\documentclass[10pt,twocolumn,twoside]{IEEEtran}
\IEEEoverridecommandlockouts                             
\usepackage[pdftex]{graphicx}
\graphicspath{{figure/}}
\usepackage{amsmath}
\usepackage{amssymb}

\usepackage{amsthm} 
\usepackage{cite}
\usepackage{color}
\usepackage{epstopdf}

\usepackage{url}

\usepackage{breakurl}

\usepackage{tabularx,booktabs}
\newcolumntype{C}{>{\centering\arraybackslash}X} % centered version of "X" type
\newcolumntype{S}{>{\hsize=.25\hsize\centering\arraybackslash}X}
\newcolumntype{M}{>{\hsize=.35\hsize\centering\arraybackslash}X}
\newcolumntype{L}{>{\hsize=1\hsize\raggedright\arraybackslash}X}
\usepackage{bm}
\usepackage{comment}
\usepackage{enumitem}

\usepackage{algorithm}
\usepackage{algorithmic}
\newcommand{\N}{\mathcal{N}}
\newcommand{\bP}{\bm{P}}

\newcommand{\bs}{\bm{s}}

\newcommand{\ba}{\bm{a}}

\newcommand{\NN}{\mathrm{NN}} % Neural Network

\newtheorem{remark}{\textbf{Remark}}

\allowdisplaybreaks

\title{Reinforcement Learning for Selective Key Applications in Power Systems: Recent Advances and Future Challenges}

\author{Xin Chen,~\IEEEmembership{Student Member,~IEEE,}
	Guannan Qu,~\IEEEmembership{Member,~IEEE,}
	Yujie Tang,~\IEEEmembership{Member,~IEEE,}\\ %<-this % stops a space
	 Steven Low,~\IEEEmembership{Fellow,~IEEE,}
Na Li,~\IEEEmembership{Member,~IEEE}
	\thanks{ X. Chen, Y. Tang, and N. Li are with the School of Engineering and Applied Sciences, Harvard University, USA. Emails: chenxin2336@gmail.com, yujietang@seas.harvard.edu,  nali@seas.harvard.edu.} 
	
\thanks{G. Qu is with Department of Electrical and Computer Engineering, Carnegie Mellon University, USA. Email: gqu@andrew.cmu.edu. }
	\thanks{ 
  S. Low is with Department of Computing and Mathematical Sciences, California Institute of Technology, USA. Email: slow@caltech.edu.
}

	\thanks{ 
The work was supported by 
NSF CAREER: ECCS-1553407,
NSF AI Institute: 2112085,
NSF     ECCS-1931662, CPS ECCS-1932611, 
Resnick Sustainability Institute, the PIMCO Fellowship, NSF   AitF-1637598, CNS-1518941, Amazon AI4Science Fellowship, Caltech Center for Autonomous Systems and Technologies (CAST).

} 
}

\begin{document}

\maketitle

\begin{abstract}

With large-scale integration of renewable generation and  distributed energy resources, 
modern power systems are confronted with  new operational challenges, such as growing complexity, increasing uncertainty, and aggravating volatility. Meanwhile, more and more data are becoming available owing to
the widespread deployment of smart meters, smart sensors, and upgraded communication networks. As  a result, data-driven control techniques, especially reinforcement learning (RL), have attracted surging attention in recent years.
This paper  provides a comprehensive review of various RL techniques and how they can be applied to decision-making and control in power systems. In particular, we 
select three key applications, i.e., frequency regulation, voltage control, and energy management, as examples to illustrate  RL-based models and solutions.
We then present  the critical issues in the application of RL,  i.e., safety, robustness, scalability, and data. Several potential future directions are discussed as well.

\end{abstract}


	\section*{Nomenclature}

\addcontentsline{toc}{section}{Nomenclature}

\subsection{Notations}
\begin{IEEEdescription}[\IEEEusemathlabelsep\IEEEsetlabelwidth{$V_1,V_2,V_3$}]
	\item [$\mathcal{A}$, $a$] Action space, action.
		\item[$\mathcal{E}$] $\subseteq \mathcal{N}\times \mathcal{N}$, the set of lines connecting buses.
	\item [$J$]   Expected  total discounted reward. 
	\item [$\mathcal{N}$] $:=\{1,\cdots,N\}$, the set of buses in a power network or the set of agents.
	\item [$\text{NN}(x;w)$]   Neural network with input $x$ and parameter $w$.
		\item [$o$]  Observation. 
		\item [$\mathbb{P}$] Transition probability.
			\item [$Q_{\pi}$]   $Q$-function (or $Q$-value) under policy $\pi$. 
				\item [$r$] Reward.
			\item [$\mathcal{S}$, $s$] State space, state.
	\item[$\mathcal{T}$] $:=\{0,1,\cdots,T\}$,  the discrete time horizon.
		\item [$\gamma$]  Discounting factor. 
		\item [$\Delta(\mathcal{A})$]  The set of probability distributions over set $\mathcal{A}$.
		\item [$\Delta t$] The time interval in $\mathcal{T}$.  
	\item [$\pi$, $\pi^*$]  Policy,  optimal policy.
\end{IEEEdescription}

\subsection{Abbreviations}
\begin{IEEEdescription}[\IEEEusemathlabelsep\IEEEsetlabelwidth{$\jmath:=\sqrt{-1}$}]
       \item[A3C] Asynchronous Advantaged Actor Critic.
       \item[ACE] Area Control Error.
       \item[AMI] Advanced Metering Infrastructure.
		\item[ANN] Artificial Neural Network.
	 \item[DDPG] Deep Deterministic Policy Gradient.
		\item[DER] Distributed Energy Resource.
		\item[DP] Dynamic Programming.
			\item[(D)RL] (Deep) Reinforcement Learning.
		\item[DQN] Deep $Q$ Network.
		\item[EMS] Energy Management System.
				\item[EV] Electric Vehicle.
		\item[FR] Frequency Regulation.
\item[HVAC]		Heating, Ventilation, and Air Conditioning.
\item[IES] Integrated Energy System.
		\item[LSPI] Least-Squares Policy Iteration.
	\item[LSTM] Long-Short Term Memory.
		\item[MDP] Markov Decision Process.
	\item[OLTC] On-Load Tap Changing Transformer.
		\item[OPF] Optimal Power Flow.
	\item[PMU] Phasor  Measurement Unit.
		\item[PV] Photovoltaic.
		\item[SAC] Soft Actor Critic. 
		\item[SCADA]  Supervisory Control and Data Acquisition.
		\item[SVC] Static Var (Reactive Power) Compensator.  
		\item[TD] Temporal Difference.
			\item[UCRL] Upper Confidence Reinforcement Learning.
\end{IEEEdescription}

\section{Introduction}

\IEEEPARstart{E}{lectric} power systems are undergoing an architectural transformation to become more sustainable, distributed, dynamic, intelligent,  and open. On the one hand,
the proliferation of renewable generation and distributed energy resources (DERs), including solar energy,  wind power, energy storage, responsive demands, electric vehicles (EVs),  etc., creates severe operational challenges. 
On the other hand, the  deployment of information,  communication, and computing technologies throughout the electric system, such as phasor measurement units (PMUs), advanced metering infrastructures (AMIs), and wide area monitoring systems (WAMS) \cite{gungor2011smart}, has been growing rapidly in recent decades.
It evolves traditional power systems towards smart grids and
offers an unprecedented opportunity to overcome these challenges through real-time data-driven monitoring and control at scale.
 This will require new advanced decision-making and control techniques  to manage
\begin{itemize}
    \item [1)] \emph{Growing complexity}. The deployment of massive DERs and the interconnection of regional power grids
    dramatically increase  system operation complexity and make it 
    difficult to obtain accurate system (dynamical) models.

\item [2)] \emph{Increasing uncertainty}. The rapid growth of renewable generation and responsive loads significantly increases uncertainty, especially when human users are involved, jeopardizing predictions and system reliability.

 \item [3)] \emph{Aggravating volatility}. The high penetration of power electronics converter-interfaced devices reduces system inertia,  which leads to faster dynamics and necessitates advanced controllers with online adaptivity.

\end{itemize}

In particular,
 reinforcement learning (RL) \cite{sutton2018reinforcement}, a prominent machine learning paradigm  
 concerned with how agents take sequential actions in an uncertain interactive environment and learn from the feedback to optimize a specific performance, can play an important role in overcoming these challenges. 
 Leveraging artificial neural networks (ANNs) for function approximation, deep RL (DRL) \cite{li2017deep}  is further developed to solve large-scale online decision problems. 
The most appealing virtue of (D)RL  is its  \emph{model-free} nature,
i.e., it makes decisions without explicitly estimating the underlying models. 
Hence,  (D)RL   has the potential to capture hard-to-model dynamics and could outperform model-based methods in highly complex tasks. 
Moreover, the data-driven nature of (D)RL   allows it to adapt to real-time observations and perform well in uncertain dynamical environments. 
Over the past decade, 
(D)RL has achieved great success in a broad spectrum of applications, such as  playing games  \cite{silver2016mastering},  robotics \cite{kober2013reinforcement},   autonomous driving \cite{sallab2017deep},  clinical trials \cite{zhao2011reinforcement}, etc.

Meanwhile, the application of  RL in power system operation and control has attracted surging  attention \cite{zhang2019deep,glavic2019deep,cao2020reinforcement,yang2020reinforcement}.  
 RL-based decision-making  mechanisms are envisioned to compensate for the limitations of existing model-based approaches  and thus are promising  to address the  emerging challenges described above. 
This paper provides a review and survey on RL-based decision-making in smart grids. 
We will introduce various RL terminologies, exemplify how to apply RL to power systems, and discuss  critical issues in their application.  Compared with  recent review articles \cite{zhang2019deep,glavic2019deep,cao2020reinforcement,yang2020reinforcement} 
on this subject,  the main merits of this paper include
\begin{itemize}
    \item [1)] We present a comprehensive and structural overview of the RL methodology, from basic concepts and theoretical fundamentals to  state-of-the-art RL techniques.
    \item [2)]  Three key applications  are selected as examples to illustrate
the overall procedure of applying RL to the  control and decision-making in power systems, from modeling, solution, to numerical implementation.
 
    \item[3)] We discuss the critical challenges and future directions for applying RL to power system problems   in depth.
\end{itemize}

 In the rest of this paper, Section \ref{sec:preonRL} presents a comprehensive overview of the RL fundamentals  and the state-of-the-art RL techniques. 
 Section \ref{sec:application} describes the application of RL to three critical power system problems, i.e., frequency regulation, voltage control,  and energy management, where paradigmatic mathematical models  are provided for illustration. Section \ref{sec:challenge} summarizes the key issues of safety, robustness, scalability, and data, and then discusses several potential future directions.  Lastly, we  conclude in Section \ref{sec:conclusion}.

\section{Preliminaries on Reinforcement Learning} \label{sec:preonRL}

\begin{figure*}
    \centering
          \includegraphics[scale=0.5]{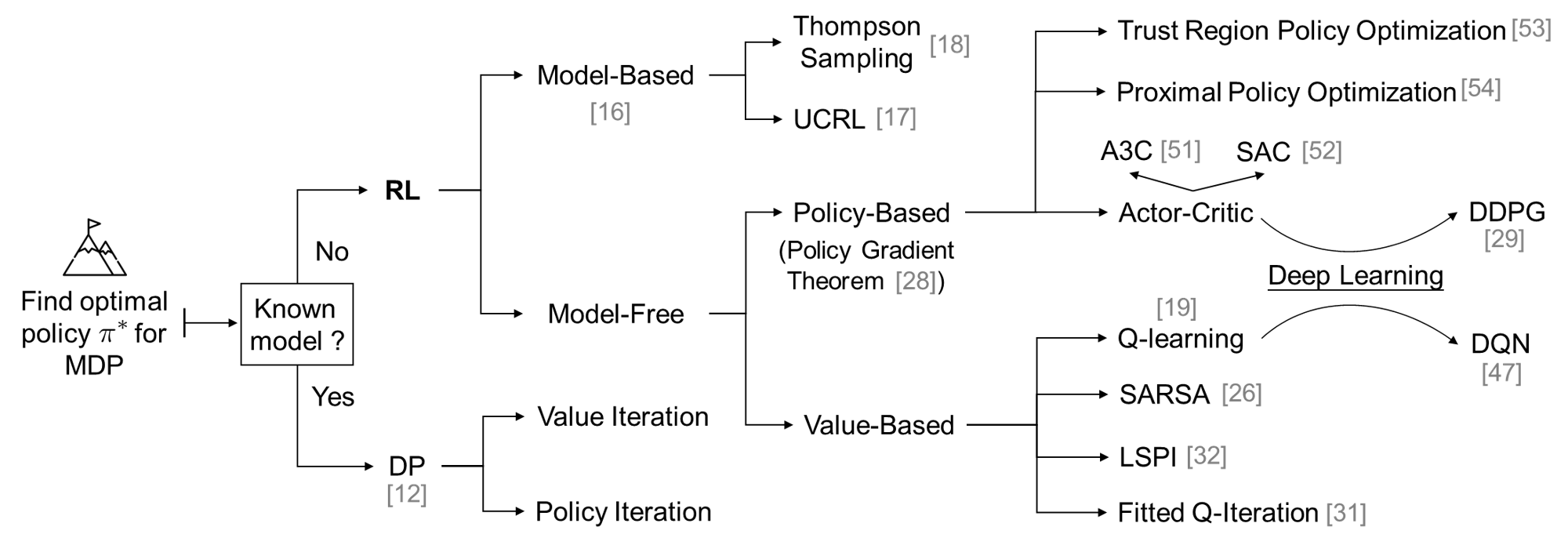}
    \caption{The structure of the RL methodology with related literature.  }   
    \label{fig:struc}
\end{figure*}

This section provides a comprehensive overview of the RL methodology. First, we set up the RL problem formulation and   key concepts, such as $Q$-function and Bellman (Optimality) Equation. Then two  categories of 
classical RL algorithms, i.e., value-based and policy-based, are introduced. With these fundamentals in place, we next present  several 
 state-of-the-art  RL techniques, including DRL, deterministic policy gradient, modern actor-critic methods, multi-agent RL, etc. The overall structure of  RL methodology with related literature is illustrated  in Fig.  \ref{fig:struc}.

\subsection{Fundamentals of Reinforcement Learning}

RL is a branch of machine learning concerned with how an agent makes  sequential decisions in an uncertain \textbf{environment} to maximize the cumulative reward. Mathematically, the decision-making problem is modeled as a Markov Decision Process (MDP), which is defined by \textbf{state} space $\mathcal{S}$, \textbf{action} space $\mathcal{A}$, the \textbf{transition} probability function $\mathbb{P}(\cdot|s,a): \mathcal{S}\times\mathcal{A}\rightarrow \Delta(\mathcal{S})$ that maps a state-action pair $(s,a)\in\mathcal{S}\times\mathcal{A}$ to a  distribution on the state space, and lastly the \textbf{reward} function $r(s,a)\footnote{ A generic reward function is given by $r(s,a,s')$ where the next state $s'$ is also included as an argument, but there is no essential difference between the case with $r(s,a)$ and the case with $r(s,a,s')$ in algorithms and results. By marginalizing over next states $s'$ according to the transition function $\mathbb{P}(s'|s,a)$, one can simply convert $r(s,a,s')$ to $r(s,a)$ 
\cite{sutton2018reinforcement}.}:\mathcal{S}\times\mathcal{A} \rightarrow \mathbb{R}$. The state space $\mathcal{S}$ and  action space $\mathcal{A}$ can be either discrete or continuous. To simplify discussion, we focus on the discrete case below.

\begin{figure}
    \centering
    \includegraphics[scale=0.38]{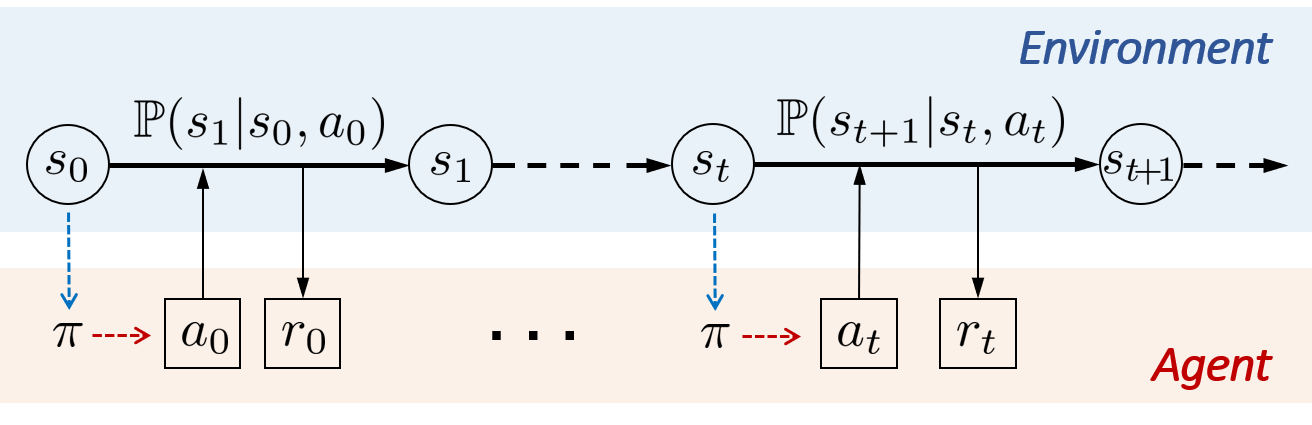}
    \caption{Illustration of a Markov Decision Process.}
    \label{fig:mdp}
\end{figure}

As illustrated in Fig.  \ref{fig:mdp},
in an MDP setting, the environment starts with an initial state $s_0\in\mathcal{S}$. At each time $t\!=\!\{0,1,\cdots\}$, given  current state $s_t \in\mathcal{S}$, the agent chooses action $a_t \in\mathcal{A}$ and receives reward $r(s_t,a_t)$ that depends on the current state-action pair $(s_t,a_t)$, 
after which the next state $s_{t+1}$ is randomly generated from the transition probability $\mathbb{P}(s_{t+1}|s_t,a_t)$.
A \textbf{policy} $\pi(a|s) \in\Delta(\mathcal{A})$ for the agent is a map from the state $s$ to  a distribution on the action space $\mathcal{A}$, which rules what action to take given a certain state $s$.\footnote{$a\sim\pi(\cdot|s)$ is  a stochastic policy, and it becomes a 
 deterministic policy $a=\pi(s)$ when the probability distribution $\pi(\cdot|s)$ is a singleton for all $s$.}
The agent aims to find an optimal policy 
 $\pi^*$ (may not be unique) that maximizes the expected infinite horizon discounted reward $J(\pi)$:
\begin{align}
  \pi^*\in \arg \max_{\pi} J(\pi) = \mathbb{E}_{s_0\sim \mu_0 } \mathbb{E}_{\pi} \sum_{t=0}^\infty \gamma^t r(s_t, a_t), \label{eq:objective}
\end{align}
where the first expectation means that $s_0$ is drawn from an initial state distribution $\mu_0$, and the second  expectation means that the action $a_t$ is taken according to the policy $\pi(\cdot|s_t)$. Parameter $\gamma\in(0,1)$ is the discounting factor that penalizes the rewards in the future.

In the MDP framework, the so-called ``\textbf{model}" specifically refers to the reward function $r$ and the transition probability $\mathbb{P}$. Accordingly, it leads to two different problem settings:
\begin{itemize}
 \item  \emph{When the model is known}, one can directly solve for an optimal policy $\pi^*$ by Dynamic Programming (DP) \cite{bertsekas2012dynamic}. 
    \item  \emph{When the model is unknown}, the agent learns an optimal policy $\pi^*$ based on the past observations from  interacting with the environment, which is the problem of RL.
\end{itemize}

Since  DP lays the foundation for RL algorithms,   we first consider the case with a known model and introduce the basic ideas of finding an optimal policy $\pi^*$ with respect to  \eqref{eq:objective}. 
The crux is the concept of \emph{$Q$-function} together with the \emph{Bellman Equation}. The $Q$-function $Q_\pi: \mathcal{S} \times \mathcal{A} \rightarrow \mathbb{R}$ for a given policy $\pi$ is defined as
\begin{align} \label{eq:qfunc}
    Q_\pi(s,a) = \mathbb{E}_{\pi} \Big [ \sum_{t=0}^\infty \gamma^t r(s_t, a_t) | s_0 = s,a_0 = a \Big ],
\end{align} 
which is the expected cumulative reward when the initial state is $s$, the initial action is $a$, and all the subsequent actions are chosen according to policy $\pi$. %It is often usefull to think of the $Q$ function as a vector
The $Q$-function $Q_\pi$ satisfies the following Bellman Equation: $\forall (s,a)\in\mathcal{S}\times\mathcal{A}$,
\begin{align}
    Q_\pi(s,a) = r(s,a) + \gamma \mathbb{E}_{s'\sim\mathbb{P}(\cdot|s,a), a'\sim \pi(\cdot|s') }Q_\pi(s',a'),\label{eq:bellman_eq}
\end{align}
where the expectation denotes that the next state $s'$ is drawn from $\mathbb{P}(\cdot|s,a)$, and the next action $a'$ is drawn from $\pi(\cdot|s')$. Here, it is helpful to think of the $Q$-function as a large table or vector filled with $Q$-values $Q_\pi(s,a)$. The Bellman Equation (\ref{eq:bellman_eq}) indicates a recursive relation that each $Q$-value equals the immediate reward plus the discounted  future value. Computing the $Q$-function for a given policy $\pi$ is called \emph{policy evaluation},
which can be done by simply solving a set of linear equations when the model, i.e., $\mathbb{P}$ and $r$, is known.

%Note that the $Q$-function $Q_\pi$ is defined for any particular policy $\pi$ which is unnecessary to be the optimal policy. 
%Such a $Q$-function is an ``evaluation'' of a given policy $\pi$: it says what is the cumulative reward under policy $\pi$. Therefore, calculating $Q_\pi(s,a)$ is often called policy evaluation and can be done by simply solving \eqref{eq:bellman_eq}, which is a set of linear equations when $\mathbb{P}$ and $r$ are known.% and it forms the ``critic'' part for the actor-critic method introduced in Section~\ref{subsec:classicrl}. 

%Obtaining $Q_\pi$ is clearly not enough for our purpose of finding the optimal policy \eqref{eq:objective}, but it is therefore we now further introduce the \emph{optimal} $Q$-function and the \emph{Bellman Optimality Equation}. 

The $Q$-function associated with an optimal policy $\pi^*$ for (\ref{eq:objective}) is called an \emph{optimal $Q$-function} and denoted as $Q^*$.  The key to find $\pi^*$ is that the optimal $Q$-function must be the unique solution to the \emph{Bellman Optimality Equation} (\ref{eq:bellman_opt_eq}): $\forall(s,a)\in\mathcal{S}\times\mathcal{A}$,
%With the definition of $Q_\pi$, we now introduce the method to find the optimal policy. Suppose an optimal policy $\pi^*$ exists for \eqref{eq:objective}. Then, the $Q$-function for this policy $Q_{\pi^*}$, which we also call \emph{optimal $Q$-function} and denote as $Q^*$, %It can be shown that the $Q$ function for an optimal policy must be the unique %\footnote{Though the optimal policy need not be unique} solution to the \emph{Bellman Optimality Equation} given by, $\forall(s,a)\in\mathcal{S}\times\mathcal{A}$,
\begin{align}
    Q^*(s,a) = r(s,a) + \gamma \mathbb{E}_{s'\sim\mathbb{P}(\cdot|s,a)} \max_{a'\in\mathcal{A}} Q^*(s',a').\label{eq:bellman_opt_eq}
\end{align}
Interested readers are referred to textbook \cite[Sec. 1.2]{bertsekas2012dynamic} on why this is true.  Based on the Bellman Optimality Equation (\ref{eq:bellman_opt_eq}), the optimal $Q$-function $Q^*$ and an optimal policy $\pi^*$ can be solved using DP or linear programming \cite[Sec. 2.4]{bertsekas2012dynamic}. In particular, \emph{policy iteration} and \emph{value iteration} are two classic DP algorithms. See \cite[Chapter 4]{sutton2018reinforcement} for details.

\begin{remark}
(Modeling Issues with MDP). \normalfont 
MDP is a generic framework to model sequential decision-making problems
and is the basis for RL algorithms. However, several issues deserve attention when modeling power system control problems in the MDP framework.
\begin{itemize}
    \item [1)] At the heart of MDP is the \emph{Markov property} that the distribution of future states depends only on the present state and action, i.e., $\mathbb{P}(s_{t+1}|s_t,a_t)$.
In other words, given the present, the future does not depend on the  past. Then for a specific control problem, one needs to check whether the choices of state and action satisfy the Markov property. A general guideline is to include all necessary known information in the enlarged state, known as state augmentation \cite{bertsekas2012dynamic}, to maintain the Markov property, however,  at the cost of complexity.  

\item[2)] Most classical MDP theories and  RL algorithms are based on \emph{discrete-time transitions}, but many power system control problems follow continuous-time dynamics, such as frequency regulation. To fit the MDP framework, continuous-time dynamics are generally discretized with a proper temporal resolution, which is a common issue for   digital control systems, and there are well-established frameworks to deal with it. Besides, there are RL variants  that are 
 built directly on continuous-time dynamics, such as integral RL  \cite{modares2014integral}. 

\item[3)] Many  MDP/RL methods assume
time-homogeneous state transitions and rewards. But in power systems, there are various time-varying exogenous inputs  and disturbances,  making the state transitions  \emph{not time-homogeneous}. This is an important problem that has not been adequately explored in the existing power literature and needs further study. \emph{Nonstationary MDP} \cite{lecarpentier2019non} and related RL algorithms \cite{cheung2020reinforcement} could be the potential directions.
\end{itemize}

In addition, the issues of continuous state/action spaces and partial observability for MDP modeling will be discussed later.
\end{remark}

\subsection{Classical Reinforcement Learning Algorithms}\label{subsec:classicrl}

This subsection considers the RL setting when the environment model is unknown, and presents classical RL algorithms for finding the optimal policy $\pi^*$.
As shown in Fig. \ref{fig:struc}, RL algorithms can be  divided into two categories, i.e., \emph{model-based} and \emph{model-free}.
``Model-based" refers to the RL algorithms that explicitly estimate and online  update an environment model from past observations and  make decisions based on this model \cite{moerland2020model}, such as  upper confidence RL (UCRL) \cite{jaksch2010near} and Thompson sampling \cite{russo2017tutorial}. In contrast, ``model-free" means that the  RL algorithms directly search for optimal policies  without estimating the environment model.
 Model-free RL algorithms  are mainly categorized into two types, i.e., \emph{value-based} and \emph{policy-based}.
Generally, value-based methods are preferred for modest-scale RL problems with finite state/action space as they do not assume a policy class and have a strong convergence guarantee.  The convergence of value-based methods to an optimal $Q$-function in the 
tabular setting (without function approximation) was proven back in the 1990s \cite{tsitsiklis1994asynchronous}.
In contrast, policy-based methods are 
more efficient for problems with high dimensions or continuous action/state space. But they are known to suffer from various convergence issues, e.g.,  local optimum,  high variance, etc.  The
convergence of policy-based methods with restricted policy classes to the global optimum   has been  shown in a recent work  \cite{agarwal2020optimality} under the tabular policy  parameterization.

\begin{remark}\label{remark:explore}
(Exploration vs. Exploitation). \normalfont A fundamental problem faced by  RL algorithms is the dilemma between exploration and exploitation. Good performance requires taking  actions  adaptively to strike an effective balance between 1) \emph{exploring} poorly-understood actions to  gather new information that may improve future reward, and 2) \emph{exploiting} what is known for decision-making to maximize immediate reward. Generally, it is natural to achieve  exploitation  with the goal of reward maximization, while different RL algorithms encourage exploration in different ways. For value-based RL algorithms, $\epsilon$-greedy is commonly used with a probability of $\epsilon$ to explore random actions. In policy-based methods,  exploration is usually achieved by injecting random perturbation to the actions, adopting a stochastic policy, or adding an entropy term to the objective, etc.
\end{remark}

Before presenting  classical RL methods, we introduce a key algorithm, 
\textbf{Temporal-Difference} (TD) learning \cite{580874}, 
for policy evaluation when the model is unknown. TD learning is central to both value-based and policy-based RL algorithms. 
It learns the $Q$-function $Q_\pi$ for a given policy $\pi$ from episodes of experience. Here, an ``episode" refers to a state-action-reward trajectory over time $(s_0,a_0,r_0, s_1,a_1,r_1,\cdots)$ until terminated. Specifically, TD learning maintains  a $Q$-function $Q(s,a)$ for all state-action pairs $(s,a)\in\mathcal{S}\times\mathcal{A}$ and updates it upon a new observation $(r_t, s_{t+1},a_{t+1})$ by 
\begin{align}\label{eq:td_learning}
\begin{split}
    Q(s_t,a_t) &\leftarrow \, Q(s_t,a_t) \\
    &+ \alpha( r_t + \gamma  Q(s_{t+1},a_{t+1}) -Q(s_t,a_t)), % \nonumber\\
   % = \,& (1\!-\alpha)Q(s_t,a_t) + \alpha ( r_t + \gamma  Q(s_{t+1},a_{t+1}))
\end{split}
\end{align}
where $\alpha$ is the step size. Readers might observe that the second term in \eqref{eq:td_learning} is very similar to  the Bellman Equation \eqref{eq:bellman_eq}, which is exactly the rationale  behind TD learning. Essentially, TD learning (\ref{eq:td_learning}) is a stochastic approximation scheme  for solving the Bellman Equation~\eqref{eq:bellman_eq} \cite{robbins1951stochastic}, and can be shown to converge to the true $Q_\pi$ under mild assumptions \cite{580874,srikant2019finite}.

%The goal of RL is to find the optimal policy \eqref{eq:objective} by learning from past interactions with the system, without knowing the reward function and the transition kernel. Here we introduce two classical RL algorithms.

%\subsubsection{Value-based RL}

\textit{1) Value-based RL algorithms}   directly learn the optimal $Q$-function $Q^*$, and the optimal (deterministic) policy $\pi^*$ is a byproduct that can be retrieved by acting greedily, i.e., $\pi^*(s) = \arg\max_{a\in\mathcal{A}} Q^*(s,a)$.  Among many, \textbf{$\bm{Q}$-learning}  \cite{watkins1992q,tsitsiklis1994asynchronous} is perhaps the most popular value-based RL algorithm.
Similar to TD-learning, $Q$-learning maintains a $Q$-function and updates it towards the optimal $Q$-function based on episodes of experience.
Specifically, at each time $t$, given  current  state $s_t$, the agent chooses action $a_t$ according to a certain behavior policy.\footnote{Such a behavior policy, also called exploratory policy,
%specifically refers to the policy that generates the actual actions in the training episodes. It
can be arbitrary as long as it visits  all the states and actions sufficiently often.} Upon observing the outcome $(r_t, s_{t+1})$, 
 $Q$-learning updates the $Q$-function  by 
\begin{align}  \label{eq:q_learning}
\begin{split}
       Q(s_t,a_t) & \leftarrow  Q(s_t,a_t)  \\
    & +\alpha (r_t + \gamma  \max_{a'\in\mathcal{A}}Q(s_{t+1},a') - Q(s_t,a_t)).
\end{split}
\end{align}
%which is same as TD learning (\ref{eq:td_learning}) except replacing $a_{t+1}$ with an optimal action $a'$. 
The rationale behind \eqref{eq:q_learning} is that  the $Q$-learning algorithm \eqref{eq:q_learning} is essentially a stochastic approximation scheme for solving the Bellman Optimality Equation~\eqref{eq:bellman_opt_eq}, and one can show the convergence to  $Q^*$ under mild assumptions \cite{tsitsiklis1994asynchronous,qu2020finite}. 

{SARSA}\footnote{In some literature, SARSA is  also called $Q$-learning.}  
 \cite{rummery1994line} is another classical value-based RL algorithm, whose name comes from the experience sequence $(s,a,r,s',a')$. SARSA is actually an \textit{on-policy} variant of $Q$-learning. The major difference  is that SARSA 
  takes actions  according to the target policy (typically $\epsilon$-greedy based on the current $Q$-function) rather than any
  arbitrary behavior policy in $Q$-learning. 
  The following remark distinguishes and compares ``on-policy" and ``off-policy" RL algorithms.
  
  %We defer the discussion on ``on-policy" and ``off-policy" to Remark \ref{remark:on-off}. 

 \begin{remark}\label{remark:on-off}
(On-Policy vs. Off-Policy).  \normalfont  On-policy RL methods continuously improve a policy (called target policy) and implement this policy to generate episodes for algorithm training. In contrast, off-policy RL methods learn a target policy based on the episodes that are generated by following a different  policy (called behavior policy) rather than the target policy itself. In short, ``on" and ``off" indicate whether the training samples are generated by following the target policy or not. 
For example, $Q$-learning is an off-policy RL method as the episodes used in training can be produced by any policies, and
the actor-critic algorithm described below  is on-policy.\footnote{There are also off-policy variants of the actor-critic algorithm, e.g., \cite{degris2012off}.} 
For power system applications,  control policies that are not well-trained are generally not allowed to be implemented in  real-world power grids for the sake of safety. Thus off-policy RL is   preferred when  high-fidelity  simulators are unavailable
 since it can learn from the vast amount of  operational data generated by incumbent controllers. Off-policy RL is also relatively easy to provide a safety guarantee due to the flexibility in choosing the behavior policies, but it is known to suffer from slower convergence and higher sample complexity.

%For power system applications, off-policy RL is quite appealing because the huge amount of real operational data  that are generated by existing controllers can be utilized. 

\end{remark}

\textit{2) Policy-based RL algorithms} restrict  the optimal policy search to a policy class  that is parameterized as $\pi_\theta$ with the  parameter $\theta\in \Theta\subseteq \mathbb{R}^K$. With this  parameterization,
the objective  \eqref{eq:objective} can be rewritten as a function of the policy parameter, i.e., $J(\theta)$, and the RL problem is 
 reformulated as an optimization problem (\ref{eq:RLopt}) that aims to find the optimal $\theta^*$:
\begin{align}\label{eq:RLopt}
   \theta^*\in \arg\max_{\theta\in\Theta} J(\theta).
\end{align}
To solve (\ref{eq:RLopt}), a straightforward idea is to employ the gradient ascent method, i.e., $\theta\leftarrow \theta + \eta \nabla J(\theta)$, where $\eta$ is the step size. However, computing the gradient $\nabla J(\theta)$ was supposed to be intrinsically hard as the environment model is unknown.  \emph{Policy Gradient Theorem} \cite{sutton2000policy} is a big breakthrough in addressing the gradient computation issue. This theorem shows that the policy gradient $\nabla J(\theta)$  can be simply expressed as 
\begin{align}\label{eq:policygrad}
     \nabla J(\theta) = \sum_{s\in \mathcal{S}} \mu_\theta (s) \sum_{a\in \mathcal{A}} \pi_\theta(a|s)Q_{\pi_\theta}(s,a) \nabla_\theta \ln \pi_\theta(a|s).
\end{align}
Here, $\mu_\theta(s)\in\Delta(\mathcal{S})$ is the on-policy state distribution \cite[Chapter 9.2]{sutton2018reinforcement}, which denotes the fraction of time steps spent in each state $s\in \mathcal{S}$.  
Equation (\ref{eq:policygrad}) is for a stochastic policy $a\sim \pi_\theta(\cdot|s)$, while  the version of policy gradient theorem for a deterministic policy $a=\pi_\theta(s)$ \cite{silver2014deterministic} will be discussed later.

The policy gradient theorem  provides a highway to estimate the  gradient $\nabla J(\theta)$, which lays the foundation for  policy-based RL algorithms. In particular, the \textbf{actor-critic} algorithm is a prominent and widely used  architecture based on policy gradient.
It comprises  two eponymous components: 1) the ``critic'' is to estimate the $Q$-function  $Q_{\pi_\theta}(s,a)$, and 2) the ``actor'' conducts the gradient ascent  based on (\ref{eq:policygrad}). The following iterative scheme is an illustrative actor-critic example:
\begin{enumerate}
\item [1)]Given state $s$, take action $a\sim \pi_\theta(a|s)$, then observe the reward $r$ and next state $s'$;
    \item  [2)](Critic) Update $Q$-function $Q_{\pi_\theta}(s,a)$ by TD learning;
    \item  [3)] (Actor) Update policy parameter  $\theta$ by
    \begin{align} \label{eq:actorup}
        \theta\leftarrow \theta + \eta\, Q_{\pi_\theta}(s,a) \nabla_\theta \ln\pi_\theta(a|s);
    \end{align}
    \item   [4)] $s\leftarrow s'$. Go to step 1) and repeat. 
\end{enumerate}
There are many actor-critic variants   with different implementation manners, e.g., how $s$ is sampled from $\mu_\theta(s)$, how $Q$-function is updated, etc. See \cite[Chapter 13]{sutton2018reinforcement} for more details.
We emphasize that the algorithms introduced above are far from complete. 
In the next subsections, we will introduce the  state-of-the-art modern (D)RL techniques  that are widely used in complex control tasks, especially for 
power system applications.
 At last, we close this subsection with the following two remarks on different RL settings.
 %Remark \ref{remark:batchRL}. % two crucial remarks.

%\emph{Trust Region Policy Optimization} \cite{schulman2015trust} is invented to stabilize the training process with guaranteed monotonic improvement over policy iteration, which  imposes a KL-divergence constraint to avoid dramatic policy changes.

%\begin{remark}\label{remark:explore}
%(Exploration vs. Exploitation). \normalfont A fundamental problem faced by the RL algorithms is the dilemma between exploration and exploitation. Good performance requires taking  actions in an adaptive way that strikes an effective balance between 1) \emph{exploring} poorly-understood actions to  gather new information that may improve future reward and 2) \emph{exploiting} what is known for decisions to maximize immediate reward. Generally, it is natural to achieve  exploitation  with the goal of reward maximization, while different RL algorithms encourage exploration in different ways. For on-policy value-based RL algorithms, $\epsilon$-greedy is commonly utilized with a probability of $\epsilon$ to explore random actions. In policy-based methods,  the exploration is usually realized by adding random perturbation to the actions or adopting a stochastic policy.
%\end{remark}

\begin{remark}\label{remark:batchRL}
(Online RL vs. Batch RL). \normalfont The algorithms introduced above are referred to as ``\emph{online RL}" that takes actions and updates the policies simultaneously. In contrast,  there is another type of RL called ``\emph{batch RL}"  \cite{lange2012batch}, which decouples the sample data collection and policy training. 
Precisely,   given a set of     experience  episodes generated by following any  arbitrary behavior policies, 
batch RL fits the optimal $Q$-function or
optimizes the target policy entirely based on this fixed sample dataset.  Some classical 
batch RL algorithms include  Fitted $Q$-Iteration \cite{ernst2005tree}, Least-Squares Policy Iteration (LSPI) \cite{lagoudakis2003least}, etc. For example,
given a batch of transition experiences $\mathcal{D}:=\{(s_i,a_i,r_i,s_i')_{i=1}^n\}$, 
Fitted $Q$-Iteration, which is seen as the batch version of $Q$-learning, aims to fit a parameterized $Q$-function $Q_\theta(s,a)$ by iterating the following two steps:
\begin{itemize}
    \item [1)] Create the target  $Q$-value $q_i$ for each sample in $\mathcal{D}$ by $$q_i = r_i + \gamma \max_{a'}Q_\theta(s_i',a').$$
    \item [2)] Apply  regression approaches to fit a new $Q_\theta(s,a)$ based on the training dataset $(s_i,a_i;q_i)_{i=1}^n$.
\end{itemize}
The crucial advantages  of batch RL lie in the stability and data-efficiency of the learning process by making the best use of the available sample datasets. However, because of  relying entirely on a given dataset, the lack of exploration is one of the major problems of batch RL.
To encourage exploration, batch RL typically iterates between exploratory sample collection and policy learning prior to application. Besides, pure batch RL, also referred to as 
``offline RL" \cite{agarwal2020optimistic}%\footnote{In literature, it is  arguable to set  clear boundaries among online RL, batch RL and offline RL.},
, has attracted increasing recent attention, which completely ignores the exploration issue and aims to learn policies fully based on a static dataset with no online interaction. Offline RL   assumes a sufficiently large and diverse dataset 
that  adequately covers high-reward transitions for learning good policies and turns 
 the RL problem into a supervised machine learning problem.  
 See \cite{levine2020offline} for a tutorial of offline RL.
\end{remark}

\begin{remark} (Passive RL, Active RL, and Inverse RL). \normalfont The terminology ``passive RL" typically refers to the RL setting where the agent 
 acts by following a fixed policy $\pi$ and aims to learn how good this policy is from observations. It is analogous to the policy evaluation task, and TD learning is one of the representative algorithms of passive RL. In contrast, ``active RL" allows the agent to update policies with the goal of finding an optimal policy, which is basically the standard RL setting that we described above. However, in some references, e.g.,  \cite{krueger2020active,daniel2014active}, ``active RL" has a completely different meaning and refers to the RL variant where the agent does not observe the reward unless it pays a query cost to account for the difficulty of collecting reward feedback. Thus, the agent chooses both an action and whether to observe the reward  at each time.  Another interesting RL variant is ``inverse RL" \cite{ng2000algorithms,fu2017learning}, in which the state-action sequence of an (expert) agent is given,  and the task is to infer the reward function that this agent seeks to maximize. Inverse RL is motivated by various practical applications where the reward engineering is complex or expensive and  one can observe an expert demonstrating the task  to learn how to perform, e.g., autonomous driving. %The idea of inverse RL  can be applied to power systems to design effective reward functions with the teaching of expert systems.

\end{remark}

\subsection{Fundamentals of Deep Learning}

\begin{figure}
    \centering
    \includegraphics [scale=0.35] {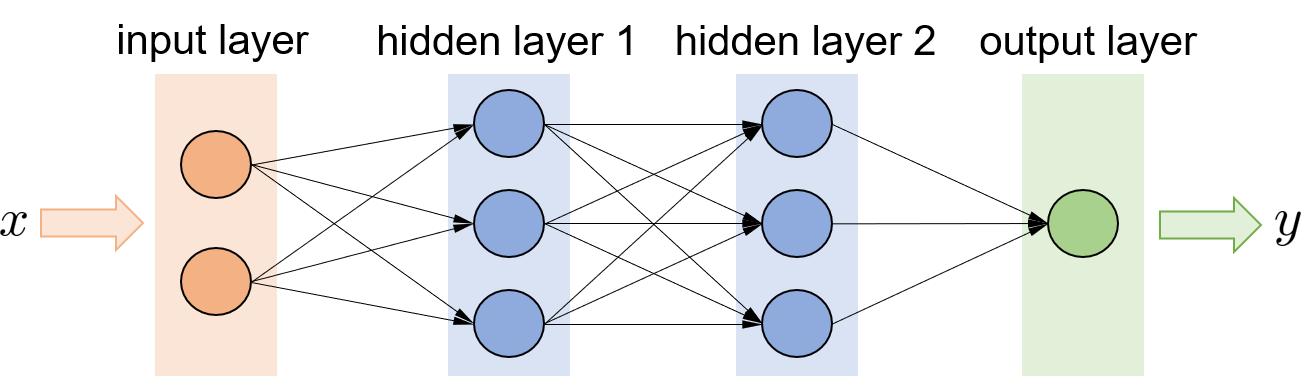}
    \caption{Illustration of a regular   four-layer feed-forward ANN \cite{annfig}.}
    \label{fig:ANN}
\end{figure}

This subsection presents the fundamentals of deep learning to set the stage for the introduction of DRL.
Deep learning refers to the machine learning technique that models with multi-layer ANNs. The history of ANNs dates back to 1940s \cite{wang2017origin}, and it has received tremendous interests in the recent decade due to the booming of data technology and computing power, which allows efficient training of wider and deeper ANNs.  Essentially, an ANN is an universal parameterized mapping $y = \text{NN}(x;w)$ from the input features $x$ to the outputs $y$ with the parameters $w$.
As illustrated in Fig.  \ref{fig:ANN}, an input feature vector $x$ is taken in by the input layer, then is processed through a series of hidden layers, and results in the output vector $y$. 
 Each hidden layer consists of a number of neurons that are the   activation functions, e.g. linear,  ReLU, or sigmoid \cite{lecun2015deep}. Based on a sample dataset $(x^i, y^i)_{i=1,\cdots,n}$, the parameter $w$ can be optimized via regression. A landmark in training ANNs is the discovery of the \emph{back-propagation} method, which  offers an efficient way to compute the gradient of the loss function over $w$ \cite{sun2019optimization}.
 Nevertheless, it is pretty tricky to train large-scale ANNs in practice, and article \cite{sun2019optimization}  provides an
overview of the optimization algorithms and theory for training ANNs.
 Three typical classes of ANNs with different architectures are introduced below. See book \cite{Goodfellow-et-al-2016} for details.

\emph{1) Convolutional Neural Networks (CNNs)} 
 are in the architecture of feed-forward neural networks (as shown in Fig.  \ref{fig:ANN}) and specialize in pattern detection, which are powerful for image analysis and computer vision tasks. 
The convolutional hidden layers  are  the basis at the heart of 
a CNN.  Each neuron $k=1,2,\cdots$ in a convolutional layer defines a  small filter  (or kernel) matrix $F_k$ of low dimension (e.g., $3\times 3$) and
convolves with the  input matrix $X$ of relatively high dimension, which leads to the output matrix $U_k = F_k\otimes X$. Here, $\otimes$ denotes the convolution operator\footnote{Specifically, the convolution operation is performed by
sliding the filter matrix $F_k$ across the input matrix $X$ and 
computing the corresponding element-wise multiplications, so that  each element in matrix $U_k$ is the sum of the element-wise multiplications between $F_k$ and the associated  sub-matrix of $X$. See \cite[Chapter 9]{Goodfellow-et-al-2016} for a detailed definition of convolution.}, and the output $(U_k)_{k=1,2,\cdots}$ is referred to as the  feature map that is passed to the next layer.
Besides, pooling layers are commonly used to reduce the dimension of the representation with
the max or average pooling. %The final layers are fully connected to make the recognition reasoning based on the extracted features.

%while the outputs are referred as the extracted features or patterns.

\emph{2) Recurrent Neural Networks (RNNs)} specialize in processing long sequential inputs and tackling tasks with context spreading over time by leveraging a recurrent  structure. Hence, RNNs achieve great success in the applications such as speech recognition and machine translation.  RNNs process an input sequence one element at a time, and maintain in their hidden units a state vector $s$ that implicitly
contains historical information about the past elements. 
Interestingly, the recurrence of RNNs is  analogous to a dynamical system \cite{Goodfellow-et-al-2016} and 
can be expressed as 
\begin{align}
    s_t = f(s_{t-1}, x_t; v),\ \ y_t = g(s_t;u),
\end{align}
where $x_t$ and $y_t$ are the input and output of the neural network at time step $t$, and $w:=(v,u)$ is the parameter for training.
$s_t$ denotes the state stored in the hidden units at step $t$ and will be passed to the processing at step $t+1$. In this  way, $s_t$ implicitly covers all historical input information $(x_1,\cdots,x_t)$.  Among many variants, long-short term memory  (LSTM) network \cite{hochreiter1997long} is a special type of RNNs that excels at handling long-term   dependencies and outperforms conventional RNNs by using special memory cells and gates.

\begin{figure}
    \centering
    \includegraphics [scale=0.32] {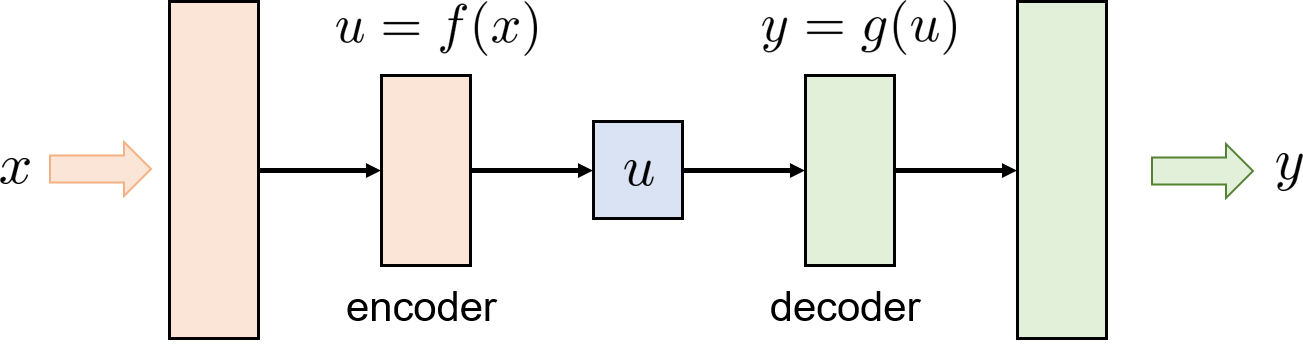}
    \caption{Illustration of a simple autoencoder network \cite{autofig}.}
    \label{fig:encoder}
\end{figure}

\emph{3) Autoencoders} \cite{hinton2006reducing} are used to obtain a low-dimensional representation of high-dimensional inputs, which is similar to, but more general than, principal components analysis (PCA). As illustrated in Fig.  \ref{fig:encoder}, an autoencoder is in an hourglass-shape feed-forward network structure and
consists of a encoder function $u=f(x)$ and a decoder function $y=g(u)$. In particular, an autoencoder is trained to learn an approximation function $y = \text{NN}(x;w)=g(f(x))\approx x$ with the loss function  $L(x, g(f(x)))$ that penalizes the dissimilarity between the input $x$ and output $y$.
The bottleneck layer has a much smaller amount of neurons, and thus it is forced to form a  compressed representation  of the input $x$.

\subsection{Deep Reinforcement Learning}

For many practical problems, the state and action spaces are large or continuous, together with complex system dynamics. As a result, it is intractable for value-based RL to compute or store a  gigantic  $Q$-value table for  all state-action pairs.  To deal with this issue, \emph{function approximation} methods
are developed to approximate  the $Q$-function with some parameterized function classes, such as  linear function or polynomial function. As for policy-based RL, finding a capable policy class  to achieve optimal control is also nontrivial in high-dimensional complex tasks.
Driven by the advances of deep learning, DRL  
that leverages ANNs for function approximation or policy parameterization is
becoming increasingly popular.
Precisely, DRL can use ANNs to 
   1) approximate the $Q$-function with a $Q$-network $\hat{Q}_w(s,a):= \NN(s,a;w)$,  and
   2) parameterize  the policy with the policy network $\pi_\theta(a|s): = \NN(a|s;\theta)$.
%   We elaborate these two usages as follows.

\textit{1)  $Q$-Function Approximation}. $Q$-network can be used to approximate the  $Q$-function in TD learning (\ref{eq:td_learning}) and $Q$-learning (\ref{eq:q_learning}). For TD learning, the parameter $w$ is updated by 
\begin{align} \label{eq:qnettd}
\begin{split}
     w\leftarrow w + \alpha \Big[r_t &+ \gamma\,\hat{Q}_w(s_{t+1},a_{t+1}) \\
     &- \hat{Q}_w(s_t,a_t) \Big] \nabla_w  \hat{Q}_w (s_t,a_t),
\end{split}
\end{align}
where the gradient $\nabla_w  \hat{Q}_w (s_t,a_t)$ can be calculated efficiently using the back-propagation method.
As for  $Q$-learning,   it is  known that adopting a nonlinear function, such as  an ANN, for
 approximation 
 may cause instability and divergence issues 
in the training process. To this end, 
 \textbf{Deep $\bm{Q}$-Network} (DQN) \cite{mnih2015human} is developed and  greatly improves
 the training stability of $Q$-learning with  the following two  tricks:

 $\bullet$ \textit{Experience Replay}.  Instead of performing on consecutive episodes, a
widely used trick is to
  store all  
 transition experiences $e:=(s,a,r,s')$ in a database  $\mathcal{D}$ called ``replay buffer". 
 At each step, a  batch of transition experiences is randomly sampled from the replay buffer $\mathcal{D}$ for  $Q$-learning update. This can enhance the data efficiency by recycling previous experiences and reduce the variance of learning updates. More importantly,  sampling uniformly from the replay buffer breaks the temporal correlations  that  jeopardize the training process, and thus improves the  stability and convergence of $Q$-learning.

 $\bullet$ \textit{Target Network}. The other trick is the introduction of the target network 
 $\hat{Q}_{\hat{w}}(s,a)$  with  parameter $\hat{w}$, which is a clone of  the  $Q$-network $\hat{Q}_w(s,a)$. Its parameter $\hat{w}$ is kept frozen and is only  updated periodically.
 Specifically,
with a batch of transition experiences $(s_i,a_i,r_i,s_i')^{n}_{i=1}$ sampled from the replay buffer, the 
 $Q$-network  $\hat{Q}_w(s,a)$ is updated by solving 
\begin{align}
      w\! \leftarrow \!\arg\min_{w}\! \sum_{i=1}^n\! \big( r_i +\!\gamma\max_{a'}\hat{Q}_{\hat{w}}(s'_i,a') \!-\!  \hat{Q}_w(s_i,a_i) \big)^2. \label{eq:dqnl}
\end{align}
The optimization (\ref{eq:dqnl}) can be viewed as finding an optimal $Q$-network $\hat{Q}_w(s,a)$ that approximately solves the 
Bellman Optimality Equation (\ref{eq:bellman_opt_eq}). The critical difference  is that the target network $\hat{Q}_{\hat{w}}$ with parameter $\hat{w}$ instead of $\hat{Q}_{w}$ is used to compute the maximization over $a'$  in (\ref{eq:dqnl}). 
 After a fixed number of updates above, 
 the target network $\hat{Q}_{\hat{w}}(s,a)$ is renewed by replacing  $\hat{w}$ with the latest learned $w$. This trick can mitigate the training instability as the short-term oscillations are circumvented. See \cite{fan2020theoretical} for more details. 
 
 In addition, there are several notable variants of DQN that further improve the performance, such as double DQN \cite{van2016deep} and  dueling DQN \cite{wang2016dueling}. Particularly, double DQN is proposed to tackle the overestimation issue of the action values in DQN by learning two sets of $Q$-functions; one  $Q$-function is used to select the action, and the other  is used to determine its value.
Dueling DQN proposes a dueling network architecture that separately estimates the state value function $V(s)$ and the state-dependent action advantage function $A(s,a)$, which are then combined to determine the  $Q$-value. 
The main benefit of this factoring is to generalize learning across actions without imposing any
change to the underlying RL
algorithm \cite{wang2016dueling}. % Besides, the above two variants can be integrated as dueling double DQN, leading to better policy evaluation.

%maintains two $Q$ estimates and addresses overestimation issues found in DQN;  Dueling DQN  generalizes learning across actions without imposing any change to the underlying reinforcement learning algorithm. 

\textit{2) Policy Parameterization.} Due to the powerful generalization capability, ANNs are widely used to parameterize control policies,  especially when the state and action spaces are continuous. 
 The resultant policy network $\NN(a|s;\theta)$ takes states as the input and outputs the probability of  action selection. In  actor-critic methods, it is common to adopt both   the $Q$-network  $\NN(s,a;w)$ and the policy network $\NN(a|s;\theta)$ simultaneously,  where   the ``actor" updates $\theta$ according to (\ref{eq:actorup}) and the ``critic"  updates $w$ according to (\ref{eq:qnettd}). The back-propagation method \cite{sun2019optimization}   can be applied to efficiently compute the gradient of ANNs.

 When function approximation is adopted, the theoretical analysis on both value-based and policy-based RL methods is little and generally limited to
the linear function approximation.
Besides,  one problem that hinders the use of value-based methods for large or continuous action space is the difficulty of 
performing the maximization step.
For example, when deep ANNs are used to approximate the $Q$-function, it is not easy to solve $\max_{a'} \hat{Q}_w(s,a')$ for the optimal action $a'$ due to the nonlinear and complex  formulation of $\hat{Q}_w(s,a)$.

\subsection{Other Modern Reinforcement Learning Techniques}\label{sec:othermodern}

This subsection summarizes  several state-of-the-art modern RL techniques that are widely used in complex  tasks.

\subsubsection{Deterministic Policy Gradient}

The RL algorithms described above focus on 
stochastic policies $a\sim\pi_\theta(\cdot|s)$, while deterministic policies $a=\pi_\theta(s)$ are more desirable for many real-world control problems with continuous state and action spaces. On the one hand,   since most incumbent controllers in physical systems, such as PID control and robust control, are all deterministic, deterministic policies  are better matched to  the  practical control architectures, e.g., in power system applications.
On the other hand, a deterministic policy is more sample-efficient as its policy gradient only integrates over the state space. In contrast, a stochastic policy gradient integrates over both state and action spaces \cite{silver2014deterministic}. Similar to the stochastic case, there is a \emph{Deterministic Policy Gradient Theorem} \cite{silver2014deterministic} showing that the  policy gradient 
with respect to a deterministic policy $\pi_\theta(s)$ can be simply expressed as
\begin{align}
    \nabla J(\theta) = \mathbb{E}_{s\sim \mu_\theta(\cdot)}  \nabla_a Q_{\pi_\theta}(s,a)|_{a=\pi_\theta(s)}  \nabla_\theta \pi_\theta(s).
\end{align}
Correspondingly, the ``actor"  in the actor-critic algorithm can update the parameter $\theta$ by 
\begin{align}
    \theta \leftarrow \theta + \eta  \nabla_a Q_{\pi_\theta}(s,a)|_{a=\pi_\theta(s)} \nabla_\theta  \pi_\theta(s).
\end{align}
One major issue regarding a deterministic policy is the lack of 
 exploration due to the determinacy of action selection. A common way to encourage exploration is to 
 perturb
a deterministic policy  with exploratory noises, e.g., adding a Gaussian noise $\xi$ to the policy with
 $a = \pi_\theta(s) + \xi$.

%two methods are usually adopted: 1) perturb the policy with exploration noises, e.g., adding a Gaussian noise $\xi$ with  $a = \pi_\theta(s) + \xi$; 2) learn the policy in an off-policy manner where the training samples are collected by following a stochastic behavior policy \cite{silver2014deterministic}. 

\subsubsection{Modern Actor-Critic Methods} Although 
achieving great success in many complex tasks, 
the actor-critic methods are known to suffer from various problems,  such as  high variance, slow convergence, local optimum, etc. Therefore, many variants have been developed to improve  the performance of actor-critic, and we list some of them below.

 $\bullet$ \emph{Advantaged Actor-Critic} \cite[Chapter 13.4]{sutton2018reinforcement}: 
The \emph{advantage function}, $A(s,a)\!=\! Q_{\pi_\theta}(s,a) -{baseline}$, i.e., the $Q$-function subtracted by a baseline,
is introduced to replace $Q_{\pi_\theta}(s,a)$ in the ``actor" update, e.g., (\ref{eq:actorup}). One  common choice for the baseline is an estimate of the state value function $V(s)$. This modification can  significantly reduce the variance of the policy gradient estimate without changing the expectation.

  $\bullet$ \emph{Asynchronous Actor-Critic} \cite{mnih2016asynchronous} presents an asynchronous  variant with parallel training to enhance sample efficiency and training stability.  In this method, multiple actors are trained in parallel with different exploration polices, then the global parameters get updated based on all the learning results and  synchronized to each actor.

  $\bullet$ \emph{Soft Actor-Critic} (SAC) \cite{haarnoja2018soft} with stochastic policies is an off-policy deep actor-critic algorithm based on the
maximum entropy RL framework, which adds  an entropy term of the policy $\mathcal{H}(\pi_\theta(\cdot|s_t))$ to  objective (\ref{eq:objective}) to encourage exploration.

\subsubsection{ Trust  Region/Proximal Policy Optimization} To improve the training stability of policy-based RL algorithms, reference \cite{schulman2015trust} proposes the Trust Region Policy Optimization (TRPO) algorithm, which enforces a trust region constraint \eqref{eq:trpo:con}. It indicates that the KL-divergence $\mathrm{D}_{KL}$ between the old and new policies  should not be larger than a given threshold $\delta$.  Denote $\rho(\theta):= \frac{\pi_\theta(a|s)}{\pi_{\theta_{\mathrm{old}}}(a|s)}$ as the probability ratio between the new policy $\pi_\theta$ and the old policy $\pi_{\theta_{\mathrm{old}}}$ with $\rho(\theta_{\mathrm{old}})=1$. Then, TRPO aims to solve the constrained optimization \eqref{eq:trpo}:
\begin{subequations}\label{eq:trpo}
\begin{align}
    \max_\theta &\  \ J(\theta)=\mathbb{E}[\rho(\theta) \hat{A}_{\theta_{\mathrm{old}}}(s,a)],\\
    \text{s.t. } &\ \ 
     \mathbb{E}[\mathrm{D}_{KL}(\pi_\theta||\pi_{\theta_{\mathrm{old}}})]\leq \delta, \label{eq:trpo:con}
\end{align}
\end{subequations}
where $\hat{A}_{\theta_{\mathrm{old}}}(s,a)$ is an estimation of the advantage function. However, TRPO is hard to implement. 
To this end,  work \cite{schulman2017proximal} proposes 
Proximal Policy Optimization (PPO) methods, which achieve the benefits of TRPO with simpler implementation and better empirical sample complexity. Specifically,  PPO simplifies the policy optimization as \eqref{eq:ppo}:
\begin{align}\label{eq:ppo}
     \max_\theta \,  \mathbb{E}\Big[\! \min\Big( \rho(\theta) \hat{A}_{\theta_{\mathrm{old}}}, \text{clip}( \rho(\theta), 1\!-\!\epsilon,1\!+\!\epsilon  )\hat{A}_{\theta_{\mathrm{old}}}  \Big)\Big],
\end{align}
where $\epsilon$ is a hyperparameter, e.g., $\epsilon=0.2$, and the clip function $\text{clip}( \rho(\theta), 1-\epsilon,1+\epsilon )$ enforces $\rho(\theta)$ to stay within the interval $[1-\epsilon,1+\epsilon]$. 
The ``$\min$" of the clipped and unclipped objectives is taken to eliminate the incentive for moving $\rho(\theta)$ outside of the
interval $[1\!-\!\epsilon,1\!+\!\epsilon]$ to prevent big   updates of $\theta$.

\subsubsection{Multi-Agent RL} Many  power system control tasks involve the coordination over  multiple agents. For example, in frequency regulation, each generator can be treated as an individual agent that makes its own generation decisions, while the frequency dynamics  are  jointly determined by  all power injections. It motivates the \emph{multi-agent RL} framework, which considers a set  $\mathcal{N}$ of agents interacting with the same environment and sharing a common state $s\in \mathcal{S}$. At each time $t$, each agent $i\in\mathcal{N}$ takes its own action $a_t^i\in \mathcal{A}_i$ given the current state $s_t\in \mathcal{S}$, and receives the reward $r^i(s_t, (a_t^i)_{i\in\mathcal{N}})$, then the system state evolves to $s_{t+1}$ based on  $(a_t^i)_{i\in\mathcal{N}}$. Multi-agent RL is an active and challenging  research area with many unsolved problems.
An overview on related  theories and algorithms is provided in \cite{zhang2019multi}. In particular, 
the decentralized (distributed)  multi-agent RL attracts a great deal of attention 
for  power system applications. A popular variant is that each agent $i$ adopts a local policy $a^i = \pi_{i,\theta_i}(o^i)$ with its parameter $\theta_i$, which determines the action $a^i$ based on local observation $o^i$ (e.g., local voltage or frequency of bus $i$). This method allows for decentralized implementation as the policy $\pi_{i,\theta_i}$ for each agent only needs local observations, but it still requires centralized training since the system state transition relies on   all agents' actions. Multi-agent RL methods with distributed training are still under   research and development.

\begin{remark}
\normalfont  Although the RL techniques above are discussed separately, they can be integrated for a single problem to achieve all the benefits. For instance, one may apply the multi-agent actor-critic framework with deterministic policies,  adopt ANNs to parameterize  the $Q$-function and the policy, and use the advantage function for actor update. Accordingly, the resultant algorithm is usually named 
after the combination of key words, e.g., deep deterministic policy gradient (DDPG), asynchronous advantaged actor-critic (A3C), etc.
\end{remark}

\section{Selective Applications of RL in Power Systems} \label{sec:application}

\begin{figure}
    \centering
       \includegraphics[width = \columnwidth]{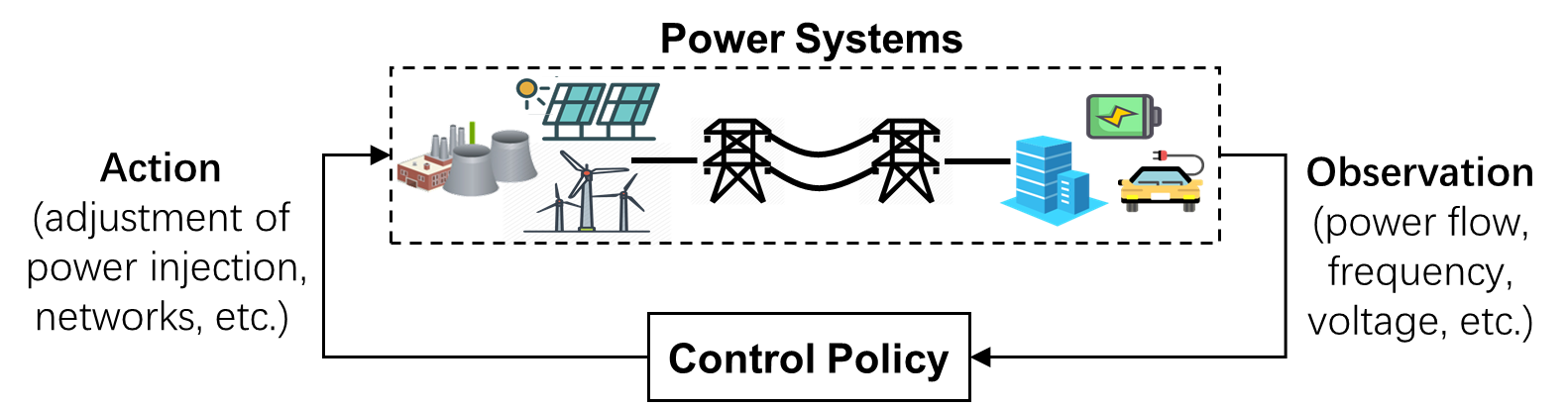}
    \caption{RL schemes for the control and decision-making in power systems.}
    \label{fig:RLpower}
\end{figure}

Throughout  the past decades,  tremendous efforts have been devoted to improving the modeling of power systems. Schemes 
 based on
(optimal) power flow techniques and precise modeling of various electric facilities are standard for the
control and optimization of   power systems. 
However, the large-scale integration of renewable generation and DERs 
significantly aggravates 
 the complexity, uncertainty, and volatility. 
   It becomes increasingly arduous
  to obtain accurate system models and power injection predictions, challenging the traditional model-based approaches. 
    Hence, model-free RL-based methods  become an appealing complement. As illustrated in Fig.  \ref{fig:RLpower}, the RL-based schemes relieve the need for accurate system models and learn control policies based on  data  collected from  actual system operation or high-fidelity simulators, when the underlying physical models and dynamics are regarded as the unknown environment.

For power systems, frequency level and voltage profile are two of the most critical indicators of system operating conditions, and reliable and efficient energy management is a core task.  
 Accordingly,  this section focuses on three key applications,
 i.e., \emph{frequency regulation}, \emph{voltage control}, and \emph{energy management}.
Frequency regulation is a fast-timescale control problem with system frequency dynamics, while energy management is usually a slow-timescale decision-making problem.  Voltage control has both fast-timescale   and slow-timescale  controllable devices. RL is a general method that is applicable to both control problems (in fast timescale) and  sequential decision-making problems (in slow timescale) under the MDP framework.  In the following, we elaborate on the overall procedure of applying RL  
to these key applications from a tutorial perspective. We summarize the related literature with a table (see Tables  \ref{tab:frequency}, \ref{tab:voltage}, \ref{tab:energy}) and use existing works to exemplify how to 
model power system applications as  RL problems. The critical  issues, future directions, and numerical implementation of RL schemes are also discussed.

 Before proceeding, a natural question  is why it is necessary to develop new RL-based approaches since traditional tools and existing controllers mostly work ``just fine" in real-world power systems. 
The answer varies from application to application, and we explain some of the main motivations below.

\begin{itemize}
    \item [1)] Although traditional methods work well in the current grid, it is  envisioned that these methods may not be sufficient for the future grid with high renewable penetration and human user participation.
    Most existing  schemes  rely heavily on  sound knowledge of power system models, and  have been challenged by various emerging issues, such as 
    the lack of accurate distribution grid models,
    highly uncertain renewable generation and  user behavior,
     coordination among massive distributed devices, the growing deployment of EVs  coupled with transportation,  etc.
    
    \item [2)] The research community has been studying various techniques to tackle these challenges, e.g., adaptive control, stochastic optimization, machine learning, zeroth-order methods, etc.  Among them, 
     RL is  a promising direction to  investigate and will play an important role in addressing these challenges because of its data-driven and model-free nature. RL is capable of dealing with highly complex and hard-to-model  problems and can adapt to rapid power fluctuations and topology changes.  %For example, RL can be used to learn human behavior, especially for demand response and EV charging, and facilitates  system operation decisions.

    \item [3)] This paper does not  suggest a dichotomy between RL and conventional methods. Instead,   RL 
    can complement existing approaches and improve them in a data-driven way. 
For instance, policy-based RL algorithms can be integrated
into existing controllers 
to online adjust key parameters  for adaptivity and achieve hard-to-model objectives.
It is necessary to identify  the  right  application  scenarios  for RL  and 
use RL schemes appropriately. This paper aims  to throw light and stimulate such discussions and relevant research. 
\end{itemize}

\subsection{Frequency Regulation}

Frequency regulation (FR) is to maintain  the power system frequency closely around its nominal value, e.g.,  60 Hz in the U.S., through balancing power generation and load demand. Conventionally,  three generation control mechanisms 
 in a hierarchical structure  are implemented at different timescales to achieve  fast response and economic efficiency.  In bulk power systems,\footnote{There are some other types of primary FR, e.g., using autonomous centralized control mechanisms, in small-scale power grids, such as microgrids.} the 
 primary FR generally operates locally to eliminate  power imbalance at the timescale of a few seconds, e.g., using \emph{droop control},
 when the governor adjusts the mechanical power input to the generator around a setpoint and based on the local frequency deviation.
 The secondary FR, known as \emph{automatic generation control} (AGC), adjusts the  setpoints of governors to bring the frequency and tie-line power interchanges  back to their nominal values, which is performed  in a centralized manner within  minutes. The tertiary FR, namely \emph{economic dispatch},  reschedules the unit commitment and restores the secondary control reserves within tens of minutes to hours. See \cite{bevrani_intelligent_nodate} for  detailed explanations of the three-level FR architecture.  In this subsection, we focus on the primary and secondary FR mechanisms, as  the tertiary FR does not involve frequency dynamics and is corresponding to the power dispatch that will be discussed  in Section \ref{sec:energymanage}.

There are a number of recent  works \cite{cui2020reinforcement,yan_multi-agent_2020,li_deep_2020,khooban_novel_2020,younesi_assessing_2020,chen_model_2020,abouheaf_load_2019,yan_data-driven_2019,wang_multiobjective_2019,adibi2019reinforcement,xi_smart_2018,yin_artificial_2017,singh_distributed_2017,yu_multi-agent_2015,rozada_load_2020} leveraging model-free RL techniques for  FR mechanism design, which are summarized in Table \ref{tab:frequency}.
 The main motivations for developing RL-based FR schemes are explained as follows.
\begin{itemize}
    \item [1)] Although the bulk transmission systems have relatively good models of  power grids, there may not be  accurate models or predictions on the large-scale renewable generation due to the inherent uncertainty and intermittency. 
    As the penetration  of renewable generation  keeps growing, 
    new challenges are posed to traditional FR 
    schemes for maintaining  nominal frequency in real-time.
   % which  will bring challenges to maintaining power balance and the nominal frequency  in the near future.
    \item [2)] FR  in the distribution level, e.g., DER-based FR, load-side FR, etc., has   attracted a great deal of recent studies. However, distribution grids may not have
 accurate system model information, 
    and it is too complex to model vast heterogeneous DERs and load devices. In such situations, RL methods     can be adopted to
      circumvent the requirement of  system model information and  directly learn control policies  from available data.
    
    \item [3)]  
    With less inertia and fast power fluctuations introduced by   inverter-based renewable energy  resources,  power systems become more and more dynamical and volatile. Conventional frequency controllers  may not adapt well to the time-varying operational environment
   \cite{yin_artificial_2017}. In addition, existing methods have difficulty  coordinating  large-scale systems at a fast time scale due to the communication and computation burdens, limiting the overall frequency regulation
performance \cite{yan_multi-agent_2020}.    Hence, (multi-agent) DRL methods can be used to develop FR schemes to improve the adaptivity and optimality.
\end{itemize}

 In the following,  we take multi-area AGC as the paradigm  to illustrate how to apply RL methods, as the mathematical models of AGC have been well established and widely used in the literature \cite{bevrani_intelligent_nodate,chen2020distributed,7944568}. We will present the  definitions of environment, state and action, the reward design, and the learning of  control policies, and then  discuss several key issues in RL-based FR. Note that the models presented  below are  examples for illustration, and there are other RL formulations and models for FR depending on the specific problem  setting.

\begin{table*}
 \caption{Literature Summary on Learning for Frequency Regulation.}
\label{tab:frequency}
%\begin{tabularx}{\textwidth} {@{}l*{6}{C}} 
\begin{tabularx}{\textwidth}{S S S  M M L}
%\begin{tabularx}{\textwidth}{@{}l*{8}{C}c@{}}
\toprule
Reference & Problem  & State/Action Space&  Algorithm   & Policy Class &   \multicolumn{1}{c}{Key Features} 
\\ 
\midrule
Yan et al. 2020 \cite{yan_multi-agent_2020}  & Multi-area AGC    &   Continuous  %discrete-time sampling for RL implementation.
& Multi-agent DDPG & ANN & 1-Offline centralized learning and  decentralized application; 2-Multiply an auto-correlated noise to the actor for exploration; 3-An initialization process is used   for ANN  training acceleration. \\ 
Li et al. 2020 \cite{li_deep_2020} & Single-area AGC & Continuous  & Twin delayed DDPG (actor-critic) & ANN & The twin delayed DDPG method is used to improve the exploration process with multiple explorers. \\
 Khooban et al. 2020 \cite{khooban_novel_2020}  &  Microgrid FR  &  Continuous  &   DDPG (actor-critic) & ANN & 1-DDPG method works as a supplementary controller for a PID-based main controller to improve the online adaptive performance;  2-Add Ornstein-Uhlenbeck process based noises to the actor  for exploration.      \\ 
 Younesi et al. 2020 \cite{younesi_assessing_2020}  & Microgrid FR    & Discrete
& $Q$-learning & $\epsilon$-greedy  &$Q$-learning  works as a supervisory controller for   PID controllers to improve the online dynamic response. \\ 
Chen et al. 2020 \cite{chen_model_2020}  & Emergency FR & Discrete &  Single/multi-agent $Q$-learning/DDPG &  $\epsilon$-greedy/ greedy/ANN & 1-Offline learning and online application;  2-The $Q$-learning/DDPG-based controller is used for limited/multiple emergency scenarios.  \\ 
Abouheaf et al. 2019 \cite{abouheaf_load_2019}  & Multi-area AGC & Continuous &   Integral RL (actor-critic)  & Linear feedback controller &  Continuous-time integral-Bellman optimality equation is considered.        \\ 
Wang et al. 2019 \cite{wang_multiobjective_2019}  & Optimization of activation rules  in AGC & Discrete & Multi-objective RL ($Q$-learning)  & Greedy &   A constrained optimization model is built to solve for optimal participation factors, where the objective is  the combination of multiple $Q$-functions.  \\
Singh et al. 2017  \cite{singh_distributed_2017}  &   Multi-area AGC&  Discrete & $Q$-learning & Stochastic policy &  An estimator agent is defined to estimate the frequency bias factor $\beta_i$ and determine the ACE signal accordingly.\\
\bottomrule
\end{tabularx}
\end{table*}

%\slow{Perhaps left align, instead of center align, the text in the last column of Table I?}

% In the follows, we take the multi-area AGC as example to elaborate how RL is applied to this problem.

%They are mainly  categorized  into two levels: multi-area level and district grid level.  The former studies the coordinated AGC over multiple control areas, while the latter focuses on the frequency control in a district power grid, such as a distribution network or a microgrid. To better present the related work, we  provide a general mathematical model on the FC problem. 

\subsubsection{Environment, State and Action}

The frequency dynamics in a power network can be expressed as  (\ref{eq:sysdyn}):
\begin{align} \label{eq:sysdyn}
 \frac{d \bs}{d t} = \bm{f}( \bs, \Delta \bP_M,  \Delta \bP_L), 
\end{align}
where  $ \bs\!:=\! ( (\Delta \omega_i)_{i\in\N}, ( \Delta P_{ij})_{ij\in\mathcal{E}})$ denotes the system \emph{state}, including the frequency deviation $\Delta \omega_i$ at each bus $i$ and the power flow deviation   $\Delta P_{ij}$ over line $ij\in\mathcal{E}$ from bus $i$ to bus $j$ (away from the nominal values). 
 $\Delta \bP_M \!:=\!(\Delta P^M_i)_{i\in \mathcal{N}}$, $\Delta \bP_L\!:=\!(\Delta P^L_i)_{i\in \mathcal{N}}$ capture the 
  deviations of generator mechanical power and other  power injections, respectively.

The governor-turbine control model \cite{bevrani_intelligent_nodate} of a  generator can be formulated as the time differential equation (\ref{eq:gendyn}):
\begin{align} \label{eq:gendyn}
 \qquad  \frac{d \Delta P_i^M}{dt} = g_i(\Delta P_i^M, \Delta \omega_i, P_i^C ), \quad i\in\mathcal{N},
\end{align}
where $P_i^C$ is the generation control command. A widely used linearized version of (\ref{eq:sysdyn}) and (\ref{eq:gendyn}) is provided in Appendix \ref{app:fredyn}. However, the real-world frequency dynamics 
(\ref{eq:sysdyn}) and  generation control model (\ref{eq:gendyn}) are  highly nonlinear and complex.  This motivates the use of model-free RL methods, since the underlying physical models (\ref{eq:sysdyn}) and (\ref{eq:gendyn}), together with operational constraints, are simply treated as the \emph{environment} in the RL setting. %In most references, high-fidelity simulators are  built to simulate the environment for offline training or testing the proposed algorithms. 

%The frequency dynamics function $\bm{f}(\cdot)$ and the generation control function $g_i(\cdot)$ are generally nonlinear and complex. See \cite{bevrani_intelligent_nodate,bergen_power} for a detailed introduction, and see \cite{singh_distributed_2017} for widely-used simplified linear models.  
 
 %The practical  frequency dynamics model (\ref{eq:sysdyn}) and the generation control model (\ref{eq:gendyn}) are  highly nonlinear and complex, especially when some operating constraints, such as generation rate constraints and generation dead bands, are considered.   This is one major motivation to develop model-free RL-based FC methods. Because in the RL setting,  the system frequency dynamics  and generation control dynamics  together with the operating constraints are treated as the \emph{environment}, which is  not necessary to be known. The FC controller (or policy) can be directly designed based on the real data collected from the system operation. Nevertheless, approximate system models are often  established to serve as a simulator  to help with the learning process.

 %They are mainly  categorized  into two levels: multi-area level and district grid level.  The former studies the coordinated AGC over multiple control areas, while the latter focuses on the frequency control in a district power grid, such as a distribution network or a microgrid. To better present the related work, we  provide a general mathematical model on the FC problem. 
 
When controlling generators for FR, the \emph{action} is defined as the concatenation of   the  generation control commands with  $\bm{a}:=(P^C_i)_{i\in\mathcal{N}}$. The corresponding action space is continuous in nature but could get discretized  in $Q$-learning-based FR schemes \cite{singh_distributed_2017,chen_model_2020}. Besides,
the continuous-time system dynamics  are generally discretized with the discrete-time horizon $\mathcal{T}$ to fit the RL framework, and the time interval $\Delta t$ depends on the sampling or control period.

We denote
$\Delta \bP_L$ in \eqref{eq:sysdyn} as the deviations of other power injections, such as
loads (negative power injection), the outputs of renewable energy resources,  the charging/discharging power of energy storage systems, etc. Depending on the actual problem setting, $\Delta \bP_L$ can be treated as exogenous states with additional dynamics, or be included in the action $\bm{a}$ if these power injections are also controlled for FR \cite{younesi_assessing_2020, khooban_novel_2020}.

%controllable loads, electric vehicles, energy storage, wind turbines, etc., are considered for FR \cite{younesi_assessing_2020, khooban_novel_2020}, they can also be included in the action $\bm{a}$. 

 \subsubsection{Reward Design} The design of the reward function plays a crucial role in successful RL applications. However, there is no general rule to follow, but one principle is to effectively reflect the control goal. 
 For multi-area AGC\footnote{For multi-area AGC problem,  each control area is generally aggregated and characterised by a single governor-turbine model (\ref{eq:gendyn}). Then the  control actions for an
individual generator within this area are allocated based on its participation factor. Thus each bus $i$  represents an aggregate control area, and $\Delta P_{ij}$ is the deviation of tie-line power interchange  from area $i$ to area $j$.}, it aims  to  restore the frequency and 
 tie-line power flow to the nominal values  after  disturbances. 
  Accordingly,  the \emph{reward} at time $t\in \mathcal{T}$ can be  defined as the minus of frequency deviation and tie-line flow deviation, e.g., in
  the square sum form (\ref{eq:frerew})  \cite{yan_multi-agent_2020}:
\begin{align} \label{eq:frerew}
        r(t) = -\Delta t \cdot \sum_{i\in\mathcal{N}} \Big(  (\beta_i \Delta \omega_{i}(t))^2 + (\sum_{j:ij\in\mathcal{E}} \Delta P_{ij}(t))^2  \Big),
\end{align}
where $\beta_i$ is the frequency bias factor. 
 For single-area FR, the goal is  to restore the system frequency, thus the term related to tie-line power flow can be removed from (\ref{eq:frerew}).
Besides, the exponential function \cite{rozada_load_2020}, absolute value function \cite{li_deep_2020}, and other
  sophisticated reward functions involving the cost of generation change and penalty for large frequency deviation \cite{li_deep_2020}, can also be used.

 \subsubsection{Policy Learning} Since the system states may not be fully observable in practice, the RL control policy is generally defined as the map $\bm{a}(t) = \pi(\bm{o}(t))$ from the available measurement observations $\bm{o}(t)$ to the action $\bm{a}(t)$. The following two steps are critical to 
 learn
 a good control policy.

 $\bullet$ \emph{Select Effective Observations.} The selection of observations typically faces a trade-off between informativeness and complexity. It is helpful to leverage domain knowledge to choose effective observations. For example,
 multi-area AGC conventionally operates  based on the area control error (ACE) signal,
given by
$\text{ACE}_i = \beta_i \Delta \omega_i  + \sum_{j:ij\in\mathcal{E}}\Delta P_{ij}$. Accordingly, the   proportional, integral, and derivative (PID) counterparts of the ACE signal, i.e., 
$(\text{ACE}_i(t), \int \text{ACE}_i(t)\, dt, \frac{d \text{ACE}_i(t)}{d t})$, are adopted as the observation in \cite{yan_multi-agent_2020}. Other measurements, such as the power injection deviations $\Delta P_i^M, \Delta P_i^L $,
could also be included in the observation \cite{singh_distributed_2017, li_deep_2020}. Reference \cite{yan_data-driven_2019} applies the stacked denoising autoencoders  to extract compact and useful  features  from the raw measurement data for FR.

  $\bullet$ \emph{Select RL Algorithm}. Both valued-based and policy-based RL algorithms 
have been applied  to FR in power systems. 
 In $Q$-learning-based FR schemes, e.g.,   \cite{singh_distributed_2017},  the state and action spaces are discretized and the $\epsilon$-greedy policy is used. Recent works  \cite{yan_multi-agent_2020, khooban_novel_2020} employ the DDPG-based actor-critic framework to develop the FR schemes, considering continuous action and observation. In addition, 
multi-agent RL  is applied to coordinate 
 multiple control areas or multiple generators in \cite{yan_multi-agent_2020, xi_smart_2018, singh_distributed_2017}, where each agent designs  its own control policy $a_i(t) = \pi_i(o_i(t))$ with the local observation $o_i$. In this way,  
the resultant algorithms can achieve centralized learning and decentralized implementation.

\subsubsection{Simulation Results} 
Reference \cite{yin_artificial_2017} conducts simulations on an interconnected power system with four provincial control areas, and demonstrates that the proposed emotional  RL algorithm outperforms SARSA, Q-learning and PI controllers, with much smaller frequency deviations and ACE values in all test cases.
In \cite{yan_multi-agent_2020}, the numerical tests on the New England 39-bus system show that the  multi-agent DDPG-based controller improves the mean absolute control error by 60.5\% over the DQN-based controller and  50.5\% over a fine-tuned PID controller.  In \cite{li_deep_2020}, the simulations on a provincial power grid with ten generator units show that the proposed DCR-TD3 
method   achieves the frequency deviation of $6.35\times 10^{-3}$ Hz and ACE of $26.48$ MW, which outperforms 
the DDPG-based controller (with   frequency deviation of $6.48\times 10^{-3}$ Hz and  ACE of $26.97$ MW)  and PI controller (with   frequency deviation of $15.54\times 10^{-3}$ Hz and  ACE of $60.9$ MW).

\subsubsection{Discussion}
Based on  the existing works listed above, we discuss
several key observations  as   follows.

 $\bullet$ \emph{Environment Model}. Most of the  references build environment models or simulators to simulate the dynamics and responses of power systems for  training and testing their proposed algorithms. These simulators are typically high fidelity with realistic component models, which are too complex to be useful for the direct development and optimization of controllers. Moreover, 
it is laborious and costly to build and  maintain such
  environment models in practice, 
 and thus they
 may not be available for many power grids. When such simulators are unavailable, a potential solution is to train off-policy RL schemes using  real system operation data.

 $\bullet$ \emph{Safety}. Since FR is  vital for power system operation, it necessitates \textit{safe} control policies. Specifically, two requirements need to be met: 1) the closed-loop system dynamics are stable when applying the RL control policies; 2) the  physical constraints, such as line thermal limits, are satisfied.
However, few existing studies consider the safety issue of applying RL to FR. A recent work
\cite{cui2020reinforcement} proposes to explicitly engineer the ANN structure of DRL  to 
 guarantee the frequency stability.

 $\bullet$ \emph{Integration with Existing Controllers}.  References
 \cite{younesi_assessing_2020,khooban_novel_2020} use the DRL-based controller as a supervisory or supplementary controller to existing  PID-based FR controllers, to improve the dynamical adaptivity with baseline performance guarantee. More discussions are provided in Section \ref{sec:future}.

  $\bullet$ \emph{Load-Side Frequency Regulation}. The  researches  mentioned above focus on controlling generators for FR. Besides, various  emerging power devices, e.g., inverter-based PV units, ubiquitous controllable loads with fast response, are  promising  complements to generation-frequency control \cite{7944568,chen2020distributed}. These are potential FR applications of RL in   smart grids.
 
 %, combines the existing controllers with the RL based control schemes, where the latter is used as the      alternative or supplementary controller to the existing traditional controller in a way that improves their fast dynamic response. 

%For multi-area AGC,  each control area is generally aggregated and characterised by a single governor-turbine model (\ref{eq:gendyn}). Then  the  control actions for an individual generator within this area are allocated based on its participation factor\footnote{The participation factor is normally fixed during the AGC cycle but is adjusted in the economic dispatch process. % Reference \cite{wang_multiobjective_2019} proposes a multi-objective RL based method to determine the optimal AGC activation rules, i.e. how to optimally allocate the real-time AGC signal to each generator. }.Accordingly, each bus $i$  represents an aggregate control area, and $\Delta P_{ij}$ is the deviation of tie-line power interchange  from area $i$ to area $j$. After disturbances, it aims to drive the frequency deviation and power interchange deviation to zero by generation control. Thus the  generation control command $\bm{u}\!:=\!(u_i)_{i\in\mathcal{N}}$ is  the \emph{action} to take, which is continuous in nature but could get discretized  in Q-learning based FC methods \cite{singh_distributed_2017}.

%As for solution, reference \cite{yan_multi-agent_2020} derives an approximate analytic formulation of the gradient of Q function with respect to the action based on the  linearized system dynamics. 

\subsection{Voltage Control}

Voltage control aims to keep the voltage magnitudes across the power networks close to the nominal values or stay within an acceptable interval. 
Most  recent works focus on   voltage control in distribution systems and propose
a variety of control mechanisms \cite{8636257,7874216, qu2019optimal,magnusson2020distributed,8859389}. 
As the penetration of renewable generation, especially solar panels and wind turbines, deepens in distribution systems,
the rapid fluctuations and significant uncertainties of renewable  generation pose new challenges to the voltage control task.  
Meanwhile, unbalanced power flow, multi-phase device integration, and the lack of accurate network models further complicate the situation. To this end, 
 a number of studies \cite{shi2021stability, lee2021graph, 9356806, 9358213,9353702,9328796,9143169,mukherjee2021scalable,kou2020safe,8985179,vlachogiannis_reinforcement_2004,xu_optimal_2020,yang_two-timescale_2020,wang_data-driven_2020,wang_safe_2020,duan_deep-reinforcement-learning-based_2020,cao_multi-agent_2020,liu2020two} propose  using model-free RL for voltage control. We summarize the related work in Table~\ref{tab:voltage} and present below how to solve the voltage control problem in the RL framework.

 \begin{table*}
 \caption{Literature Summary on Learning for Voltage Control. }
\label{tab:voltage}
%\begin{tabularx}{\textwidth} {@{}l*{6}{C}} 
\begin{tabularx}{\textwidth}{S M S S S L}
%\begin{tabularx}{\textwidth}{@{}l*{8}{C}c@{}}
\toprule
Reference   & Control Scheme & State/Action Space& Algorithm & Policy Class &  \multicolumn{1}{c}{Key Features} 
\\ \midrule
Shi et. al. 2021 \cite{shi2021stability} & Reactive power control & Continuous  &  DDPG &ANN    & Through monotone policy
network design, a Lyapunov function-based Stable-DDPG method  is proposed for voltage control with stability guarantee. \\
Lee et. al. 2021 \cite{lee2021graph} &Voltage regulators, capacitors,  batteries &  Hybrid & Proximal policy
optimization   &   Graph convolutional network  & A graph neural networks (topology)-based RL method
 is proposed and extensive simulations are performed to 
study the properties of graph-based policies.   \\
Liu et. al. 2021 \cite{9356806} & SVCs and PV units &Continuous   & Multi-agent
constrained soft actor-critic   &  ANN  & 
Multiple agents are
trained in a centralized manner to learn the coordination control
strategies and are executed in a decentralized manner  based on local information.

\\
 Yin et. al. 2021 \cite{9358213} &Automatic voltage control  & Discrete &  Q-learning&   ANN & The
emotional factors are added to the ANN structure and Q-learning  to improve  the accuracy and performance of the
control algorithm.\\
Gao et. al. 2021 \cite{9353702} &  Voltage regulator, capacitor banks, OLTC &  Hybrid/ discrete  &  Consensus multi-agent DRL&  ANN  &  1- The maximum entropy method is used to encourage exploration; 2- A consensus multi-agent RL algorithm is developed, which enables distributed control and efficient communication.\\
 Sun et. al. 2021 \cite{9328796} & PV reactive power control & Hybrid/ continuous   &   Multi-agent DDPG &  ANN   & A voltage sensitivity based DDPG method is proposed, which analytically computes the gradient of value function to action rather than using the critic ANN. \\
Zhang et. al. 2021 \cite{9143169} & Smart inverters, voltage regulators, and capacitors & Continuous/ discrete  &   Multi-agent DQN &  $\epsilon$-greedy   & 1- Both the network loss and voltage violation are considered in the reward definition; 2- Multi-agent DQN is used to enhance the scalability of the algorithm. \\
Mukherjee et. al. 2021 \cite{mukherjee2021scalable} &  Load shedding  &  Continuous   &  Hierarchical DRL&    LSTM  &   1- A hierarchical multi-agent RL algorithm with two levels  is developed to accelerate learning; 2- Augmented random search  is used to solve for optimal policies. \\
 Kou et. al. 2020 \cite{kou2020safe} &  Reactive power control &  Continuous &  DDPG &  ANN &  A safety layer is formed on top of the actor network  to ensure safe exploration, which predicts the state change and prevents the violation of constraints.  \\
   Al-Saffar et. al. 2020 \cite{8985179} &   Battery energy storage systems  &  Discrete   & Monte-Carlo tree search   &    Greedy &  The proposed approach divides a network
into multiple smaller segments based on impacted regions; and  it solves the over-voltage problem via  Monte-Carlo tree search  and model predictive control.  \\
Xu et. al. 2020 \cite{xu_optimal_2020} & Set OLTC tap positions &Hybrid/ discrete  & LSPI (batch RL) & %off & 
Greedy & 1- ``Virtual'' sample generation is used for better exploration; 2- Adopt a multi-agent trick to handle scalability and use Radial basis function as state features.  \\
Yang et. al. 2020 \cite{yang_two-timescale_2020} & On-off switch of capacitors   & Hybrid/ discrete & DQN 
& Greedy & Power injections are determined via traditional OPF in fast timescale, and the  switching  of capacitors is determined via DQN in slow timescale. \\
Wang et. al. 2020 \cite{wang_data-driven_2020} & Generator voltage setpoints&  Continuous& Multi-agent DDPG
& ANN & Adopt a competitive (game) formulation with specially designed reward for each agent. \\
Wang et. al. 2020 \cite{wang_safe_2020} & Set tap/on-off positions& Hybrid/ Discrete  & Constrained SAC %& Off??? 
& ANN & 1- Model the voltage violation  as constraint using the constrained MDP framework; 2- Reward is defined as the negative of power loss and switching cost.  \\
Duan et. al. 2020 \cite{duan_deep-reinforcement-learning-based_2020} & Generator voltage setpoints &Hybrid & DQN/DDPG %& On 
& Decaying $\epsilon$-greedy/ANN & 1- DQN is used for discrete action and DDPG is used for continuous action; 2- Decaying $\epsilon$-greedy policy is employed in DQN to encourage exploration. \\
Cao et. al. 2020 \cite{cao_multi-agent_2020} & PV reactive power control& Continuous & Multi-agent DDPG% & on 
& ANN & The attention neural network is used to develop the critic to enhance the algorithm scalability. \\
Liu et. al. 2020 \cite{liu2020two} & Reactive power control&Continuous & Adversarial RL \cite{pinto2017robust}/SAC
& Stochastic policy& 1- A two-stage RL method is proposed to improve the online safety and efficiency via offline pre-training; 2-  Adversarial SAC is used to make the online application robust to the transfer gap.\\
%Vlachogiannis et. al. 2004 \cite{vlachogiannis_reinforcement_2004} & & Discrete & $Q$-learning %& off?  & Greedy & ? \\
\bottomrule
\end{tabularx}
\end{table*}

%The goal of voltage control in distribution grids is to keep the voltage magnitudes across the grid close to the nominal value (e.g. 120V at the household level in the U.S.). Several mechanisms can be used for this purpose, including the adjustment of reactive power output of Distributed Energy Resources (DERs) like solar panels, and changing capacity bank/load tap set points \cite{Baran1989a,Baran1989b,li2014real, zhang2013local, bolognani2013distributed, zhu2016fast,kekatos2015fast,cavraro2016value,liu2018hybrid,tang2019fast,qu2019optimal,magnusson2019voltage,magnusson2020distributed}. The model that describes the relationship between the control mechanisms and the voltage magnitudes can be very complex, and in the absence of accurate models,  a number of work has proposed using model-free RL for voltage control \cite{vlachogiannis_reinforcement_2004,xu_optimal_2020,yang_two-timescale_2020,wang_data-driven_2020,wang_safe_2020,duan_deep-reinforcement-learning-based_2020,cao_multi-agent_2020}.We summarize the related work in Table~\ref{tab:voltage} and present below how the voltage control problem can be solved in the RL framework.

\subsubsection{Environment, State and Action} 
%Typically, there is a distribution grid consisting of $\mathcal{N} = \{0,1,\dots,n\}$ of buses and a set $\mathcal{E}\in \mathcal{N}\times \mathcal{N}$ of edges. 

 In distribution systems, the controllable devices  for voltage control can be classified into  slow  timescale  and fast timescale. Slow  timescale devices, such as on-load tap changing  transformers (OLTCs), voltage regulators, and capacitor banks, are   discretely controlled on an hourly or daily basis. The \emph{states} and control \emph{actions} for them can be defined as 
\begin{subequations}
\begin{align} \label{eq:volsta}
     \bm{s}_{\mathrm{slow}}  & := \big( (v_i)_{i\in\mathcal{N}}, (P_{ij},Q_{ij})_{ij\in\mathcal{E}},\boldsymbol{\tau}^{\mathrm{TC}},\boldsymbol{\tau}^{\mathrm{VR}},  \boldsymbol{\tau}^{\mathrm{CB}}\big),\\
     \bm{a}_{\mathrm{slow}} &:=  \big(\Delta\boldsymbol{\tau}^{\mathrm{TC}}, \Delta\boldsymbol{\tau}^{\mathrm{VR}}, \Delta \boldsymbol{\tau}^{\mathrm{CB}}\big),
 \end{align}
\end{subequations}
where $v_i$ is the voltage magnitude of bus $i$, %$p_i, q_i$ are the active/reactive power injections at bus $i$, 
and $P_{ij}, Q_{ij}$ are the active and reactive power flows over line $ij$. $\boldsymbol{\tau}^{\mathrm{TC}}$, $\boldsymbol{\tau}^{\mathrm{VR}}$,  $\boldsymbol{\tau}^{\mathrm{CB}}$ denote  the tap positions of the OLTCs, voltage regulators, and capacitor banks respectively, which are discrete values. $\Delta\boldsymbol{\tau}^{\mathrm{TC}}, \Delta\boldsymbol{\tau}^{\mathrm{VR}}, \Delta \boldsymbol{\tau}^{\mathrm{CB}}$ denote the  discrete  changes of corresponding tap positions.

The fast timescale devices include inverter-based  DERs and  static Var compensators (SVCs), whose (active/reactive) power outputs\footnote{Due to
 the comparable magnitudes of line resistance and reactance in  distribution networks, the conditions for active-reactive power decoupling are no longer met. Thus \emph{active power} outputs also play a  role in voltage control, and alternating current  (AC) power flow models are generally needed.
} can be continuously controlled within seconds. Their \emph{states} and control \emph{actions} can be defined as 
\begin{subequations}
\begin{align} \label{eq:volsta:fast}
     \bm{s}_{\mathrm{fast}}  & := \big( (v_i)_{i\in\mathcal{N}}, (P_{ij},Q_{ij})_{ij\in\mathcal{E}}\big),\\
     \bm{a}_{\mathrm{fast}} &:=  \big(\bm{p}^{\mathrm{DER}}, \bm{q}^{\mathrm{DER}},\bm{q}^{\mathrm{SVC}}\big),
 \end{align}
\end{subequations}
where $\bm{p}^{\mathrm{DER}}, \bm{q}^{\mathrm{DER}}$ collect the continuous active and reactive power outputs of  DERs respectively, and $\bm{q}^{\mathrm{SVC}}$ denotes the reactive power outputs of SVCs.

%Note that the state (\ref{eq:volsta}) and the action (\ref{eq:volact}) defined above  contain both continuous variables, e.g., $v_i, \bm{p}^{\mathrm{DER}}$, and discrete variables, e.g., $\boldsymbol{\tau}^{\mathrm{CB}},\Delta \boldsymbol{\tau}^{\mathrm{CB}}$. 
Since RL methods  handle continuous and discrete actions  differently, most existing studies only consider either continuous control actions (e.g., $\bm{q}^{\mathrm{DER}}$) \cite{wang_data-driven_2020,cao_multi-agent_2020}, or discrete control actions (e.g., $\boldsymbol{\tau}^{\mathrm{CB}}$ and/or $\boldsymbol{\tau}^{\mathrm{TC}} $) \cite{vlachogiannis_reinforcement_2004,xu_optimal_2020}. Nevertheless, the recent works \cite{yang_two-timescale_2020,liu2021bi}
 propose  two-timescale or bi-level RL-based voltage control algorithms, taking {into} account both fast continuous devices and slow discrete devices. {In the rest of this subsection},  we uniformly  use $\bm{s}$ and $\bm{a}$ to denote the state and action.

Given the definitions of state and action, the system dynamics that depict the \emph{environment} can be formulated as 
\begin{align} \label{eq:voldyn}
    \bm{s}(t+1) = \bm{f}\big(\bm{s}(t),\bm{a}(t), \bm{p}^{\mathrm{ex}}(t),\bm{q}^{\mathrm{ex}}(t)\big),
\end{align}
where  $\bm{p}^{\mathrm{ex}},\bm{q}^{\mathrm{ex}}$ denote the exogenous active power and reactive power injections to the grid, including load demands and other  generations. The transition function $\bm{f}$ captures the tap position evolution and the power flow equations, 
which could be very complex or even unknown in reality.  The exogenous injections $\bm{p}^{\mathrm{ex}},\bm{q}^{\mathrm{ex}}$ include the uncontrollable renewable generations that are difficult to  predict    well. These issues motivate the RL-based voltage control schemes as 
 the dynamics model (\ref{eq:voldyn}) is not required in the RL setting.

\subsubsection{Reward Design} The goal of voltage control is to maintain the voltage magnitudes  close to the nominal value (denoted as $1$ per unit (p.u.)) or   within an acceptable range.
Accordingly,
 the reward function is typically in the form  of penalization on the voltage deviation from $1$ p.u. For example, in \cite{xu_optimal_2020,yang_two-timescale_2020,cao_multi-agent_2020}, the reward is defined as (\ref{eq:volrew}), 
 \begin{align}\label{eq:volrew}
     r(\bm{s},\bm{a}) = - \sum_{i\in\mathcal{N}} (v_i - 1)^2.
 \end{align}  
% where the negative sign indicates the smaller the voltage violation, the higher the reward. 
An alternative   is to set the reward to be negative (e.g., $-1$) when the voltage is outside an acceptable range (e.g., $\pm 5\%$ of the nominal value), and positive (e.g., $+1$) when inside the range  \cite{duan_deep-reinforcement-learning-based_2020}.
Moreover, the reward can incorporate the operation cost of   controllable devices (e.g., switching cost of discrete devices),  power loss \cite{wang_safe_2020}, and other sophisticated metrics \cite{wang_data-driven_2020}.

\subsubsection{RL Algorithms} Both value-based and policy-based RL algorithms have been applied for voltage control:

%As mentioned in section~\ref{subsec:classicrl}, there are mainly two types of RL methods: value-based and policy-based. 

 $\bullet$ \emph{Value-Based RL.} Several works \cite{vlachogiannis_reinforcement_2004,xu_optimal_2020,yang_two-timescale_2020,duan_deep-reinforcement-learning-based_2020} adopt value-based algorithms, such as DQN and LSPI, to learn the optimal $Q$-function with function approximation, typically using ANNs \cite{yang_two-timescale_2020,duan_deep-reinforcement-learning-based_2020} or radial basis functions \cite{xu_optimal_2020}. 
Based on the learned $Q$-function, these works use the greedy policy as the control policy, i.e., $\bm{a}(t) = \arg\max_{\tilde{\bm{a}}\in\mathcal{A}} Q(\bm{s}(t),\tilde{\bm{a}})$.
Two limitations of the greedy policy include 1) the action selection depends on the state of the entire system, which hinders  distributed implementation; 2) it is usually not suitable for continuous action space since the maximization may be difficult to compute, especially when complex function approximation, e.g., with ANNs, is adopted.

 $\bullet$ \emph{Policy-Based RL.} Compared with value-based RL methods, the voltage control schemes based on  actor-critic algorithms, e.g., \cite{wang_data-driven_2020,wang_safe_2020,duan_deep-reinforcement-learning-based_2020,cao_multi-agent_2020},
are more flexible, which can  accommodate both continuous and discrete actions and enable distributed implementation. Typically, a parameterized deterministic policy class $q_i^{\mathrm{DER}} = \pi_{i,\bm{\theta}_i}(\bm{o}_i^{\mathrm{DER}})$ is employed for 
each DER device  $i$, which determines the reactive power output $q_i^{\mathrm{DER}}$ based on local observation $\bm{o}_i^{\mathrm{DER}}$ with parameter $\bm{\theta}_i$. The policy class $\pi_{i,\bm{\theta}_i}$ is often parameterized using ANNs. 
Then some actor-critic methods, e.g., multi-agent DDPG, are used to optimize  parameter $\bm{\theta}_i$, where a centralized critic
learns the $Q$-function with ANN approximation and each   DER device  performs the policy gradient update  as the actor.

\subsubsection{Simulation Results} In \cite{9143169},  the simulations on the IEEE 123-bus system show that 
the  proposed  multi-agent DQN method converges to a stable reward level after about 4000 episodes during the training process, and  it achieves an average power loss reduction of 75.23 kW compared with a baseline method. In \cite{wang_safe_2020}, the tests on a 4-bus feeder system show that the  constrained SAC  and DQN methods take about $1\times 10^4$ training
samples to achieve stable performance, while the constrained policy optimization (CPO) method  requires
up to $5\times 10^5$ training samples to converge. Besides, the   constrained SAC achieves the highest return and almost zero voltage violations  for the 34-bus
and 123-bus test feeders.
In \cite{xu_optimal_2020}, it takes about 30 seconds for the proposed LSPI-based algorithm to converge in the case of IEEE 13-bus test feeder, which is faster than the exhaustive search approach
by several orders of magnitude but maintains a similar level of reward.

\subsubsection{Discussion}
Some key issues of applying RL to voltage control are discussed below:

$\bullet$ \emph{Scalability}. As the network scale and the number of controllable devices increase, the  size of the state/action space {grows} exponentially, which poses severe challenges in learning the $Q$-function. 
Reference \cite{xu_optimal_2020} proposes a useful trick that defines different $Q$-functions for different actions, which 
leads to a scalable method under its special problem formulation. 

$\bullet$  \emph{Data Availability}.
To learn the $Q$-function for a given policy,
on-policy RL methods, such as actor-critic, need to implement the policy and collect sample data. This could be problematic since the policy is not necessarily safe, and thus the implementation on  real-world power systems may be catastrophic. One remedy is to train the policy on high-fidelity simulators. Reference 
 \cite{xu_optimal_2020} proposes a novel method to generate virtual sample data  for a certain policy, based on the data collected from implementing another safe policy. More discussions on safety are provided in Section \ref{sec:safety}.

 $\bullet$ \emph{Topology Change}. 
The network topology, primarily for distribution systems, is  subject to changes from time to time due to network reconfiguration, line faults, and other operational factors. A voltage control policy trained for a specific topology may not work well under a different network topology. To cope with this issue,  the network topology can be included as one of the states or a parameter of the policy. 
Reference \cite{lee2021graph} represents  the power network topology   with graph neural networks and studies the properties of 
graph-based voltage control policies.
Besides,  if the set of network topologies is not large, one can train a separate control policy offline for each possible topology and apply the corresponding policy online.  To avoid learning from scratch and enhance efficiency, one may use
transfer RL \cite{parisotto2015actor}   to transplant the well-trained policy for a given topology to another.

\subsection{Energy Management} \label{sec:energymanage}

Energy management is an advanced  application that utilizes information flow to manage power flow and maintain power balance
in a reliable and efficient manner. To this end, 
energy management systems (EMSs) are developed for electric power control centers  to monitor, control, and optimize the system operation.  
 With the assistance of the
supervisory control and data acquisition (SCADA) system, 
the EMS for transmission systems is technically mature.
However, for many sub-regional power systems, such as medium/low-voltage distribution grids and microgrids,  EMS is still under development due to the integration of various DER facilities and the lack of 
 metering units. 
Moreover,  an EMS family \cite{sun2012family} with a hierarchical structure is necessitated to facilitate different levels of energy management, including
grid-level EMS, EMS for coordinating a cluster of DERs, home EMS (HEMS),  etc.

In practice, there are  significant uncertainties in energy management,
which result from 
 unknown models and parameters of power networks and DER facilities, uncertain user  behaviors and weather conditions, etc. Hence,  many recent studies    
 adopt (D)RL techniques to develop data-driven EMS. 
  A summary of the related literature is provided in Table \ref{tab:energy}.  
 In the rest of this subsection, we first introduce the RL models of  DERs and adjustable loads, then review the RL-based schemes for different levels of energy management problems.

%The following models only consider active power for simplicity, while the reactive power can be similarly included.
% \subsection{Models of DERs and Adjustable loads}

\subsubsection{State, Action, and Environment} We present the action, state and environment models for several typical DER facilities, buildings, and residential loads.

$\bullet$ \emph{Distributed Energy Resources:}
For compact expression, we consider a bundle of several typical DERs, including   a   dispatchable PV unit,  a battery, an EV,  and a diesel generator (DG). The \emph{action} at time $t\in \mathcal{T}$ is defined as 
\begin{align} \label{eq:der:a}
    \ba^{\mathrm{DER}}(t):= \big(p^{\mathrm{PV}}(t),\, p^{\mathrm{Bat}}(t),\, p^{\mathrm{EV}}(t),\,  p^{\mathrm{DG}}(t)\big).
\end{align}
Here, $p^{\mathrm{PV}}, p^{\mathrm{Bat}}, p^{\mathrm{EV}}, p^{\mathrm{DG}}$ are the power outputs of the PV unit, battery, EV, and DG respectively, which are continuous. $p^{\mathrm{Bat}},p^{\mathrm{EV}}$ can be 
either positive (discharging) or negative (charging). 
The DER \emph{state} at time $t$ can be defined as
\begin{align} \label{eq:der}
\bs^{\mathrm{DER}}(t):= \big(\bar{p}^{\mathrm{PV}}(t),\, {E}^{\mathrm{Bat}}(t),\, {E}^{\mathrm{EV}}(t),\, \bm{x}^{\mathrm{EV}}(t) \big),
\end{align}
where  $\bar{p}^{\mathrm{PV}}$ is the maximal PV  generation power
determined by the solar irradiance.  The PV output $p^{\mathrm{PV}}$ can be adjusted within the interval $[0, \bar{p}^{\mathrm{PV}}]$, and  $p^{\mathrm{PV}}(t) = \bar{p}^{\mathrm{PV}}(t)$ when the PV unit operates in the maximum power point tracking (MPPT)  mode.
${E}^{\mathrm{Bat}},{E}^{\mathrm{EV}}$ denote the associated state of charge (SOC) levels. 
$\bm{x}^{\mathrm{EV}}$ captures other related states of the EV, e.g. current  location (at home or outside), travel plan, etc.

$\bullet$ \emph{Building HVAC:} Buildings account for a large share of the total energy usage, about half of which is consumed by the heating, ventilation, and air conditioning (HVAC) systems \cite{buildweb}.
Smartly scheduling  HVAC  operation has huge potential to save energy cost, but the building climate dynamics are intrinsically hard to model and affected by various environmental factors. 
Generally, a building is divided into multiple  thermal zones,  and the \emph{action} at time $t$ is defined as 
\begin{align} 
    \ba^{\mathrm{HVAC}} (t) := \big( T_{c}(t),\,T_{s}(t), \, (m^{i}(t))_{i\in\mathcal{N}}  \big),
\end{align}
where $T_c$ and $T_s$ are the conditioned air temperature and the supply air temperature, respectively. $m^{i}$ is the supply air flow rate at zone $i\in\mathcal{N}$. 
The choice of states is subtle, since many exogenous factors may affect the indoor climate. 
A typical definition of the \emph{HVAC state} is 
\begin{align}  \label{eq:hvacstate}
    \bs^{\mathrm{HVAC}} (t):= \big(  T_{out}(t), \big(T^{i}_{in}(t), h^i(t), e^i(t)\big)_{i\in\mathcal{N} }  \big),
\end{align}
where $T_{out}$ and $T^i_{in}$ are the outside temperature and  indoor temperature of zone $i$; $h^i$ and $e^i$ are the humidity  and  occupancy rate of zone $i$, respectively. 
Besides, the solar irradiance, the carbon dioxide concentration and other environmental factors may also be included in the state $\bs^{\mathrm{HVAC}}$.

$\bullet$ \emph{Residential Loads:} 
Residential demand response (DR) \cite{demandresene} that motivates changes in electric consumption by end-users in response to time-varying electricity price or incentive payments  attracts considerable recent attention. 
The domestic electric appliances are classified as 1) \emph{non-adjustable loads}, e.g., computers and refrigerators,  which are critical  and must be satisfied; 2) \emph{adjustable loads}, e.g., air conditioners and washing machines, whose  operating power or usage time can be adjusted.  The \emph{action} for an adjustable load $i$ at time $t\in\mathcal{T}$ can be defined as 
 \begin{align}
  \quad    \ba_i^{\mathrm{L}}(t):= \big(z_{i}^{\mathrm{L}} (t),\, p_{i}^{\mathrm{L}} (t)\big),\ i\in\mathcal{N}_L,
 \end{align}
 where binary $z_{i}^{\mathrm{L}}\in \{0,1\}$ denotes  whether switching the on/off working mode (equal to $1$) or keeping unchanged (equal to $0$). $p_{i}^{\mathrm{L}}$ is the power consumption of load $i$, which can be adjusted either discretely or continuously depending on the load characteristics. 
The operational \emph{state} of  load $i$  can be defined as 
\begin{align}  \label{eq:loadstate}
    \quad     {\bs}_{i}^{\mathrm{L}} (t) :=  (\alpha_{i}^{\mathrm{L}} (t),  \bm{x}_{i}^{\mathrm{L}}(t)),\ i\in\mathcal{N}_L,
\end{align}
where binary $\alpha_{i}^{\mathrm{L}}$ equals $0$ for the off status and $1$ for the on status. 
$\bm{x}_{i}^{\mathrm{L}}$ collects other related states of load $i$. 
 For example,   the indoor and outdoor temperatures are contained in $\bm{x}_{i}^{\mathrm{L}}$ if load $i$ is an air conditioner \cite{chen2020online}; and $\bm{x}_{i}^{\mathrm{L}}$ captures the task progress and the remaining time to the  deadline for a washing machine load.

$\bullet$ \emph{Other System States:} 
 In addition to the operational states above, there are some critical \emph{system states} for EMS, e.g.,
 \begin{align} \label{eq:sys:s}
     \bs^{\mathrm{Sys}}(t): = \big(t,\, \bm{\ell}(t-K_p:t+K_f), \bm{v}(t),\bm{P}(t),\cdots    \big),
 \end{align}
 including the current time $t$, electricity price $\bm{\ell}$ (from past $K_p$ time steps to future $K_f$ time predictions)  \cite{lu2019demand}, voltage profile $\bm{v}:=(v_i)_{i\in\mathcal{N}}$,  power flow  $\bm{P}:=(P_{ij})_{ij\in\mathcal{E}}$, etc.

  The state definitions  in (\ref{eq:der}), (\ref{eq:hvacstate}), and (\ref{eq:loadstate}) only contain the present status at time $t$, which can also include the past values and  future predictions to capture the temporal patterns. In addition, the previous actions may also be considered as one of the states, e.g., adding $\ba^{\mathrm{DER}}(t\!-\!1)$ to the state $\bs^{\mathrm{DER}}(t)$.  For  different energy management problems, 
their state $\bm{s}$  and action $\bm{a}$ are determined accordingly by selecting and combining the definitions in (but not limited to) \eqref{eq:der:a}-\eqref{eq:sys:s}. And the \emph{environment} model is given by
\begin{align}
    \bs(t+1) = \bm{f}(\bs(t), \ba(t), \bm{u}^{\mathrm{ex}}(t)),
\end{align}
where $\bm{u}^{\mathrm{ex}}$ captures other related exogenous factors. We note that the RL models presented above are   examples for illustrative purposes, and one needs to build its own models to fit   specific applications.

\subsubsection{Energy Management Applications}

Energy management indeed covers a broad range of sub-topics, including integrated energy systems (IESs),  grid-level power dispatch, management of DERs, building HVAC control, and HEMS, etc.
We present these sub-topics in a hierarchical order below,
and summarize the basic attributes and key features of representative references in Table \ref{tab:energy}.

$\bullet$ \emph{Integrated Energy Systems} \cite{huang2020multienergy}, also referred to as multi-energy systems, incorporate  the power grids with heat networks and gas networks   to accommodate renewable energy and enhance the overall energy efficiency and flexibility \cite{9361643}. Reference \cite{9016168} proposes  a DDPG-based real-time  control strategy to manage a residential multi-energy system, where   DERs, heat pumps, gas boilers, and thermal 
energy storage are controlled to minimize the total operational cost. 
In \cite{9346692}, the management of IESs with integrated demand response is modeled as a Stackelberg game, and an  actor-critic scheme is developed for the energy provider to adjust pricing and power dispatching strategies  to cope with unknown private parameters of users. Extensive  
case studies are conducted 
in  \cite{ceusters2021model} to compare the performance of a  twin delayed DDPG scheme against a benchmark linear model-predictive-control method, which empirically show that RL is a viable optimal control technique for IES management and can outperform conventional approaches.

$\bullet$ \emph{Grid-Level Power Dispatch} aims to schedule the power outputs of generators and DERs to optimize the operating cost of the entire grid while satisfying the operational constraints.
Optimal power flow (OPF) is a fundamental tool of traditional power dispatch schemes. Several recent
works \cite{9069289,8853513,9272783,9275611,9465777} propose DRL-based methods to solve the OPF problem in order to achieve fast solution and cope with the absence of accurate grid models. 
Most existing references  \cite{8957677,8789677,9464628,9245590,8769895,8306311} focus on the power dispatch  in  distribution grids or microgrids. In \cite{9245590}, a model-based DRL algorithm is developed to online schedule a residential microgrid, and Monte-Carlo tree search is adopted to find the optimal decisions.
Reference \cite{8306311} proposes a cooperative RL algorithm  for distributed economic dispatch in microgrids, where a diffusion strategy is used to coordinate the actions of many DERs.

$\bullet$ \emph{Device-Level Energy Management} focuses on the optimal control of DER devices and adjustable loads, such as EV, energy storage system, HVAC, and residential electric appliances, which usually aims to minimize the total energy cost under time-varying electricity price.
In \cite{wan_modelfree_realtime_2019,li_constrained_ev_2020,achiam_constrained_policy_2017}, various RL techniques are studied to design EV charging policies to deal with the  randomness in the arrival and departure time of an EV. 
See \cite{9371688} for a  review on RL-based EV charging management systems. 
References  \cite{bui_double_deep_2020, 9061038, 9097915} adopt  DQN and DDPG to learn the charging/discharging strategy for controlling battery systems considering unknown degradation models. In terms of building HVAC control, there are multiple uncertainty factors such as random zone occupancy, unknown thermal dynamics models, uncertain outdoor temperature and electricity price, etc. Moreover,
 the thermal comfort and air quality need to be guaranteed. 
To this end, a number of studies \cite{wei2017deep, 9085925,9146920,mocanu2018line,9161266} leverage DRL  for HVAC system control. 
 In  \cite{xu2020multi,lu2019demand,li_real-time_2020,alfaverh2020demand}, DRL-based HEMS is developed to optimally schedule household electric appliances, considering
   resident's preferences, uncertain electricity price and weather conditions.

 \subsubsection{Simulation Results} In \cite{liu2020optimization}, the offline training  of DQN over 1000 episodes takes about 2 hours, and the simulations show that the double DQN-based HEMS can reduce the  user's  daily electricity payment by about $50\%$.  In \cite{yu2019deep}, the   DDPG-based EMS  method achieves a  relatively stable reward after 3000 episodes in the training process, and reduces the total energy cost by $15.2\%$ and $8.1\%$ compared with two  baseline algorithms.
 In the case studies of \cite{wan_modelfree_realtime_2019},  the training of ANN 
 converges after about 35000 epochs;  the DQN-based EV charging   scheme
 decreases the charging cost by $77.3\%$ in comparison with the uncontrolled solution and performs better than other benchmark solutions. Reference \cite{ye_modelfree_realtime_2020} compares the training efficiency of DQN, DPG, and prioritized DDPG, which take about 8000, 13000, and 6200 episodes to reach the benchmark performance, respectively.

\subsubsection{Discussion}
Some key issues are discussed as follows.

  $\bullet$ \emph{Challenges in Controlling DERs and Loads}. Large-scale distributed renewable generation introduces significant uncertainty and intermittency to   energy management, which requires highly accurate  forecasting techniques and fast adaptive controllers to cope. The partial observability issue of complex facilities  and the heterogeneity of various devices lead to further difficulties in coordinating massive loads. Moreover, the  control of HVACs and residential loads involves interaction with human users; thus it is necessary   to take into account user comfort and learn unknown and diverse user behaviors.

 $\bullet$ \emph{Physical Constraints}. There are various physical constraints, e.g., the state of charge limits for batteries and EVs, that should be satisfied when taking control actions. 
Reference \cite{8306311} formulates the constraint violation as a penalty term in the reward,   in the form of a logarithmic barrier function. Reference \cite{li_constrained_ev_2020} builds a constrained MDP problem to take into account the physical constraints  and solves the problem with the constrained policy optimization method \cite{achiam_constrained_policy_2017}.  These methods
 impose the constraints in a ``soft" manner, and there is still a chance to
 violate the constraints. More discussions are provided in Section \ref{sec:safety}.

 $\bullet$ \emph{Hybrid of Discrete and Continuous State/Action.} Energy management often involves the control of a hybrid of discrete devices and continuous devices, yet the basic RL methods only focus on handling either discrete or continuous actions.
Some $Q$-learning-based  work \cite{xu2020multi}  discretizes the continuous action space to fit the algorithm framework.
Reference  \cite{li_real-time_2020} proposes an ANN-based stochastic policy to handle both discrete and continuous actions, combining a Bernoulli  policy for on/off switch actions and a Gaussian policy for continuous actions.

\begin{table*}
 \caption{Literature Summary on Learning for Energy Management.}
\label{tab:energy}
\begin{tabularx}{\textwidth}{S S S  M M L}
\toprule
Reference & Problem  & State/Action Space&  Algorithm   & Policy Class & \multicolumn{1}{c}{Key Features}  
\\ 
\midrule
Ye et al. 2020 \cite{9016168} & IES management   & Hybrid & DDPG (actor-critic)    &  Gaussian policy & The prioritized experience replay method is used to enhance the sampling efficiency of the experience replay mechanism. \\
Xu et al. 2021 \cite{9361643} &   IES management  &  Discrete & $Q$-learning    &  Stochastic policy &   RL-based differential evolution algorithms are developed to solve the complex IES scheduling issue. %problem, where RL is used to find the optimal mutation strategy and associated parameters.
\\ 
 Yan et al. 2020 \cite{9069289}  &   Optimal power flow  &  Continuous &  DDPG   & ANN &   Rather than using the critic network, the deterministic gradient of a Lagrangian-based  value function is derived
analytically. \\ 
 Woo et al. 2020  \cite{9272783}&   Optimal power flow  &  Continuous &   Twin delayed DDPG   &  ANN &   The appropriate reward vector in
the training process is set to build decision policies, considering power system constraints. \\ 
Zhou et al. 2020  \cite{9275611}&   Optimal power flow  &   Continuous  &   Proximal  policy  optimization    & Gaussian policy &  Imitation learning is adopted to generate initial weights for ANNs and  proximal policy optimization is used to train a DRL agent for fast OPF solution. \\ 
Hao et al. 2021  \cite{9464628}&   Microgrid power dispatch  &  Continuous   &  Hierarchical RL     &  Two knowledge
rule-based policies&  1- Hierarchical RL is used to reduce complexity and improve learning efficiency; 2-
Incorporated with domain knowledge, it avoids baseline
violation and additional learning beyond feasible action space.
\\ 
Lin et al. 2020 \cite{8957677} & 
Power dispatch &
Continuous & Soft A3C & Gaussian  policy
& The edge computing technique is employed to accelerate the computation and communication in a cloud environment. \\
Zhang et al. 2020 \cite{8789677} & Distribution power dispatch   &Continuous
& Fitted $Q$-iteration    &  $\epsilon$-greedy & The  $Q$-function is parameterized by polynomial approximation and optimized using a regularized recursive least square method with a forgetting factor.\\ 
%Foruzan et al. 2018 \cite{foruzan2018reinforcement} & Microgrid power dispatch & Discrete & $Q$-leaning & Greedy &\color{red}  Multi-agent and distributed energy management, formulated as a Markov game, not finished \\
%Kim et al. 2016 \cite{kim2015dynamic}  &    Pricing scheme & Discrete & $Q$-learning  & Greedy  & The post-decision state learning  is used to remove the need of exploration and improve the learning speed. \\

Wan et al. 2019 \cite{wan_modelfree_realtime_2019}  & EV charging scheduling &  Continuous/ discrete
& Double DQN  & $\epsilon$-greedy & A representation network is constructed to extract features from the electricity price. \\ 
Li et al. 2020 \cite{li_constrained_ev_2020} & 
EV charging scheduling &
Continuous
& Constrained policy optimization \cite{achiam_constrained_policy_2017} & Gaussian   policy & A constrained MDP is formulated to schedule the charging of an EV, considering  charging constraints.  \\
Silva et al. 2020 \cite{silva2020coordination} &
EV charging scheduling & Discrete
& Multi-agent $Q$-learning & $\epsilon$-greedy & Use multi-agent multi-objective RL to mode the EV charging coordination  with the $W$-learning method. \\
%Sadeg. et al. 2020 \cite{8727484} & EV charging scheduling & Discrete  & Fitted $Q$-iteration &  Greedy& Study the coordination of a set of EV charging stations.\\
Bui et al. 2020 \cite{bui_double_deep_2020} & Battery management &
Hybrid/ discrete & Double DQN & $\epsilon$-greedy & 
To mitigate the overestimation problem, double DQN  with a primary network for action selection and a target network is used. \\
%Zhang et al. 2016 \cite{zhang2016optimal} & HEMS    &  & &   &  \\
%Wen et al. 2015 \cite{wen2015optimal} & HEMS    &  discrete  & Q-learning& greedy  & Focus on designing a EMS framework on job rescheduling  and interacting with users. \\
%Mieth et al. 2019 \cite{mieth2019online} & Pricing Scheme    &  continuous & least square estimation  & distributionally robust optimization  & \\
%Li et al. 2019 \cite{li2019distributed}  & Pricing Scheme    &  continuous & Iterative linear regression  & robust optimization & \\
%Mocanu et al. 2018 \cite{mocanu2018line} & Building EM & Continuous/ discrete & Deep policy gradient & DNN \\
Cao et al. 2020  \cite{9061038}& Battery management  &  Hybrid/ discrete  &  Double DQN   &  Greedy &  A hybrid CNN  and LSTM model is adopted to predict the future electricity
price.
\\ 
Yu et al. 2021  \cite{9146920}& Building HVAC & Continuous  &   Multi-actor-attention-critic    & Stochastic Policy &   A scalable HVAC control algorithm is proposed to solve
the Markov game based on multi-agent DRL with attention mechanism. 
\\ 
 Gao et al. 2020  \cite{9085925}& Building HVAC & Continuous   &    DDPG &  ANN&  A feed-forward ANN with Bayesian regularization is built for
predicting thermal comfort. 
\\ 
 Mocanu et al. 2019   \cite{mocanu2018line} & Building  energy management&  Hybrid   &   DPG and DQN  &  ANN&  Both DPG and DQN are implemented for building energy control and their performances are compared numerically.
\\ 
Xu et al. 2020 \cite{xu2020multi}  & HEMS    & Continuous/ discrete 
&Multi-agent $Q$-learning& $\epsilon$-greedy  & Use extreme learning machine based ANNs to predict future PV output and electricity price. \\ 
Li et al. 2020 \cite{li_real-time_2020}  & HEMS    & Hybrid
& Trust region policy optimization   &  ANNs-based  stochastic policy & A policy network  determines both discrete actions (on/off switch with Bernoulli  policy) and continuous actions (power control with Gaussian policy).  \\ 
 Alfaverh et al. 2020 \cite{alfaverh2020demand}  & HEMS    & Discrete
& $Q$-learning&  Stochastic policy &  Fuzzy sets and fuzzy reasoning rules are used to simplify the action-state space.
\\
 Chen et al. 2021 \cite{chen2020online} & Residential load control    &  Continuous & Thompson sampling  & Stochastic policy  & Logistic regression is employed to predict customers' opt-out behaviors in demand response and Thompson sampling is used for online learning. \\
\bottomrule
\end{tabularx}
\end{table*}

\subsection{Other Applications}

In addition to the three critical applications above, other applications of RL in  power systems include   electricity market \cite{8826539,8805177},  %multi-energy system management \cite{9016168},
network reconfiguration \cite{gao2020batch}, service restoration \cite{8862818,5871714},  emergency control \cite{8787888}, maximum power point tracking \cite{7081385,9079637}, cyber security \cite{8786811,8248780}, maintenance scheduling \cite{8998302},   protective relay control \cite{9029268}, electric vehicle charging navigation \cite{8845652}, demand response customer selection \cite{online2021mab}, power flexibility aggregation \cite{li2020real},  etc.

%{\color{red}add more existing applications...}

\subsection{Numerical Implementation}\label{sec:numerical}

In this subsection, we present the overall procedure, useful tools, and available testbeds for the numerical implementation of (D)RL in power system applications. The implementation procedure generally comprises the following three steps:

\subsubsection{Environment Setup}
First, an environment simulator  specifying states,   actions, observations, and internal transition needs to be built to simulate the real system and   applications. 
The  \emph{OpenAI Gym}  \cite{brockman2016openai} is a prominent  toolkit that provides many simulation environments of physical systems, games, and robots for RL research.
As for power system applications, most existing works build their own  synthetic environments to train and test RL algorithms based on standard IEEE test systems or  real power grids.  In    \cite{henry2021gym}, 
a framework called \emph{Gym-ANM} is developed
to establish RL 
environments for active network management tasks in  distribution systems.
Besides, the \emph{Gird2Op} framework\footnote{Grid2Op package [Online]: \url{https://github.com/rte-france/Grid2Op}.} is an open-source environment for training RL agents to operate power networks, which is the testbed for the Learning to Run a Power Network (L2RPN) challenge \cite{marot2021learning}. Other recently developed RL environments  include \emph{RLGC} \cite{8787888} for power system control, \emph{gymgrid}  \cite{henri2020pymgrid} and  \emph{OMG} \cite{heid2020omg} for microgrid simulation and control, and \emph{PowerGym} \cite{fan2021powergym} for voltage control in distribution systems, etc.
A variety of test  systems and test cases are available in these environments.

\subsubsection{Agent Setup} Then, one needs to create an agent (or   agents) that specifies the reward function, RL methods, and policies to interact with the environment, i.e., 
receiving observations and taking control actions. For the implementation of DRL with ANNs, the   mainstream
 open-source deep learning frameworks include \emph{TensorFlow} \cite{abadi2016tensorflow}, \emph{PyTorch} \cite{paszke2019pytorch}, \emph{Keras}\footnote{Keras library [Online]: \url{https://keras.io/}.}, \emph{MXNet}\footnote{Apache MXNet library [Online]: \url{mxnet.apache.org}.}, \emph{CNTK}\cite{seide2016cntk}, etc. Building on top of these deep learning frameworks, there are several widely-used open-source RL libraries, such as 
\emph{Tensorforce}\footnote{Tensorforce library [Online]: \url{https://github.com/tensorforce/tensorforce}.}, \emph{Stable Baselines}\footnote{Stable Baselines [Online]: \url{https://stable-baselines.readthedocs.io/}.},  \emph{RL Coach}, \emph{KerasRL}, \emph{TF Agents}, etc.,  which basically cover the implementation of all state-of-the-art RL algorithms.
%See \cite{rlframework} for a summary and comparison of RL frameworks. 

 \subsubsection{Agent Training and Testing} With the environment and the agent in place, the embedded functions in the RL frameworks  introduced above can be   directly used to train and test the agent, or researchers can code their own implementation with tailored designs. Besides, there are some commercial toolboxes available for RL implementation. For examples, the MathWorks RL toolbox \cite{matrltool} can be used to build and train agents under the  
 environments modeled in Matlab or Simulink.

\section{Challenges and Perspectives} \label{sec:challenge}

This section presents the critical challenges of using RL in  power system applications, i.e., safety and robustness, scalability,  and data. Several   future directions are then discussed.

\subsection{Safety and Robustness} \label{sec:safety}
Power systems are vital infrastructures of modern societies. Hence,  it is necessary to ensure that the applied controllers   are safe, in the sense that they do not drive the power system operational states to violate critical physical constraints, or cause instability or reliability issues. Regarding RL-based control schemes, 
there are  two aspects of safety concern:

\emph{1) Guarantee that the learning process is safe} (also referred to as \emph{safe exploration}). For this issue,  off-policy RL methods \cite{xu_optimal_2020} are more desirable, where the training data are generated  from existing controllers that are known to be safe. In contrast,
 it remains  an open question for on-policy RL to guarantee safe exploration. Some attempts \cite{berkenkamp2017safe, koller2018learning, chow2018lyapunov, fan2019safety} propose safe on-policy exploration schemes based on  Lyapunov criterion and Gaussian process. The basic idea is to construct a certain safety region, and special actions are taken to drive the state back once approaching the boundary of this safety region.
 See \cite{garcia2015comprehensive} for a comprehensive survey on safe RL. However,
 almost all the existing works  train their RL control policies only based on high-fidelity power system simulators, 
and it is plausible that  the safe exploration problem is circumvented. 
Nevertheless, one can argue that there might be a substantial gap between the simulator and the real-world system, leading to the failure of generalization in real implementation. 
A possible remedy is  to employ the robust (adversarial) RL methods \cite{morimoto2005robust,pinto2017robust, liu2020two} in simulator-based policy training.

\emph{2) Guarantee that the final learned control policy is safe}. It  is generally hard to verify whether a policy is safe or the generated actions can respect  physical operational constraints. Some
common methods to 
deal with  constraints include 1) formulating the constraint violation as a penalty term  to the reward, 2) training the control policy based  on a constrained MDP  \cite{wang_safe_2020,li_constrained_ev_2020}, 3) adding a heuristic safety layer to adjust the  actions such that the constraints are respected.
Specifically, the second method aims to learn an optimal policy $\pi^*$  that maximizes the expected total return $J(\pi)$ and is subject to a budget constraint:
   \begin{align} \label{eq:cmdp}
    \pi^* \in \arg\max_{\pi} J(\pi),\quad \text{s.t.} \ J^c(\pi)\leq d, 
\end{align}
where $J^c(\pi)$ is the expected total cost and $d$ is the budget. By defining  the physical constraint violation as certain costs in
$J^c(\pi)$, (\ref{eq:cmdp}) imposes safety requirements to some degrees. Typical approaches to solve the constrained MDP problem  (\ref{eq:cmdp}) include the Lagrangian methods \cite{wang_safe_2020,tessler2018reward},  constrained policy update rules \cite{achiam_constrained_policy_2017}, etc.

We briefly introduce three related RL variants below, i.e., constrained RL, adversarial RL, and robust RL. 
 
$\bullet$ \emph{Constrained RL}  deals with the safety issue and constraints. Two types of constraints, i.e., soft  and hard constraints, are generally considered in  the literature. The common ways to handle soft constraints include 
 1) using barrier functions or penalty functions to integrate the constraints to the reward function \cite{cheng2019end};  2) modeling as a chance constraint (i.e., the probability of constraint violation is no larger than a predefined threshold)
  \cite{qin2021density,wabersich2018linear} or a budget constraint (such as the constraint in model \eqref{eq:cmdp}) \cite{altman1999constrained,achiam2017constrained}. In terms of hard constraints, the predominant approach is to take  conservative actions to ensure that the hard constraints   are satisfied at all times,  despite the problem uncertainties \cite{dean2019safely}.
  However, such schemes usually lead to   conservativeness and may not work well when the underlying system is complex.

$\bullet$
\emph{Adversarial RL} \cite{pinto2017robust,pattanaik2017robust} adopts a two-player game structure
where the learner agent learns to take actions against an adversarial agent whose objective is different from or even opposite to the learner agent.
When applying adversarial RL to the  power system cyber security problem \cite{9494982,8944283,pan2021improving}, one can model the   cyber attacker as the adversarial agent, whose 
 attack actions, attack schemes, and payoffs  depend  on the practical settings.
% date, there are  few studies \cite{9494982,8944283,pan2021improving} on the application of adversarial RL to  the cyber security of power systems.
Reference \cite{8944283} formulates  a repeated game
 to mimic the  interactions between the
attackers and defenders in  power systems. Reference  \cite{pan2021improving} proposes  an agent-specific  adversary MDP to learn an
adversarial policy and  uses it to improve the robustness of
RL methods via adversarial training.

$\bullet$ \emph{Robust RL} \cite{morimoto2005robust,mankowitz2019robust} employs a  $\min$-$\max$ framework to learn robust control policies,
where  ``$\min$" corresponds to the learner and  ``$\max$" corresponds to the adversary. The adversary is generally designed to choose uncertain parameters (e.g. future renewable generation) from an  uncertainty set or select the worst-case scenarios  from a predefined contingency  event set. Embedding the $\min$-$\max$ structure in RL algorithms 
has been an active  research area.  Early studies \cite{iyengar2005robust,nilim2005robust} focus on robust MDP with uncertain parameters. Recent work \cite{zhang2020robust} extends single-agent robust RL \cite{mankowitz2019robust,derman2018soft} to deal with parametric uncertainty in multi-agent RL.  Applying robust RL to power system applications  is an important future direction to  deal with  parametric uncertainties, data errors, and  mismatches between simulators and  real-world systems.

\subsection{Scalability} 

It is observed that
most  existing studies run simulations and tests on small-scale power systems with a few decision-making agents. 
  To the best of our knowledge, no real-world implementation of RL control schemes has been reported yet. 
  A crucial limitation for  RL in large-scale multi-agent systems, such as power systems, is the scalability issue, since the state and action spaces expand dramatically as the number of agents increases,   known as   ``curse of dimensionality".
Multi-agent RL  and  function approximation techniques are useful for improving the scalability,  while they are still under development with  many limitations. For example, 
there are limited provable guarantees on how well $Q$-function can be approximated with ANNs, and it is unclear whether it works for real-size power grids. 
Moreover,  even though each agent can deploy 
a local policy to determine its own action, 
most existing multi-agent RL methods still need centralized learning among  all the agents because the $Q$-function depends on the global state and all  agents' actions.
One potential direction 
to enable  distributed learning is to leverage  local dependency properties (e.g., the fast decaying property) to  find near-optimal localized policies \cite{qu2020scalable,lin2020distributed}.
Besides, 
some application-specific approximation methods can be utilized to design scalable RL algorithms. For instance, reference 
 \cite{xu_optimal_2020} develops a scalable LSPI-based voltage control scheme, which   sequentially  learns an approximate $Q$-function for each component of the action, when the other components are assumed to behave greedily according to their own approximate $Q$-functions.

\subsection{Data}

\subsubsection{Data Quantity, Quality, and Availability}
Evaluating  the amount of data   needed for training a good policy, namely  \emph{sample complexity}, is a challenging and active research area in the RL community.
For classical RL algorithms, such as $Q$-learning, the sample complexity depends on the size of the state and action spaces;    the larger the state and action spaces are, the more data are generally needed to find a near-optimal policy  \cite{qu2020finite,li2020sample}.
For
modern RL methods   commonly used in power systems, such as DQN and actor-critic, the sample complexity also depends on the complexity of the function class adopted to approximate the $Q$-function and the intrinsic approximation error of the function class \cite{fan2020theoretical, kumar2019sample}. 
In addition, \emph{data quality}  is one of the critical factors affecting   learning efficiency. Real  measurement and operational data of power grids suffer from various issues, such as  missing  data,  outlier data,  noisy data,  etc., thus a pre-processing on raw data is  needed. 
Theoretically,  larger variance in noisy observations typically leads to higher sample complexity for achieving a certain level of accuracy. Besides,
almost all the references   reviewed above assume 
that high-fidelity simulators or accurate environment models  are available 
to simulate the system dynamics and response, 
which are the sources of  sample data for  training and testing  RL policies.  When such simulators are unavailable,  \emph{data  availability}  becomes  an  issue  for  the  application  of  on-policy  RL  algorithms.

\subsubsection{Potential Directions to Address Data Issues} 
Despite   successful   simulation results, 
 theoretical understanding of the sample complexity of  modern RL algorithms is limited. 
In addition, many power system applications use  multi-agent training methods with partial observation and adopt ANNs for function approximation, further complicating the theoretical analysis.  
A key solution to improve the  sample complexity of training RL policies is the use of \emph{warm starts}. Empirical results \cite{silva2021encoding} validate that good initialization can significantly  enhance  training efficiency. There are 
multiple ways to achieve a warm start, such as 1) utilizing existing  controllers  for pre-training \cite{yan_multi-agent_2020,qu2020combining}, 2) encoding domain knowledge 
into the design of control policies \cite{silva2021encoding},  3) transfer learning \cite{parisotto2015actor} that transplants well-trained policies to a similar task to avoid learning from scratch,  4) imitation learning \cite{hussein2017imitation} that learns from available demonstrations or  expert systems, etc. 

In terms of data availability and quality, one can deal with them from   algorithmic and physical levels.   At the \emph{algorithmic level}, when high-fidelity simulators are unavailable, a
potential solution is to construct  training samples  from existing system operational data  and employ off-policy RL methods to learn control policies.  Other training techniques such as  generating  virtual samples from limited data to  boost the data availability \cite{xu_optimal_2020} can also be adopted.
There have been extensive studies on  data quality  improvement in the data science field,
such as data sanity check, missing data imputation, bad data identification, etc.  At the \emph{physical level},  1) deploying more advanced sensors and smart meters and 2) upgrading communication  infrastructure and technologies  can   improve   data  availability and quality
at the source.

%{\color{red} data available. Simulator or operational data}

\subsubsection{Standardized Dataset and Testbed} 
Existing works in the power literature mostly use   synthetic test systems and  datasets  to simulate and test the proposed RL-based algorithms, and they may not provide 
many implementation details and codes. Hence,
 it is necessary to develop benchmark datasets and authoritative testbeds for power system applications to standardize the testing of RL algorithms and facilitate fair performance comparison. We summarize several  available RL environments
 for power system applications in Section \ref{sec:numerical}.

 %are described in Section \ref{sec:numerical}, and more works are needed to cover various power system applications.

\subsubsection{Big Data Techniques} The big data in smart grids include  1) measurement data from  SCADA,  PMUs, AMI, and other advanced metering facilities, 2) electricity market pricing and bidding data,  3) equipment
monitoring, control, maintenance, and management
data, 4) meteorological data, etc. They
can benefit the application of data-driven RL in various ways \cite{8809368}. 
For example, big data mining techniques for  knowledge discovery can be adopted to detect special events,
determine effective observations, and identify critical latent states. Pattern extraction from massive datasets can be utilized to classify and cluster similar events, agents and user behaviors, to improve the data efficiency   of RL algorithms.

\subsection{Other Key Future Directions} \label{sec:future}

Regarding the challenges in applying RL to power systems,
we present several potential future directions as below.

\subsubsection{Integrate Model-free and Model-based Methods} Actual power system operation is not a black box and has  abundant model information to use.
Purely model-free approaches may be too radical to exploit available  information and suffer from their own limitations,  such as the safety and scalability issues discussed above. Since  existing
model-based methods have already been well studied in theory and applied in the industry with acceptable performance, one promising future direction is to \emph{combine  model-based and model-free methods}
for complementarity and achieve  both advantages.   For instance,  model-based methods can serve as  warm-starts or the nominal model,  or be adopted to  identify critical features for model-free methods. 
Besides,
model-free methods can   coordinate and adjust the parameters of incumbent model-based controllers to improve their adaptivity
with baseline performance guarantees. 
Reference \cite{wang_integrating_2019} summarizes three potential integration ways: implementing model-based and model-free methods  in serial,  in parallel,  or embedding one as an inner module in the other. Despite 
limited  work, e.g. \cite{qu2020combining,che2018combining}, 
on this subject so far, integrating model-free RL    with existing model-based control schemes is envisioned to be an important future direction.

\subsubsection{Exploit Suitable RL Variants} RL is a fundamental and vibrant research field  attracting a great deal of attention.  New advances in RL algorithms appear frequently. 
Besides  DRL, multi-agent RL, and robust RL mentioned above, 
 a wide range of branches in the RL  field, such as transfer RL \cite{parisotto2015actor}, meta RL \cite{wang2016learning}, federated RL \cite{zhuo2019federated}, inverse RL\cite{ng2000algorithms}, integral RL \cite{modares2014integral},  Bayesian RL \cite{ghavamzadeh2016bayesian}, hierarchical RL \cite{barto2003recent}, interpretable RL \cite{verma2018programmatically}, etc., can  improve the learning efficiency  and tackle specific problems in suitable  application scenarios.  For instance,   transfer RL can be used to transplant the well-trained policies for one task to another similar task, so that  it does not have to learn from scratch and thus can enhance the training efficiency.

\subsubsection{Leverage Domain Knowledge and Problem Structures} The naive application of existing RL algorithms may encounter many troubles in practice. 
In addition to algorithmic advances, 
\emph{leveraging domain knowledge and exploiting application-specific structures  to design tailored RL algorithms}  are necessary to achieve superior performance. Specifically, domain  knowledge and empirical evidence  can   guide the definition of state and reward, the initialization of the policy, and the selection of   RL algorithms.
 For example,  
 the area  control  error (ACE) signal is often used as the state in RL-based 
  frequency regulation.  Besides, the specific problem structures are useful in determining the policy class, approximation function class, hyperparameters, etc., to improve training efficiency and provide performance guarantees. For example,   reference \cite{cui2020reinforcement} leverages two special properties of the frequency regulation problem and designs the policy network in a particular structure to ensure the stability of the resultant RL controllers.

\subsubsection{Satisfy Practical Requirements} 
The following concrete requirements on RL-based methods need to be met to enable  practical implementation in power systems.

\begin{itemize}
\item As discussed above, the safety, scalability, and data issues of RL-based methods need to be addressed. 
    \item RL-based algorithms should be \emph{robust} to the noises and failures in measurement, communication, computation, and actuation, to ensure reliable operation.
    \item To be used with confidence, RL-based  methods need to be \emph{interpretable} and have theoretical performance guarantee.
\item Since RL   requires a large amount data from multi-stakeholders, the data privacy should be preserved.

    \item As power systems generally operate under normal conditions,  it remains an unsolved problem to ensure that   RL control policies learned from real system data have sufficient exploration and  perform well in extreme scenarios.
    \item Since RL-based approaches heavily rely on information flow, the 
     cyber security should be guaranteed under various malicious cyber attacks.
 \item  Existing RL-based algorithms mostly take tens of thousands of iterations to converge, which suggests that  the \emph{training efficiency} needs to be improved.
        \item Necessary computing resources, communications infrastructure and technology need to be deployed and upgraded to  support the application of RL schemes.  We elaborate on this requirement below.
\end{itemize}

%In addition, necessary computing resources, communications infrastructure and technologies need to be deployed or upgraded to support the application of RL schemes.
In  many existing works, multi-agent DRL is used to develop scalable control  algorithms with  centralized (offline) training and  decentralized (online) implementation. To enable  centralized training of DRL, the  coordination center needs large-scale data storage,  high-performance computers, and advanced computing techologies, such as accelerated computing (e.g., GPUs), cloud and edge computing, etc. As for decentralized or distributed implementation, although the computational burden is lighter, each  device (agent) typically requires  local sensors, meters, microchip-embedded solvers, and automated actuators. Moreover, to support the application of DRL, advanced
communication infrastructures  are necessary to enable the two-way 
communication and  real-time streaming of high-fidelity data from massive devices. Various 
 communication and networking technologies, such as (optic) cable lines, power line carrier,  cellular,
satellite, 5G,  WiMAX, WiFi, Xbee, Zigbee, etc., can be used for different RL applications. In short,
both algorithmic advances and  infrastructure development are envisioned to facilitate the practical application of 
RL schemes.

%One of the most promising directions  is to combine  model-based and model-free methods  for complementarity and improved performance. On the one hand, purely model-free approaches suffer from their own limitations,  such as the safety and scalability issues discussed above, and some useful and practically available  model information may be wasted. On the other hand,  existing model-based methods have already been well studied in theory and applied in industry with acceptable performance. Hence, the integration of model-based and model-free methods is envisioned to compensate the intrinsic limitations of both and achieve  the advantages of both. According to the integration architecture,  the integration manners of model-based and model-free methods can  be classified as three types \cite{wang_integrating_2019}:
%\begin{itemize}
%    \item [(1)] Model-based and model-free methods are implemented \textbf{in serial}. For example, model-based approaches work as a warm-start or are used to identify critical features for the model-free methods.  
%    \item [(2)]  Model-based and model-free methods work \textbf{in parallel}. For example, the outputs from the parallel model-based and model-free controllers are  weighted  to determine the final decisions.
%    \item [(3)] Model-based methods work as inner    modules \textbf{embedded} in model-free methods, and vice versa. For example,  available model information can be utilized to design policies with  particular structures instead of general NNs. or use RL to tune the (PID) parameters.
%\end{itemize}

\section{Conclusion} \label{sec:conclusion}
Although a number of works have been devoted to applying RL to the  power system field, many critical problems remain unsolved, and there is still a   substantial distance from practical implementation. On the one hand, this subject is new and still under development and needs much more studies.
  On the other hand, it is time to  step back and rethink the advantages and limitations of  applying RL to power systems (the world's most complex and vital engineered systems) and figure out where and when to use RL.  In fact, RL is not 
envisioned to completely replace  existing model-based methods but a viable alternative in specific tasks. For instance, RL and other data-driven methods are promising
when the models are too complex to be useful or when the problems are intrinsically hard to model, such as  the human-in-loop control (e.g., in demand response).
It is highly expected to identify the right application scenarios for RL and use it appropriately.

\appendices

\section{System Frequency Dynamics} \label{app:fredyn}

According to \cite{chen2020distributed,singh_distributed_2017,bevrani_intelligent_nodate}, 
the system frequency dynamics (\ref{eq:sysdyn}) can be linearized as  (\ref{eq:fredyn}), including the generator swing dynamics (\ref{eq:fregene}) and   power flow dynamics (\ref{eq:frepow}).
\begin{subequations} \label{eq:fredyn}
\begin{align}
 & \Delta \dot{\omega}_i =\! -\frac{1}{M_i} (D_i \Delta \omega_i - \Delta P_i^M + \Delta P_i^L +\!\!\sum_{j:ij\in \mathcal{E}}\!\Delta P_{ij} ), \, i\in \mathcal{N} \label{eq:fregene} \\
 & \Delta \dot{ P}_{ij}  = B_{ij} (\Delta \omega_i - \Delta \omega_j), \quad   ij\in \mathcal{E}, \label{eq:frepow}
\end{align}
\end{subequations}
where $M_i, D_i, B_{ij}$ denote the  generator  inertia, damping coefficient, and synchronization coefficient, respectively. Besides,
the governor-turbine control model (\ref{eq:gendyn}) for a  generator can be simplified as  (\ref{eq:dyngene}), including 
the turbine dynamics (\ref{eq:dyngene:tur}) and the governor dynamics (\ref{eq:dyngene:gov}):
\begin{subequations} \label{eq:dyngene}
\begin{align}  
  \Delta  \dot{P}_i^M & = -\frac{1}{T^{\text{tur}}_i}(\Delta P_i^M - \Delta P_i^G ), \label{eq:dyngene:tur}\\
    \Delta \dot{P}_i^G & = -\frac{1}{T^{\text{gov}}_i}( \frac{1}{R_i}\Delta \omega_i+\Delta P_i^G - P_i^C ),  \label{eq:dyngene:gov}
\end{align}
\end{subequations}
where $\Delta P_i^G$ is the turbine valve position deviation, and $P^C_i$ is the generation control command.  $T^{\text{tur}}_i, T^{\text{gov}}_i$ denote the turbine and governor time constants, and $R_i$ is the droop coefficient.

\bibliography{IEEEabrv, ref_intro,ref_RL, Learning_Frequency, Learning_PowerDispatch, Learning_Voltage_Control,VoltageBackgroundRef,Learning_DR} 

\begin{thebibliography}{100}
\providecommand{\url}[1]{#1}
\csname url@samestyle\endcsname
\providecommand{\newblock}{\relax}
\providecommand{\bibinfo}[2]{#2}
\providecommand{\BIBentrySTDinterwordspacing}{\spaceskip=0pt\relax}
\providecommand{\BIBentryALTinterwordstretchfactor}{4}
\providecommand{\BIBentryALTinterwordspacing}{\spaceskip=\fontdimen2\font plus
\BIBentryALTinterwordstretchfactor\fontdimen3\font minus
  \fontdimen4\font\relax}
\providecommand{\BIBforeignlanguage}[2]{{%
\expandafter\ifx\csname l@#1\endcsname\relax
\typeout{** WARNING: IEEEtran.bst: No hyphenation pattern has been}%
\typeout{** loaded for the language `#1'. Using the pattern for}%
\typeout{** the default language instead.}%
\else
\language=\csname l@#1\endcsname
\fi
#2}}
\providecommand{\BIBdecl}{\relax}
\BIBdecl

\bibitem{gungor2011smart}
V.~C. Gungor, D.~Sahin, T.~Kocak, S.~Ergut, C.~Buccella, C.~Cecati, and G.~P.
  Hancke, ``Smart grid technologies: Communication technologies and
  standards,'' \emph{IEEE Trans. Ind. Informat.}, vol.~7, no.~4, pp. 529--539,
  Nov. 2011.

\bibitem{sutton2018reinforcement}
R.~S. Sutton and A.~G. Barto, \emph{Reinforcement {Learning}: An
  {Introduction}}.\hskip 1em plus 0.5em minus 0.4em\relax MIT press, 2018.

\bibitem{li2017deep}
Y.~Li, ``Deep reinforcement learning: An overview,'' \emph{arXiv preprint
  arXiv:1701.07274}, 2017.

\bibitem{silver2016mastering}
D.~Silver, A.~Huang, C.~J. Maddison \emph{et~al.}, ``Mastering the game of go
  with deep neural networks and tree search,'' \emph{Nature}, vol. 529, no.
  7587, pp. 484--489, Jan. 2016.

\bibitem{kober2013reinforcement}
J.~Kober, J.~A. Bagnell, and J.~Peters, ``Reinforcement learning in robotics: A
  survey,'' \emph{The Int. J. of Robot. Res.}, vol.~32, no.~11, pp. 1238--1274,
  Aug. 2013.

\bibitem{sallab2017deep}
A.~E. Sallab, M.~Abdou, E.~Perot, and S.~Yogamani, ``Deep reinforcement
  learning framework for autonomous driving,'' \emph{Electronic Imaging}, vol.
  2017, no.~19, pp. 70--76, Jan. 2017.

\bibitem{zhao2011reinforcement}
Y.~Zhao, D.~Zeng, M.~A. Socinski, and M.~R. Kosorok, ``Reinforcement learning
  strategies for clinical trials in nonsmall cell lung cancer,''
  \emph{Biometrics}, vol.~67, no.~4, pp. 1422--1433, Dec. 2011.

\bibitem{zhang2019deep}
Z.~Zhang, D.~Zhang, and R.~C. Qiu, ``Deep reinforcement learning for power
  system applications: An overview,'' \emph{CSEE J. Power Energy Syst.},
  vol.~6, no.~1, pp. 213--225, Mar. 2020.

\bibitem{glavic2019deep}
M.~Glavic, ``{(Deep)} reinforcement learning for electric power system control
  and related problems: A short review and perspectives,'' \emph{Annual Reviews
  in Control}, vol.~48, pp. 22--35, Oct. 2019.

\bibitem{cao2020reinforcement}
D.~Cao, W.~Hu, J.~Zhao, G.~Zhang, B.~Zhang, Z.~Liu, Z.~Chen, and F.~Blaabjerg,
  ``Reinforcement learning and its applications in modern power and energy
  systems: A review,'' \emph{J. Mod. Power Syst. Clean Energy}, vol.~8, no.~6,
  pp. 1029--1042, Nov. 2020.

\bibitem{yang2020reinforcement}
T.~Yang, L.~Zhao, W.~Li, and A.~Y. Zomaya, ``Reinforcement learning in
  sustainable energy and electric systems: A survey,'' \emph{Annual Reviews in
  Control}, vol.~49, pp. 145--163, Apr. 2020.

\bibitem{bertsekas2012dynamic}
D.~Bertsekas, \emph{Dynamic Programming and Optimal Control}.\hskip 1em plus
  0.5em minus 0.4em\relax Athena Scientific Belmont, MA, 2012.

\bibitem{modares2014integral}
H.~Modares, F.~L. Lewis, and M.-B. Naghibi-Sistani, ``Integral reinforcement
  learning and experience replay for adaptive optimal control of
  partially-unknown constrained-input continuous-time systems,''
  \emph{Automatica}, vol.~50, no.~1, pp. 193--202, Jan. 2014.

\bibitem{lecarpentier2019non}
E.~Lecarpentier and E.~Rachelson, ``Non-stationary markov decision processes, a
  worst-case approach using model-based reinforcement learning, extended
  version,'' \emph{arXiv preprint arXiv:1904.10090}, 2019.

\bibitem{cheung2020reinforcement}
W.~C. Cheung, D.~Simchi-Levi, and R.~Zhu, ``Reinforcement learning for
  non-stationary markov decision processes: The blessing of (more) optimism,''
  in \emph{Proc. Int. Conf. Mach. Learn.}\hskip 1em plus 0.5em minus
  0.4em\relax PMLR, 2020, pp. 1843--1854.

\bibitem{moerland2020model}
T.~M. Moerland, J.~Broekens, and C.~M. Jonker, ``Model-based reinforcement
  learning: A survey,'' \emph{arXiv preprint arXiv:2006.16712}, 2020.

\bibitem{jaksch2010near}
T.~Jaksch, R.~Ortner, and P.~Auer, ``Near-optimal regret bounds for
  reinforcement learning.'' \emph{J. Mach. Learn. Res.}, vol.~11, no.~4, Apr.
  2010.

\bibitem{russo2017tutorial}
D.~Russo, B.~Van~Roy, A.~Kazerouni, I.~Osband, and Z.~Wen, ``A tutorial on
  thompson sampling,'' \emph{arXiv preprint arXiv:1707.02038}, 2017.

\bibitem{tsitsiklis1994asynchronous}
J.~N. Tsitsiklis, ``Asynchronous stochastic approximation and {Q}-learning,''
  \emph{Machine Learning}, vol.~16, no.~3, pp. 185--202, 1994.

\bibitem{agarwal2020optimality}
A.~Agarwal, S.~M. Kakade, J.~D. Lee, and G.~Mahajan, ``Optimality and
  approximation with policy gradient methods in markov decision processes,'' in
  \emph{Conf. Learning Theory}.\hskip 1em plus 0.5em minus 0.4em\relax PMLR,
  2020, pp. 64--66.

\bibitem{580874}
J.~N. {Tsitsiklis} and B.~{Van Roy}, ``An analysis of temporal-difference
  learning with function approximation,'' \emph{IEEE Trans. Autom. Control},
  vol.~42, no.~5, pp. 674--690, May 1997.

\bibitem{robbins1951stochastic}
H.~Robbins and S.~Monro, ``A stochastic approximation method,'' \emph{Ann.
  Math. Stat.}, pp. 400--407, 1951.

\bibitem{srikant2019finite}
R.~Srikant and L.~Ying, ``Finite-time error bounds for linear stochastic
  approximation and {TD} learning,'' \emph{arXiv preprint arXiv:1902.00923},
  2019.

\bibitem{watkins1992q}
C.~J. Watkins and P.~Dayan, ``Q-learning,'' \emph{Machine Learning}, vol.~8,
  no. 3-4, pp. 279--292, 1992.

\bibitem{qu2020finite}
G.~Qu and A.~Wierman, ``Finite-time analysis of asynchronous stochastic
  approximation and $ q $-learning,'' in \emph{Conf. Learning Theory}.\hskip
  1em plus 0.5em minus 0.4em\relax PMLR, 2020, pp. 3185--3205.

\bibitem{rummery1994line}
G.~A. Rummery and M.~Niranjan, \emph{On-line {Q}-Learning Using Connectionist
  Systems}.\hskip 1em plus 0.5em minus 0.4em\relax University of Cambridge,
  Department of Engineering Cambridge, UK, 1994.

\bibitem{degris2012off}
T.~Degris, M.~White, and R.~S. Sutton, ``Off-policy actor-critic,'' \emph{arXiv
  preprint arXiv:1205.4839}, 2012.

\bibitem{sutton2000policy}
R.~S. Sutton, D.~A. McAllester, S.~P. Singh, and Y.~Mansour, ``Policy gradient
  methods for reinforcement learning with function approximation,'' in
  \emph{Adv. Neural Inf. Process. Syst.}, 2000, pp. 1057--1063.

\bibitem{silver2014deterministic}
D.~Silver, G.~Lever, N.~Heess, T.~Degris, D.~Wierstra, and M.~Riedmiller,
  ``Deterministic policy gradient algorithms,'' in \emph{Proc. Int. Conf. Mach.
  Learn.}, 2014, pp. 387--395.

\bibitem{lange2012batch}
S.~Lange, T.~Gabel, and M.~Riedmiller, ``Batch reinforcement learning,'' in
  \emph{Reinforcement learning}.\hskip 1em plus 0.5em minus 0.4em\relax
  Springer, 2012, pp. 45--73.

\bibitem{ernst2005tree}
D.~Ernst, P.~Geurts, and L.~Wehenkel, ``Tree-based batch mode reinforcement
  learning,'' \emph{J. Mach. Learn. Res.}, vol.~6, pp. 503--556, Apr. 2005.

\bibitem{lagoudakis2003least}
M.~G. Lagoudakis and R.~Parr, ``Least-squares policy iteration,'' \emph{J.
  Mach. Learn. Res.}, vol.~4, pp. 1107--1149, Dec. 2003.

\bibitem{agarwal2020optimistic}
R.~Agarwal, D.~Schuurmans, and M.~Norouzi, ``An optimistic perspective on
  offline reinforcement learning,'' in \emph{Proc. Int. Conf. Mach.
  Learn.}\hskip 1em plus 0.5em minus 0.4em\relax PMLR, 2020, pp. 104--114.

\bibitem{levine2020offline}
S.~Levine, A.~Kumar, G.~Tucker, and J.~Fu, ``Offline reinforcement learning:
  Tutorial, review, and perspectives on open problems,'' \emph{arXiv preprint
  arXiv:2005.01643}, 2020.

\bibitem{krueger2020active}
D.~Krueger, J.~Leike, O.~Evans, and J.~Salvatier, ``Active reinforcement
  learning: Observing rewards at a cost,'' \emph{arXiv preprint
  arXiv:2011.06709}, 2020.

\bibitem{daniel2014active}
C.~Daniel, M.~Viering, J.~Metz, O.~Kroemer, and J.~Peters, ``Active reward
  learning,'' \emph{Robot. Scien. Syst.}, vol.~98, Jul. 2014.

\bibitem{ng2000algorithms}
A.~Y. Ng and S.~J. Russell, ``Algorithms for inverse reinforcement learning.''
  in \emph{Proc. Int. Conf. Mach. Learn.}, vol.~1, 2000, p.~2.

\bibitem{fu2017learning}
J.~Fu, K.~Luo, and S.~Levine, ``Learning robust rewards with adversarial
  inverse reinforcement learning,'' \emph{arXiv preprint arXiv:1710.11248},
  2017.

\bibitem{wang2017origin}
H.~Wang and B.~Raj, ``On the origin of deep learning,'' \emph{arXiv preprint
  arXiv:1702.07800}, 2017.

\bibitem{lecun2015deep}
Y.~LeCun, Y.~Bengio, and G.~Hinton, ``Deep learning,'' \emph{Nature}, vol. 521,
  no. 7553, pp. 436--444, May 2015.

\bibitem{sun2019optimization}
R.~Sun, ``Optimization for deep learning: theory and algorithms,'' \emph{arXiv
  preprint arXiv:1912.08957}, 2019.

\bibitem{Goodfellow-et-al-2016}
I.~Goodfellow, Y.~Bengio, and A.~Courville, \emph{Deep Learning}.\hskip 1em
  plus 0.5em minus 0.4em\relax MIT Press, 2016,
  \url{http://www.deeplearningbook.org}.

\bibitem{annfig}
\BIBentryALTinterwordspacing
{CS231n: Convolutional Neural Networks for Visual Recognition}.
  (Stanford-Spring 2021). [Online]. Available:
  \url{https://cs231n.github.io/convolutional-networks/#conv.}
\BIBentrySTDinterwordspacing

\bibitem{hochreiter1997long}
S.~Hochreiter and J.~Schmidhuber, ``Long short-term memory,'' \emph{Neural
  Comput.}, vol.~9, no.~8, pp. 1735--1780, Nov. 1997.

\bibitem{hinton2006reducing}
G.~E. Hinton and R.~R. Salakhutdinov, ``Reducing the dimensionality of data
  with neural networks,'' \emph{Science}, vol. 313, no. 5786, pp. 504--507,
  Jul. 2006.

\bibitem{autofig}
\BIBentryALTinterwordspacing
A.~Dertat. Applied deep learning-part 3: Autoencoders. (Oct. 2017). [Online].
  Available:
  \url{https://towardsdatascience.com/applied-deep-learning-part-3-autoencoders-1c083af4d798.}
\BIBentrySTDinterwordspacing

\bibitem{mnih2015human}
V.~Mnih, K.~Kavukcuoglu, D.~Silver \emph{et~al.}, ``Human-level control through
  deep reinforcement learning,'' \emph{Nature}, vol. 518, no. 7540, pp.
  529--533, Feb. 2015.

\bibitem{fan2020theoretical}
J.~Fan, Z.~Wang, Y.~Xie, and Z.~Yang, ``A theoretical analysis of deep
  q-learning,'' in \emph{Learning for Dynamics and Control}.\hskip 1em plus
  0.5em minus 0.4em\relax PMLR, 2020, pp. 486--489.

\bibitem{van2016deep}
H.~Van~Hasselt, A.~Guez, and D.~Silver, ``Deep reinforcement learning with
  double q-learning,'' in \emph{Proc. AAAI Conf. on Artif. Intel.}, vol.~30,
  no.~1, 2016.

\bibitem{wang2016dueling}
Z.~Wang, T.~Schaul, M.~Hessel, H.~Hasselt, M.~Lanctot, and N.~Freitas,
  ``Dueling network architectures for deep reinforcement learning,'' in
  \emph{Proc. Int. Conf. Mach. Learn.}\hskip 1em plus 0.5em minus 0.4em\relax
  PMLR, 2016, pp. 1995--2003.

\bibitem{mnih2016asynchronous}
V.~Mnih, A.~P. Badia, M.~Mirza, A.~Graves, T.~Lillicrap, T.~Harley, D.~Silver,
  and K.~Kavukcuoglu, ``Asynchronous methods for deep reinforcement learning,''
  in \emph{Proc. Int. Conf. Mach. Learn.}, 2016, pp. 1928--1937.

\bibitem{haarnoja2018soft}
T.~Haarnoja, A.~Zhou, P.~Abbeel, and S.~Levine, ``Soft actor-critic: Off-policy
  maximum entropy deep reinforcement learning with a stochastic actor,''
  \emph{arXiv preprint arXiv:1801.01290}, 2018.

\bibitem{schulman2015trust}
J.~Schulman, S.~Levine, P.~Abbeel, M.~Jordan, and P.~Moritz, ``Trust region
  policy optimization,'' in \emph{Proc. Int. Conf. Mach. Learn.}, 2015, pp.
  1889--1897.

\bibitem{schulman2017proximal}
J.~Schulman, F.~Wolski, P.~Dhariwal, A.~Radford, and O.~Klimov, ``Proximal
  policy optimization algorithms,'' \emph{arXiv preprint arXiv:1707.06347},
  2017.

\bibitem{zhang2019multi}
K.~Zhang, Z.~Yang, and T.~Ba{\c{s}}ar, ``Multi-agent reinforcement learning: A
  selective overview of theories and algorithms,'' \emph{arXiv preprint
  arXiv:1911.10635}, 2019.

\bibitem{bevrani_intelligent_nodate}
H.~Bevrani and T.~Hiyama, \emph{\BIBforeignlanguage{en}{Intelligent {Automatic}
  {Generation} {Control}}}.\hskip 1em plus 0.5em minus 0.4em\relax CRC press,
  2017.

\bibitem{cui2020reinforcement}
W.~Cui and B.~Zhang, ``Reinforcement learning for optimal frequency control: A
  lyapunov approach,'' \emph{arXiv preprint arXiv:2009.05654}, 2020.

\bibitem{yan_multi-agent_2020}
Z.~Yan and Y.~Xu, ``A multi-agent deep reinforcement learning method for
  cooperative load frequency control of a multi-area power system,''
  \emph{{IEEE} Trans. Power Syst.}, vol.~35, no.~6, Nov. 2020.

\bibitem{li_deep_2020}
J.~Li and T.~Yu, ``Deep reinforcement learning based multi-objective integrated
  automatic generation control for multiple continuous power disturbances,''
  \emph{IEEE Access}, vol.~8, pp. 156\,839--156\,850, Aug. 2020.

\bibitem{khooban_novel_2020}
M.~H. Khooban and M.~Gheisarnejad, ``A novel deep reinforcement learning
  controller based type-{II} fuzzy system: Frequency regulation in
  microgrids,'' \emph{IEEE Trans. Emerg. Topics Comput. Intell.}, vol.~5,
  no.~4, pp. 689--699, Aug. 2021.

\bibitem{younesi_assessing_2020}
A.~Younesi, H.~Shayeghi, and P.~Siano, ``Assessing the use of reinforcement
  learning for integrated voltage/frequency control in {AC} microgrids,''
  \emph{Energies}, vol.~13, no.~5, Mar. 2020.

\bibitem{chen_model_2020}
C.~{Chen}, M.~{Cui}, F.~F. {Li}, S.~{Yin}, and X.~{Wang}, ``Model-free
  emergency frequency control based on reinforcement learning,'' \emph{{IEEE}
  Trans. Ind. Informat.}, vol.~17, no.~4, pp. 2336--2346, Apr. 2021.

\bibitem{abouheaf_load_2019}
M.~Abouheaf, Q.~Gueaieb, and A.~Sharaf, ``Load frequency regulation for
  multi-area power system using integral reinforcement learning,'' \emph{IET
  Gener. Transm. Distrib.}, vol.~13, no.~19, pp. 4311--4323, Oct. 2019.

\bibitem{yan_data-driven_2019}
Z.~Yan and Y.~Xu, ``Data-driven load frequency control for stochastic power
  systems: {A} deep reinforcement learning method with continuous action
  search,'' \emph{{IEEE} Trans. Power Syst.}, vol.~34, no.~2, pp. 1653--1656,
  Mar. 2019.

\bibitem{wang_multiobjective_2019}
H.~Wang, Z.~Lei, X.~Zhang, J.~Peng, and H.~Jiang, ``Multiobjective
  reinforcement learning-based intelligent approach for optimization of
  activation rules in automatic generation control,'' \emph{IEEE Access},
  vol.~7, pp. 17\,480--17\,492, Feb. 2019.

\bibitem{adibi2019reinforcement}
M.~Adibi and J.~van~der Woude, ``A reinforcement learning approach for
  frequency control of inverted-based microgrids,'' \emph{IFAC-PapersOnLine},
  vol.~52, no.~4, pp. 111--116, Jun. 2019.

\bibitem{xi_smart_2018}
L.~Xi, J.~Chen, Y.~Huang, Y.~Xu, L.~Liu, Y.~Zhou, and Y.~Li, ``Smart generation
  control based on multi-agent reinforcement learning with the idea of the time
  tunnel,'' \emph{Energy}, vol. 153, pp. 977--987, Jun. 2018.

\bibitem{yin_artificial_2017}
L.~Yin, T.~Yu, L.~Zhou, L.~Huang, X.~Zhang, and B.~Zheng, ``Artificial
  emotional reinforcement learning for automatic generation control of
  large-scale interconnected power grids,'' \emph{IET Gener. Transm. Distrib.},
  vol.~11, no.~9, pp. 2305--2313, Jun. 2017.

\bibitem{singh_distributed_2017}
V.~P. Singh, N.~Kishor, and P.~Samuel, ``Distributed multi-agent system-based
  load frequency control for multi-area power system in smart grid,''
  \emph{{IEEE} Trans. Ind. Electron.}, vol.~64, no.~6, pp. 5151--5160, Jun.
  2017.

\bibitem{yu_multi-agent_2015}
T.~Yu, H.~Z. Wang, B.~Zhou, K.~W. Chan, and J.~Tang, ``Multi-agent correlated
  equilibrium {Q}($\lambda$) learning for coordinated smart generation control
  of interconnected power grids,'' \emph{{IEEE} Trans. Power Syst.}, vol.~30,
  no.~4, pp. 1669--1679, Jul. 2015.

\bibitem{rozada_load_2020}
S.~Rozada, D.~Apostolopoulou, and E.~Alonso, ``Load frequency control: {A} deep
  multi-agent reinforcement learning approach,'' in \emph{Proc. IEEE Power
  Energy Soc. Gen. Meeting}, Montreal, Canada, Aug. 2020.

\bibitem{chen2020distributed}
X.~Chen, C.~Zhao, and N.~Li, ``Distributed automatic load frequency control
  with optimality in power systems,'' \emph{IEEE Control Netw. Syst.}, vol.~8,
  no.~1, pp. 307--318, Mar. 2021.

\bibitem{7944568}
E.~{Mallada}, C.~{Zhao}, and S.~{Low}, ``Optimal load-side control for
  frequency regulation in smart grids,'' \emph{IEEE Trans. Autom. Control},
  vol.~62, no.~12, pp. 6294--6309, Dec. 2017.

\bibitem{8636257}
H.~Sun, Q.~Guo, J.~Qi, V.~Ajjarapu, R.~Bravo, J.~Chow, Z.~Li, R.~Moghe,
  E.~Nasr-Azadani, U.~Tamrakar, G.~N. Taranto, R.~Tonkoski, G.~Valverde, Q.~Wu,
  and G.~Yang, ``Review of challenges and research opportunities for voltage
  control in smart grids,'' \emph{IEEE Trans. Power Syst.}, vol.~34, no.~4, pp.
  2790--2801, Jul. 2019.

\bibitem{7874216}
K.~E. {Antoniadou-Plytaria}, I.~N. {Kouveliotis-Lysikatos}, P.~S.
  {Georgilakis}, and N.~D. {Hatziargyriou}, ``Distributed and decentralized
  voltage control of smart distribution networks: Models, methods, and future
  research,'' \emph{IEEE Trans. Smart Grid}, vol.~8, no.~6, pp. 2999--3008,
  Nov. 2017.

\bibitem{qu2019optimal}
G.~Qu and N.~Li, ``Optimal distributed feedback voltage control under limited
  reactive power,'' \emph{IEEE Trans. Power Syst.}, vol.~35, no.~1, pp.
  315--331, Jan. 2019.

\bibitem{magnusson2020distributed}
S.~{Magnússon}, G.~{Qu}, and N.~{Li}, ``Distributed optimal voltage control
  with asynchronous and delayed communication,'' \emph{IEEE Trans. Smart Grid},
  vol.~11, no.~4, pp. 3469--3482, Jul. 2020.

\bibitem{8859389}
S.~M. Mohiuddin and J.~Qi, ``Droop-free distributed control for ac microgrids
  with precisely regulated voltage variance and admissible voltage profile
  guarantees,'' \emph{IEEE Trans. Smart Grid}, vol.~11, no.~3, pp. 1956--1967,
  May 2020.

\bibitem{shi2021stability}
Y.~Shi, G.~Qu, S.~Low, A.~Anandkumar, and A.~Wierman, ``Stability constrained
  reinforcement learning for real-time voltage control,'' \emph{arXiv preprint
  arXiv:2109.14854}, 2021.

\bibitem{lee2021graph}
X.~Y. Lee, S.~Sarkar, and Y.~Wang, ``A graph policy network approach for
  volt-var control in power distribution systems,'' \emph{arXiv preprint
  arXiv:2109.12073}, 2021.

\bibitem{9356806}
H.~Liu and W.~Wu, ``Online multi-agent reinforcement learning for decentralized
  inverter-based volt-var control,'' \emph{IEEE Trans. Smart Grid}, vol.~12,
  no.~4, pp. 2980--2990, Jul. 2021.

\bibitem{9358213}
L.~Yin, C.~Zhang, Y.~Wang, F.~Gao, J.~Yu, and L.~Cheng, ``Emotional deep
  learning programming controller for automatic voltage control of power
  systems,'' \emph{IEEE Access}, vol.~9, pp. 31\,880--31\,891, Feb. 2021.

\bibitem{9353702}
Y.~Gao, W.~Wang, and N.~Yu, ``Consensus multi-agent reinforcement learning for
  volt-var control in power distribution networks,'' \emph{IEEE Trans. Smart
  Grid}, vol.~12, no.~4, pp. 3594--3604, Jul. 2021.

\bibitem{9328796}
X.~Sun and J.~Qiu, ``Two-stage volt/var control in active distribution networks
  with multi-agent deep reinforcement learning method,'' \emph{IEEE Trans.
  Smart Grid}, vol.~12, no.~4, pp. 2903--2912, Jul. 2021.

\bibitem{9143169}
Y.~Zhang, X.~Wang, J.~Wang, and Y.~Zhang, ``Deep reinforcement learning based
  volt-var optimization in smart distribution systems,'' \emph{IEEE Trans.
  Smart Grid}, vol.~12, no.~1, pp. 361--371, Jan. 2021.

\bibitem{mukherjee2021scalable}
S.~Mukherjee, R.~Huang, Q.~Huang, T.~L. Vu, and T.~Yin, ``Scalable voltage
  control using structure-driven hierarchical deep reinforcement learning,''
  \emph{arXiv preprint arXiv:2102.00077}, 2021.

\bibitem{kou2020safe}
P.~Kou, D.~Liang, C.~Wang, Z.~Wu, and L.~Gao, ``Safe deep reinforcement
  learning-based constrained optimal control scheme for active distribution
  networks,'' \emph{Applied Energy}, vol. 264, p. 114772, Apr. 2020.

\bibitem{8985179}
M.~Al-Saffar and P.~Musilek, ``Reinforcement learning-based distributed {BESS}
  management for mitigating overvoltage issues in systems with high pv
  penetration,'' \emph{IEEE Trans. Smart Grid}, vol.~11, no.~4, pp. 2980--2994,
  Jul. 2020.

\bibitem{vlachogiannis_reinforcement_2004}
J.~Vlachogiannis and N.~Hatziargyriou, ``Reinforcement learning for reactive
  power control,'' \emph{IEEE Trans. Power Syst.}, vol.~19, no.~3, pp.
  1317--1325, Aug. 2004.

\bibitem{xu_optimal_2020}
H.~Xu, A.~D. Domínguez-García, and P.~W. Sauer, ``Optimal tap setting of
  voltage regulation transformers using batch reinforcement learning,''
  \emph{IEEE Trans. Power Syst.}, vol.~35, no.~3, pp. 1990--2001, May 2020.

\bibitem{yang_two-timescale_2020}
Q.~Yang, G.~Wang, A.~Sadeghi, G.~B. Giannakis, and J.~Sun, ``Two-timescale
  voltage control in distribution grids using deep reinforcement learning,''
  \emph{IEEE Trans. Smart Grid}, vol.~11, no.~3, pp. 2313--2323, May 2020.

\bibitem{wang_data-driven_2020}
S.~Wang, J.~Duan, D.~Shi, C.~Xu, H.~Li, R.~Diao, and Z.~Wang, ``A data-driven
  multi-agent autonomous voltage control framework using deep reinforcement
  learning,'' \emph{IEEE Trans. Power Syst.}, vol.~35, no.~6, pp. 4644--4654,
  Nov. 2020.

\bibitem{wang_safe_2020}
W.~Wang, N.~Yu, Y.~Gao, and J.~Shi, ``Safe off-policy deep reinforcement
  learning algorithm for {Volt}-{VAR} control in power distribution systems,''
  \emph{IEEE Trans. Smart Grid}, vol.~11, no.~4, pp. 3008--3018, Jul. 2020.

\bibitem{duan_deep-reinforcement-learning-based_2020}
J.~Duan, D.~Shi, R.~Diao, H.~Li, Z.~Wang, B.~Zhang, D.~Bian, and Z.~Yi,
  ``Deep-reinforcement-learning-based autonomous voltage control for power grid
  operations,'' \emph{IEEE Trans. Power Syst.}, vol.~35, no.~1, pp. 814--817,
  Jan. 2020.

\bibitem{cao_multi-agent_2020}
D.~Cao, W.~Hu, J.~Zhao, Q.~Huang, Z.~Chen, and F.~Blaabjerg, ``A multi-agent
  deep reinforcement learning based voltage regulation using coordinated pv
  inverters,'' \emph{IEEE Trans. Power Syst.}, vol.~35, no.~5, pp. 4120--4123,
  Sept. 2020.

\bibitem{liu2020two}
H.~Liu and W.~Wu, ``Two-stage deep reinforcement learning for inverter-based
  volt-var control in active distribution networks,'' \emph{IEEE Trans. Smart
  Grid}, vol.~12, no.~3, pp. 2037--2047, May 2021.

\bibitem{pinto2017robust}
L.~Pinto, J.~Davidson, R.~Sukthankar, and A.~Gupta, ``Robust adversarial
  reinforcement learning,'' \emph{arXiv preprint arXiv:1703.02702}, 2017.

\bibitem{liu2021bi}
H.~Liu and W.~Wu, ``Bi-level off-policy reinforcement learning for volt/var
  control involving continuous and discrete devices,'' \emph{arXiv preprint
  arXiv:2104.05902}, 2021.

\bibitem{parisotto2015actor}
E.~Parisotto, J.~L. Ba, and R.~Salakhutdinov, ``Actor-mimic: Deep multitask and
  transfer reinforcement learning,'' \emph{arXiv preprint arXiv:1511.06342},
  2015.

\bibitem{sun2012family}
H.~Sun, B.~Zhang, W.~Wu, and Q.~Guo, ``Family of energy management system for
  smart grid,'' in \emph{Proc. 3rd IEEE PES Innov. Smart Grid Technol. Int.
  Conf. Exhibit.}, Berlin, Germany, 2012, pp. 1--5.

\bibitem{buildweb}
\BIBentryALTinterwordspacing
{Office of Energy Efficiency \& Renewable Energy (EERE)}. 2011 buildings energy
  data book. [Online]. Available:
  \url{https://catalog.data.gov/dataset/buildings-energy-data-book.}
\BIBentrySTDinterwordspacing

\bibitem{demandresene}
{U.S. Department of Energy}, ``Benefit of demand response in electricity market
  and recommendations for achieving them,'' Feb. 2006.

\bibitem{chen2020online}
X.~Chen, Y.~Li, J.~Shimada, and N.~Li, ``Online learning and distributed
  control for residential demand response,'' \emph{IEEE Trans. Smart Grid},
  vol.~12, no.~6, pp. 4843--4853, Nov. 2021.

\bibitem{lu2019demand}
R.~Lu, S.~H. Hong, and M.~Yu, ``Demand response for home energy management
  using reinforcement learning and artificial neural network,'' \emph{IEEE
  Trans. Smart Grid}, vol.~10, no.~6, pp. 6629--6639, Nov. 2019.

\bibitem{huang2020multienergy}
W.~Huang, N.~Zhang, Y.~Cheng, J.~Yang, Y.~Wang, and C.~Kang, ``Multienergy
  networks analytics: standardized modeling, optimization, and low carbon
  analysis,'' \emph{Proceedings of the IEEE}, vol. 108, no.~9, pp. 1411--1436,
  Sept. 2020.

\bibitem{9361643}
Z.~Xu, G.~Han, L.~Liu, M.~Martínez-García, and Z.~Wang, ``Multi-energy
  scheduling of an industrial integrated energy system by reinforcement
  learning based differential evolution,'' \emph{IEEE Trans. Green Commun.
  Netw.}, vol.~5, no.~3, pp. 1077--1090, Sept. 2021.

\bibitem{9016168}
Y.~{Ye}, D.~{Qiu}, X.~{Wu}, G.~{Strbac}, and J.~{Ward}, ``Model-free real-time
  autonomous control for a residential multi-energy system using deep
  reinforcement learning,'' \emph{IEEE Trans. Smart Grid}, vol.~11, no.~4, pp.
  3068--3082, Jul. 2020.

\bibitem{9346692}
Y.~Wang, Z.~Yang, L.~Dong, S.~Huang, and W.~Zhou, ``Energy management of
  integrated energy system based on stackelberg game and deep reinforcement
  learning,'' in \emph{Proc. IEEE 4th Conf. Energy Inter. and Energy System
  Integ.}, 2020, pp. 2645--2651.

\bibitem{ceusters2021model}
G.~Ceusters, R.~C. Rodr{\'\i}guez, A.~B. Garc{\'\i}a, R.~Franke, G.~Deconinck,
  L.~Helsen, A.~Now{\'e}, M.~Messagie, and L.~R. Camargo, ``Model-predictive
  control and reinforcement learning in multi-energy system case studies,''
  \emph{arXiv preprint arXiv:2104.09785}, 2021.

\bibitem{9069289}
Z.~{Yan} and Y.~{Xu}, ``Real-time optimal power flow: A lagrangian based deep
  reinforcement learning approach,'' \emph{IEEE Trans. Power Syst.}, vol.~35,
  no.~4, pp. 3270--3273, Jul. 2020.

\bibitem{8853513}
C.~{Jiang}, Z.~{Li}, J.~H. {Zheng}, Q.~H. {Wu}, and X.~{Shang}, ``Two-level
  area-load modelling for opf of power system using reinforcement learning,''
  \emph{IET Gener., Transm. Distrb.}, vol.~13, no.~18, pp. 4141--4149, Sept.
  2019.

\bibitem{9272783}
J.~H. Woo, L.~Wu, J.-B. Park, and J.~H. Roh, ``Real-time optimal power flow
  using twin delayed deep deterministic policy gradient algorithm,'' \emph{IEEE
  Access}, vol.~8, pp. 213\,611--213\,618, Nov. 2020.

\bibitem{9275611}
Y.~Zhou, B.~Zhang, C.~Xu, T.~Lan, R.~Diao, D.~Shi, Z.~Wang, and W.-J. Lee, ``A
  data-driven method for fast {A}{C} optimal power flow solutions via deep
  reinforcement learning,'' \emph{J. Mod. Power Syst. Clean Energy}, vol.~8,
  no.~6, pp. 1128--1139, Nov. 2020.

\bibitem{9465777}
D.~Cao, W.~Hu, X.~Xu, Q.~Wu, Q.~Huang, Z.~Chen, and F.~Blaabjerg, ``Deep
  reinforcement learning based approach for optimal power flow of distribution
  networks embedded with renewable energy and storage devices,'' \emph{J. Mod.
  Power Syst. Clean Energy}, vol.~9, no.~5, pp. 1101--1110, Sept. 2021.

\bibitem{8957677}
L.~{Lin}, X.~{Guan}, Y.~{Peng}, N.~{Wang}, S.~{Maharjan}, and T.~{Ohtsuki},
  ``Deep reinforcement learning for economic dispatch of virtual power plant in
  internet of energy,'' \emph{IEEE Internet of Things J.}, vol.~7, no.~7, pp.
  6288--6301, Jul. 2020.

\bibitem{8789677}
Q.~{Zhang}, K.~{Dehghanpour}, Z.~{Wang}, and Q.~{Huang}, ``A learning-based
  power management method for networked microgrids under incomplete
  information,'' \emph{IEEE Trans. Smart Grid}, vol.~11, no.~2, pp. 1193--1204,
  Mar. 2020.

\bibitem{9464628}
R.~Hao, T.~Lu, Q.~Ai, and H.~He, ``Distributed online dispatch for microgrids
  using hierarchical reinforcement learning embedded with operation
  knowledge,'' \emph{IEEE Trans. Power Syst.}, pp. 1--1, 2021.

\bibitem{9245590}
H.~Shuai and H.~He, ``Online scheduling of a residential microgrid via
  monte-carlo tree search and a learned model,'' \emph{IEEE Trans. Smart Grid},
  vol.~12, no.~2, pp. 1073--1087, Mar. 2021.

\bibitem{8769895}
Y.~Du and F.~Li, ``Intelligent multi-microgrid energy management based on deep
  neural network and model-free reinforcement learning,'' \emph{IEEE Trans.
  Smart Grid}, vol.~11, no.~2, pp. 1066--1076, Mar. 2020.

\bibitem{8306311}
W.~{Liu}, P.~{Zhuang}, H.~{Liang}, J.~{Peng}, and Z.~{Huang}, ``Distributed
  economic dispatch in microgrids based on cooperative reinforcement
  learning,'' \emph{IEEE Trans. Neural Netw. Learn. Syst.}, vol.~29, no.~6, pp.
  2192--2203, Jun. 2018.

\bibitem{wan_modelfree_realtime_2019}
Z.~Wan, H.~Li, H.~He, and D.~Prokhorov, ``Model-free real-time {{EV}} charging
  scheduling based on deep reinforcement learning,'' \emph{{IEEE} Trans. Smart
  Grid}, vol.~10, no.~5, pp. 5246--5257, Sep. 2019.

\bibitem{li_constrained_ev_2020}
H.~Li, Z.~Wan, and H.~He, ``Constrained {{EV}} charging scheduling based on
  safe deep reinforcement learning,'' \emph{{IEEE} Trans. Smart Grid}, vol.~11,
  no.~3, pp. 2427--2439, May 2020.

\bibitem{achiam_constrained_policy_2017}
J.~Achiam, D.~Held, A.~Tamar, and P.~Abbeel, ``Constrained policy
  optimization,'' in \emph{Proc. of the 34th Int. Conf. Mach. Learn.}, 2017, p.
  22–31.

\bibitem{9371688}
H.~M. Abdullah, A.~Gastli, and L.~Ben-Brahim, ``Reinforcement learning based
  {E}{V} charging management systems–a review,'' \emph{IEEE Access}, vol.~9,
  pp. 41\,506--41\,531, 2021.

\bibitem{bui_double_deep_2020}
V.-H. Bui, A.~Hussain, and H.-M. Kim, ``Double deep ${{Q}}$ -learning-based
  distributed operation of battery energy storage system considering
  uncertainties,'' \emph{{IEEE} Trans. Smart Grid}, vol.~11, no.~1, pp.
  457--469, Jan. 2020.

\bibitem{9061038}
J.~Cao, D.~Harrold, Z.~Fan, T.~Morstyn, D.~Healey, and K.~Li, ``Deep
  reinforcement learning-based energy storage arbitrage with accurate
  lithium-ion battery degradation model,'' \emph{IEEE Trans. Smart Grid},
  vol.~11, no.~5, pp. 4513--4521, Sept. 2020.

\bibitem{9097915}
F.~Sanchez~Gorostiza and F.~M. Gonzalez-Longatt, ``Deep reinforcement
  learning-based controller for {S}{O}{C} management of multi-electrical energy
  storage system,'' \emph{IEEE Trans. Smart Grid}, vol.~11, no.~6, pp.
  5039--5050, Nov. 2020.

\bibitem{wei2017deep}
T.~Wei, Y.~Wang, and Q.~Zhu, ``Deep reinforcement learning for building
  {H}{V}{A}{C} control,'' in \emph{Proc. of the 54th Annual Design Autom.
  Conf.}, 2017, pp. 1--6.

\bibitem{9085925}
G.~Gao, J.~Li, and Y.~Wen, ``Deepcomfort: Energy-efficient thermal comfort
  control in buildings via reinforcement learning,'' \emph{IEEE Internet of
  Things J.}, vol.~7, no.~9, pp. 8472--8484, Sept. 2020.

\bibitem{9146920}
L.~Yu, Y.~Sun, Z.~Xu, C.~Shen, D.~Yue, T.~Jiang, and X.~Guan, ``Multi-agent
  deep reinforcement learning for {HVAC} control in commercial buildings,''
  \emph{IEEE Trans. Smart Grid}, vol.~12, no.~1, pp. 407--419, Jan. 2021.

\bibitem{mocanu2018line}
E.~Mocanu, D.~C. Mocanu, P.~H. Nguyen, A.~Liotta, M.~E. Webber, M.~Gibescu, and
  J.~G. Slootweg, ``On-line building energy optimization using deep
  reinforcement learning,'' \emph{IEEE Trans. Smart Grid}, vol.~10, no.~4, pp.
  3698--3708, Jul. 2019.

\bibitem{9161266}
X.~Zhang, D.~Biagioni, M.~Cai, P.~Graf, and S.~Rahman, ``An edge-cloud
  integrated solution for buildings demand response using reinforcement
  learning,'' \emph{IEEE Trans. Smart Grid}, vol.~12, no.~1, pp. 420--431, Jan.
  2021.

\bibitem{xu2020multi}
X.~Xu, Y.~Jia, Y.~Xu, Z.~Xu, S.~Chai, and C.~S. Lai, ``A multi-agent
  reinforcement learning based data-driven method for home energy management,''
  \emph{IEEE Trans. Smart Grid}, vol.~11, no.~4, pp. 3201--3211, Jul. 2020.

\bibitem{li_real-time_2020}
H.~Li, Z.~Wan, and H.~He, ``Real-time residential demand response,'' \emph{IEEE
  Trans. Smart Grid}, vol.~11, no.~5, pp. 4144--4154, Sep. 2020.

\bibitem{alfaverh2020demand}
F.~Alfaverh, M.~Denai, and Y.~Sun, ``Demand response strategy based on
  reinforcement learning and fuzzy reasoning for home energy management,''
  \emph{IEEE Access}, vol.~8, pp. 39\,310--39\,321, Feb. 2020.

\bibitem{liu2020optimization}
Y.~Liu, D.~Zhang, and H.~B. Gooi, ``Optimization strategy based on deep
  reinforcement learning for home energy management,'' \emph{CSEE J. Power
  Energy Syst.}, vol.~6, no.~3, pp. 572--582, Sept. 2020.

\bibitem{yu2019deep}
L.~Yu, W.~Xie, D.~Xie, Y.~Zou, D.~Zhang, Z.~Sun, L.~Zhang, Y.~Zhang, and
  T.~Jiang, ``Deep reinforcement learning for smart home energy management,''
  \emph{IEEE Internet Things J.}, vol.~7, no.~4, pp. 2751--2762, Apr. 2019.

\bibitem{ye_modelfree_realtime_2020}
Y.~Ye, D.~Qiu, X.~Wu, G.~Strbac, and J.~Ward, ``Model-free real-time autonomous
  control for a residential multi-energy system using deep reinforcement
  learning,'' \emph{{IEEE} Trans. Smart Grid}, vol.~11, no.~4, pp. 3068--3082,
  Jul. 2020.

\bibitem{silva2020coordination}
F.~L. Silva, C.~E.~H. Nishida, D.~M. Roijers, and A.~H.~R. Costa,
  ``Coordination of electric vehicle charging through multiagent reinforcement
  learning,'' \emph{{IEEE} Trans. Smart Grid}, vol.~11, no.~3, pp. 2347--2356,
  May 2020.

\bibitem{8826539}
Y.~{Ye}, D.~{Qiu}, J.~{Li}, and G.~{Strbac}, ``Multi-period and multi-spatial
  equilibrium analysis in imperfect electricity markets: A novel multi-agent
  deep reinforcement learning approach,'' \emph{IEEE Access}, vol.~7, pp.
  130\,515--130\,529, Sept. 2019.

\bibitem{8805177}
Y.~{Ye}, D.~{Qiu}, M.~{Sun}, D.~{Papadaskalopoulos}, and G.~{Strbac}, ``Deep
  reinforcement learning for strategic bidding in electricity markets,''
  \emph{IEEE Trans. Smart Grid}, vol.~11, no.~2, pp. 1343--1355, Mar. 2020.

\bibitem{gao2020batch}
Y.~Gao, W.~Wang, J.~Shi, and N.~Yu, ``Batch-constrained reinforcement learning
  for dynamic distribution network reconfiguration,'' \emph{IEEE Trans. Smart
  Grid}, vol.~11, no.~6, pp. 5357--5369, Nov. 2020.

\bibitem{8862818}
L.~R. {Ferreira}, A.~R. {Aoki}, and G.~{Lambert-Torres}, ``A reinforcement
  learning approach to solve service restoration and load management
  simultaneously for distribution networks,'' \emph{IEEE Access}, vol.~7, pp.
  145\,978--145\,987, Oct. 2019.

\bibitem{5871714}
D.~{Ye}, M.~{Zhang}, and D.~{Sutanto}, ``A hybrid multiagent framework with
  q-learning for power grid systems restoration,'' \emph{IEEE Trans. Power
  Syst.}, vol.~26, no.~4, pp. 2434--2441, Nov. 2011.

\bibitem{8787888}
Q.~{Huang}, R.~{Huang}, W.~{Hao}, J.~{Tan}, R.~{Fan}, and Z.~{Huang},
  ``Adaptive power system emergency control using deep reinforcement
  learning,'' \emph{IEEE Trans. Smart Grid}, vol.~11, no.~2, pp. 1171--1182,
  Mar. 2020.

\bibitem{7081385}
C.~{Wei}, Z.~{Zhang}, W.~{Qiao}, and L.~{Qu}, ``Reinforcement-learning-based
  intelligent maximum power point tracking control for wind energy conversion
  systems,'' \emph{IEEE Trans. Ind. Electron.}, vol.~62, no.~10, pp.
  6360--6370, Oct. 2015.

\bibitem{9079637}
A.~{Kushwaha}, M.~{Gopal}, and B.~{Singh}, ``Q-learning based maximum power
  extraction for wind energy conversion system with variable wind speed,''
  \emph{IEEE Trans. Energy Convers.}, vol.~35, no.~3, pp. 1160--1170, Sept.
  2020.

\bibitem{8786811}
D.~{An}, Q.~{Yang}, W.~{Liu}, and Y.~{Zhang}, ``Defending against data
  integrity attacks in smart grid: A deep reinforcement learning-based
  approach,'' \emph{IEEE Access}, vol.~7, pp. 110\,835--110\,845, Aug. 2019.

\bibitem{8248780}
Y.~{Chen}, S.~{Huang}, F.~{Liu}, Z.~{Wang}, and X.~{Sun}, ``Evaluation of
  reinforcement learning-based false data injection attack to automatic voltage
  control,'' \emph{IEEE Trans. Smart Grid}, vol.~10, no.~2, pp. 2158--2169,
  Mar. 2019.

\bibitem{8998302}
Y.~{Shang}, W.~{Wu}, J.~{Liao}, J.~{Guo}, J.~{Su}, W.~{Liu}, and Y.~{Huang},
  ``Stochastic maintenance schedules of active distribution networks based on
  monte-carlo tree search,'' \emph{IEEE Trans. Power Syst.}, vol.~35, no.~5,
  pp. 3940--3952, Sept. 2020.

\bibitem{9029268}
D.~{Wu}, X.~{Zheng}, D.~{Kalathil}, and L.~{Xie}, ``Nested reinforcement
  learning based control for protective relays in power distribution systems,''
  in \emph{Proc. IEEE 58th Conf. Decision and Control}, 2019, pp. 1925--1930.

\bibitem{8845652}
T.~{Qian}, C.~{Shao}, X.~{Wang}, and M.~{Shahidehpour}, ``Deep reinforcement
  learning for {E}{V} charging navigation by coordinating smart grid and
  intelligent transportation system,'' \emph{IEEE Trans. Smart Grid}, vol.~11,
  no.~2, pp. 1714--1723, Mar. 2020.

\bibitem{online2021mab}
X.~{Chen}, Y.~{Nie}, and N.~{Li}, ``Online residential demand response via
  contextual multi-armed bandits,'' \emph{IEEE Contr. Syst. Lett.}, vol.~5,
  no.~2, pp. 433--438, Apr. 2021.

\bibitem{li2020real}
T.~Li, B.~Sun, Y.~Chen, Z.~Ye, S.~H. Low, and A.~Wierman, ``Real-time aggregate
  flexibility via reinforcement learning,'' \emph{arXiv preprint
  arXiv:2012.11261}, 2020.

\bibitem{brockman2016openai}
G.~Brockman, V.~Cheung, L.~Pettersson, J.~Schneider, J.~Schulman, J.~Tang, and
  W.~Zaremba, ``Openai gym,'' \emph{arXiv preprint arXiv:1606.01540}, 2016.

\bibitem{henry2021gym}
R.~Henry and D.~Ernst, ``Gym-anm: Reinforcement learning environments for
  active network management tasks in electricity distribution systems,''
  \emph{Energy and AI}, p. 100092, 2021.

\bibitem{marot2021learning}
A.~Marot, B.~Donnot, G.~Dulac-Arnold, A.~Kelly, A.~O'Sullivan, J.~Viebahn,
  M.~Awad, I.~Guyon, P.~Panciatici, and C.~Romero, ``Learning to run a power
  network challenge: a retrospective analysis,'' \emph{arXiv preprint
  arXiv:2103.03104}, 2021.

\bibitem{henri2020pymgrid}
G.~Henri, T.~Levent, A.~Halev, R.~Alami, and P.~Cordier, ``pymgrid: An
  open-source python microgrid simulator for applied artificial intelligence
  research,'' \emph{arXiv preprint arXiv:2011.08004}, 2020.

\bibitem{heid2020omg}
S.~Heid, D.~Weber, H.~Bode, E.~H{\"u}llermeier, and O.~Wallscheid, ``Omg: A
  scalable and flexible simulation and testing environment toolbox for
  intelligent microgrid control,'' \emph{J. Open Source Softw.}, vol.~5,
  no.~54, p. 2435, Oct. 2020.

\bibitem{fan2021powergym}
T.-H. Fan, X.~Y. Lee, and Y.~Wang, ``Powergym: A reinforcement learning
  environment for volt-var control in power distribution systems,'' \emph{arXiv
  preprint arXiv:2109.03970}, 2021.

\bibitem{abadi2016tensorflow}
M.~Abadi, P.~Barham, J.~Chen, Z.~Chen, A.~Davis, J.~Dean, M.~Devin,
  S.~Ghemawat, G.~Irving, M.~Isard \emph{et~al.}, ``Tensorflow: A system for
  large-scale machine learning,'' in \emph{12th Sympos. Operat. Syst. Design
  and Implem.}, Nov. 2016, pp. 265--283.

\bibitem{paszke2019pytorch}
A.~Paszke, S.~Gross, F.~Massa, A.~Lerer, J.~Bradbury, G.~Chanan, T.~Killeen,
  Z.~Lin, N.~Gimelshein, L.~Antiga \emph{et~al.}, ``Pytorch: An imperative
  style, high-performance deep learning library,'' \emph{Adv. Neural Inf.
  Process.}, vol.~32, pp. 8026--8037, 2019.

\bibitem{seide2016cntk}
F.~Seide and A.~Agarwal, ``{CNTK}: Microsoft's open-source deep-learning
  toolkit,'' in \emph{Proc. 22nd ACM SIGKDD Intern. Conf. Knowl. Discov. and
  Data Mining}, 2016, pp. 2135--2135.

\bibitem{matrltool}
{MATLAB and Reinforcement Learning Toolbox (R2021a)}.\hskip 1em plus 0.5em
  minus 0.4em\relax Natick, Massachusetts, U.S.: The MathWorks Inc., 2021.

\bibitem{berkenkamp2017safe}
F.~Berkenkamp, M.~Turchetta, A.~Schoellig, and A.~Krause, ``Safe model-based
  reinforcement learning with stability guarantees,'' in \emph{Adv. Neural Inf.
  Process. Syst.}, 2017, pp. 908--918.

\bibitem{koller2018learning}
T.~Koller, F.~Berkenkamp, M.~Turchetta, and A.~Krause, ``Learning-based model
  predictive control for safe exploration,'' in \emph{Proc. IEEE Conf. on
  Decision and Control}, Miami Beach, FL, 2018, pp. 6059--6066.

\bibitem{chow2018lyapunov}
Y.~Chow, O.~Nachum, E.~Duenez-Guzman, and M.~Ghavamzadeh, ``A lyapunov-based
  approach to safe reinforcement learning,'' in \emph{Adv. Neural Inf. Process.
  Syst.}, 2018, pp. 8092--8101.

\bibitem{fan2019safety}
J.~Fan and W.~Li, ``Safety-guided deep reinforcement learning via online
  gaussian process estimation,'' \emph{arXiv preprint arXiv:1903.02526}, 2019.

\bibitem{garcia2015comprehensive}
J.~Garc{\i}a and F.~Fern{\'a}ndez, ``A comprehensive survey on safe
  reinforcement learning,'' \emph{J. Mach. Learn. Res.}, vol.~16, no.~1, pp.
  1437--1480, Aug. 2015.

\bibitem{morimoto2005robust}
J.~Morimoto and K.~Doya, ``Robust reinforcement learning,'' \emph{Neural
  Comput.}, vol.~17, no.~2, pp. 335--359, Feb. 2005.

\bibitem{tessler2018reward}
C.~Tessler, D.~J. Mankowitz, and S.~Mannor, ``Reward constrained policy
  optimization,'' \emph{arXiv preprint arXiv:1805.11074}, 2018.

\bibitem{cheng2019end}
R.~Cheng, G.~Orosz, R.~M. Murray, and J.~W. Burdick, ``End-to-end safe
  reinforcement learning through barrier functions for safety-critical
  continuous control tasks,'' in \emph{Proc. AAAI Conf. Artif. Intel.},
  vol.~33, no.~01, 2019, pp. 3387--3395.

\bibitem{qin2021density}
Z.~Qin, Y.~Chen, and C.~Fan, ``Density constrained reinforcement learning,'' in
  \emph{Inter. Conf. Machi. Learn.}\hskip 1em plus 0.5em minus 0.4em\relax
  PMLR, 2021, pp. 8682--8692.

\bibitem{wabersich2018linear}
K.~P. Wabersich and M.~N. Zeilinger, ``Linear model predictive safety
  certification for learning-based control,'' in \emph{2018 IEEE Conf. Decision
  and Control (CDC)}.\hskip 1em plus 0.5em minus 0.4em\relax IEEE, 2018, pp.
  7130--7135.

\bibitem{altman1999constrained}
E.~Altman, \emph{Constrained Markov decision processes}.\hskip 1em plus 0.5em
  minus 0.4em\relax CRC Press, 1999, vol.~7.

\bibitem{achiam2017constrained}
J.~Achiam, D.~Held, A.~Tamar, and P.~Abbeel, ``Constrained policy
  optimization,'' \emph{arXiv preprint arXiv:1705.10528}, 2017.

\bibitem{dean2019safely}
S.~Dean, S.~Tu, N.~Matni, and B.~Recht, ``Safely learning to control the
  constrained linear quadratic regulator,'' in \emph{2019 Ameri. Contr. Conf.
  (ACC)}.\hskip 1em plus 0.5em minus 0.4em\relax IEEE, 2019, pp. 5582--5588.

\bibitem{pattanaik2017robust}
A.~Pattanaik, Z.~Tang, S.~Liu, G.~Bommannan, and G.~Chowdhary, ``Robust deep
  reinforcement learning with adversarial attacks,'' \emph{arXiv preprint
  arXiv:1712.03632}, 2017.

\bibitem{9494982}
L.~Omnes, A.~Marot, and B.~Donnot, ``Adversarial training for a continuous
  robustness control problem in power systems,'' in \emph{2021 IEEE Madrid
  PowerTech}, 2021, pp. 1--6.

\bibitem{8944283}
S.~Paul, Z.~Ni, and C.~Mu, ``A learning-based solution for an adversarial
  repeated game in cyber–physical power systems,'' \emph{IEEE Trans. Neural
  Netw. Learn. Syst.}, vol.~31, no.~11, pp. 4512--4523, Nov. 2020.

\bibitem{pan2021improving}
A.~Pan, H.~Zhang, Y.~Chen, Y.~Shi \emph{et~al.}, ``Improving robustness of
  reinforcement learning for power system control with adversarial training,''
  \emph{arXiv preprint arXiv:2110.08956}, 2021.

\bibitem{mankowitz2019robust}
D.~J. Mankowitz, N.~Levine, R.~Jeong, Y.~Shi, J.~Kay, A.~Abdolmaleki, J.~T.
  Springenberg, T.~Mann, T.~Hester, and M.~Riedmiller, ``Robust reinforcement
  learning for continuous control with model misspecification,'' \emph{arXiv
  preprint arXiv:1906.07516}, 2019.

\bibitem{iyengar2005robust}
G.~N. Iyengar, ``Robust dynamic programming,'' \emph{Math. Oper. Res.},
  vol.~30, no.~2, pp. 257--280, May 2005.

\bibitem{nilim2005robust}
A.~Nilim and L.~El~Ghaoui, ``Robust control of markov decision processes with
  uncertain transition matrices,'' \emph{Oper. Res.}, vol.~53, no.~5, pp.
  780--798, Oct. 2005.

\bibitem{zhang2020robust}
K.~Zhang, T.~Sun, Y.~Tao, S.~Genc, S.~Mallya, and T.~Basar, ``Robust
  multi-agent reinforcement learning with model uncertainty.'' in
  \emph{NeurIPS}, 2020.

\bibitem{derman2018soft}
E.~Derman, D.~J. Mankowitz, T.~A. Mann, and S.~Mannor, ``Soft-robust
  actor-critic policy-gradient,'' \emph{arXiv preprint arXiv:1803.04848}, 2018.

\bibitem{qu2020scalable}
G.~Qu, A.~Wierman, and N.~Li, ``Scalable reinforcement learning of localized
  policies for multi-agent networked systems,'' \emph{Proc. Mach. Learn. Res.},
  vol.~1, p.~38, 2020.

\bibitem{lin2020distributed}
Y.~Lin, G.~Qu, L.~Huang, and A.~Wierman, ``Multi-agent reinforcement learning
  in time-varying networked systems,'' \emph{arXiv preprint arXiv:2006.06555},
  2020.

\bibitem{li2020sample}
G.~Li, Y.~Wei, Y.~Chi, Y.~Gu, and Y.~Chen, ``Sample complexity of asynchronous
  q-learning: Sharper analysis and variance reduction,'' \emph{arXiv preprint
  arXiv:2006.03041}, 2020.

\bibitem{kumar2019sample}
H.~Kumar, A.~Koppel, and A.~Ribeiro, ``On the sample complexity of actor-critic
  method for reinforcement learning with function approximation,'' \emph{arXiv
  preprint arXiv:1910.08412}, 2019.

\bibitem{silva2021encoding}
A.~Silva and M.~Gombolay, ``Encoding human domain knowledge to warm start
  reinforcement learning,'' in \emph{Proc. AAAI Conf. Artif. Intel.}, vol.~35,
  no.~6, 2021, pp. 5042--5050.

\bibitem{qu2020combining}
G.~Qu, C.~Yu, S.~Low, and A.~Wierman, ``Combining model-based and model-free
  methods for nonlinear control: A provably convergent policy gradient
  approach,'' \emph{arXiv preprint arXiv:2006.07476}, 2020.

\bibitem{hussein2017imitation}
A.~Hussein, M.~M. Gaber, E.~Elyan, and C.~Jayne, ``Imitation learning: A survey
  of learning methods,'' \emph{ACM Computing Surveys (CSUR)}, vol.~50, no.~2,
  pp. 1--35, 2017.

\bibitem{8809368}
M.~Ghorbanian, S.~H. Dolatabadi, and P.~Siano, ``Big data issues in smart
  grids: A survey,'' \emph{IEEE Systems Journal}, vol.~13, no.~4, pp.
  4158--4168, Dec. 2019.

\bibitem{wang_integrating_2019}
Q.~Wang, F.~Li, Y.~Tang, and Y.~Xu, ``Integrating model-driven and data-driven
  methods for power system frequency stability assessment and control,''
  \emph{{IEEE} Trans. Power Syst.}, vol.~34, no.~6, pp. 4557--4568, Nov. 2019.

\bibitem{che2018combining}
T.~Che, Y.~Lu, G.~Tucker, S.~Bhupatiraju, S.~Gu, S.~Levine, and Y.~Bengio,
  ``Combining model-based and model-free {RL} via multi-step control
  variates,'' \emph{OpenReview}, 2018.

\bibitem{wang2016learning}
J.~X. Wang, Z.~Kurth-Nelson, D.~Tirumala, H.~Soyer, J.~Z. Leibo, R.~Munos,
  C.~Blundell, D.~Kumaran, and M.~Botvinick, ``Learning to reinforcement
  learn,'' \emph{arXiv preprint arXiv:1611.05763}, 2016.

\bibitem{zhuo2019federated}
H.~H. Zhuo, W.~Feng, Q.~Xu, Q.~Yang, and Y.~Lin, ``Federated reinforcement
  learning,'' \emph{arXiv preprint arXiv:1901.08277}, 2019.

\bibitem{ghavamzadeh2016bayesian}
M.~Ghavamzadeh, S.~Mannor, J.~Pineau, and A.~Tamar, ``Bayesian reinforcement
  learning: A survey,'' \emph{arXiv preprint arXiv:1609.04436}, 2016.

\bibitem{barto2003recent}
A.~G. Barto and S.~Mahadevan, ``Recent advances in hierarchical reinforcement
  learning,'' \emph{Discrete Event Dynamic Syst.}, vol.~13, no.~1, pp. 41--77,
  Jan. 2003.

\bibitem{verma2018programmatically}
A.~Verma, V.~Murali, R.~Singh, P.~Kohli, and S.~Chaudhuri, ``Programmatically
  interpretable reinforcement learning,'' in \emph{Proc. Int. Conf. Mach.
  Learn.}, 2018, pp. 5045--5054.

\end{thebibliography}

% add more bib files here


\vskip -4pt plus -1fil

\begin{IEEEbiography}[{\includegraphics[width=1in,height=1.25in,clip,keepaspectratio]{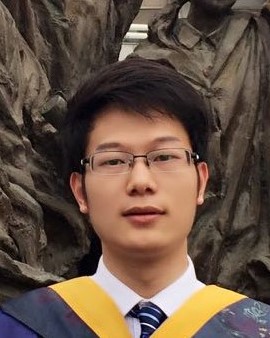}}]{Xin Chen}
received the double B.S. degrees in engineering physics and economics and the master’s degree in electrical engineering from Tsinghua University, Beijing, China, in 2015 and 2017, respectively. He is currently pursuing the Ph.D. degree in electrical engineering with Harvard University, MA, USA. He was a recipient of the Outstanding 
Student Paper Award in 
IEEE Conference on Decision and Control in 2021,  the Best Student Paper Award Finalist in the IEEE Conference on Control Technology and Applications in 2018, and 
the Best Conference Paper Award in the IEEE PES General Meeting in 2016. His research interests lie in data-driven decision-making,
reinforcement learning, distributed optimization and control of networked systems, with applications  to power and energy systems.
\end{IEEEbiography}

\vskip 0pt plus -1fil

\begin{IEEEbiography}[{\includegraphics[width=1in,height=1.25in,clip,keepaspectratio]{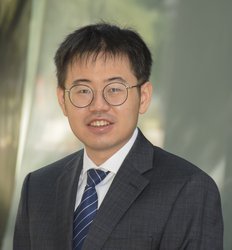}}]{Guannan Qu}
is an assistant professor in the Electrical and Computer Engineering Department of Carnegie Mellon University. He received his B.S. degree in electrical engineering from Tsinghua University in Beijing, China in 2014, and his Ph.D. in applied mathematics from Harvard University in Cambridge, Massachusetts in 2019. He was a CMI and Resnick postdoctoral scholar in the Department of Computing and Mathematical Sciences at California Institute of Technology from 2019 to 2021. He is the recipient of Caltech Simoudis Discovery Award, PIMCO Fellowship, Amazon AI4Science Fellowship, and IEEE SmartGridComm Best Student Paper Award. His research interest lies in control, optimization, and machine/reinforcement learning with applications to power systems, multi-agent systems, Internet of things, and smart cities.
\end{IEEEbiography}

\vskip 0pt plus -1fil

\begin{IEEEbiography}[{\includegraphics[width=1in,height=1.25in,clip,keepaspectratio]{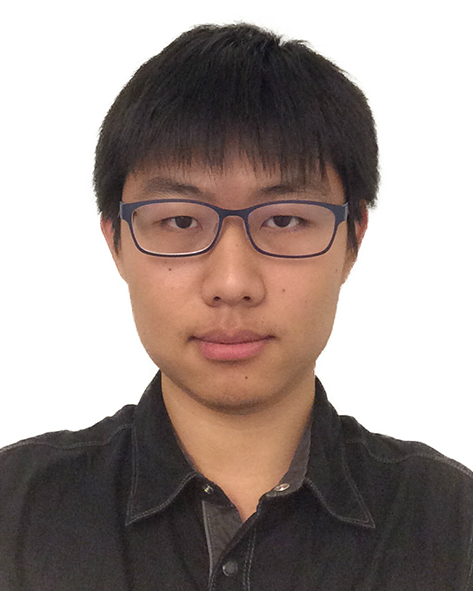}}]{Yujie Tang}
received the B.S. degree in electronic engineering from Tsinghua University, Beijing, China, in 2013, and the Ph.D. degree in electrical engineering from the California Institute of Technology, Pasadena, CA, USA, in 2019.
He is currently a Postdoctoral Fellow with the School of Engineering and Applied Sciences, Harvard University, Allston, MA, USA. His research interests include distributed optimization, control and reinforcement learning, and their applications in cyber-physical networks.
\end{IEEEbiography}

\vskip 0pt plus -1fil

\begin{IEEEbiography}[{\includegraphics[width=1in,height=1.25in,clip,keepaspectratio]{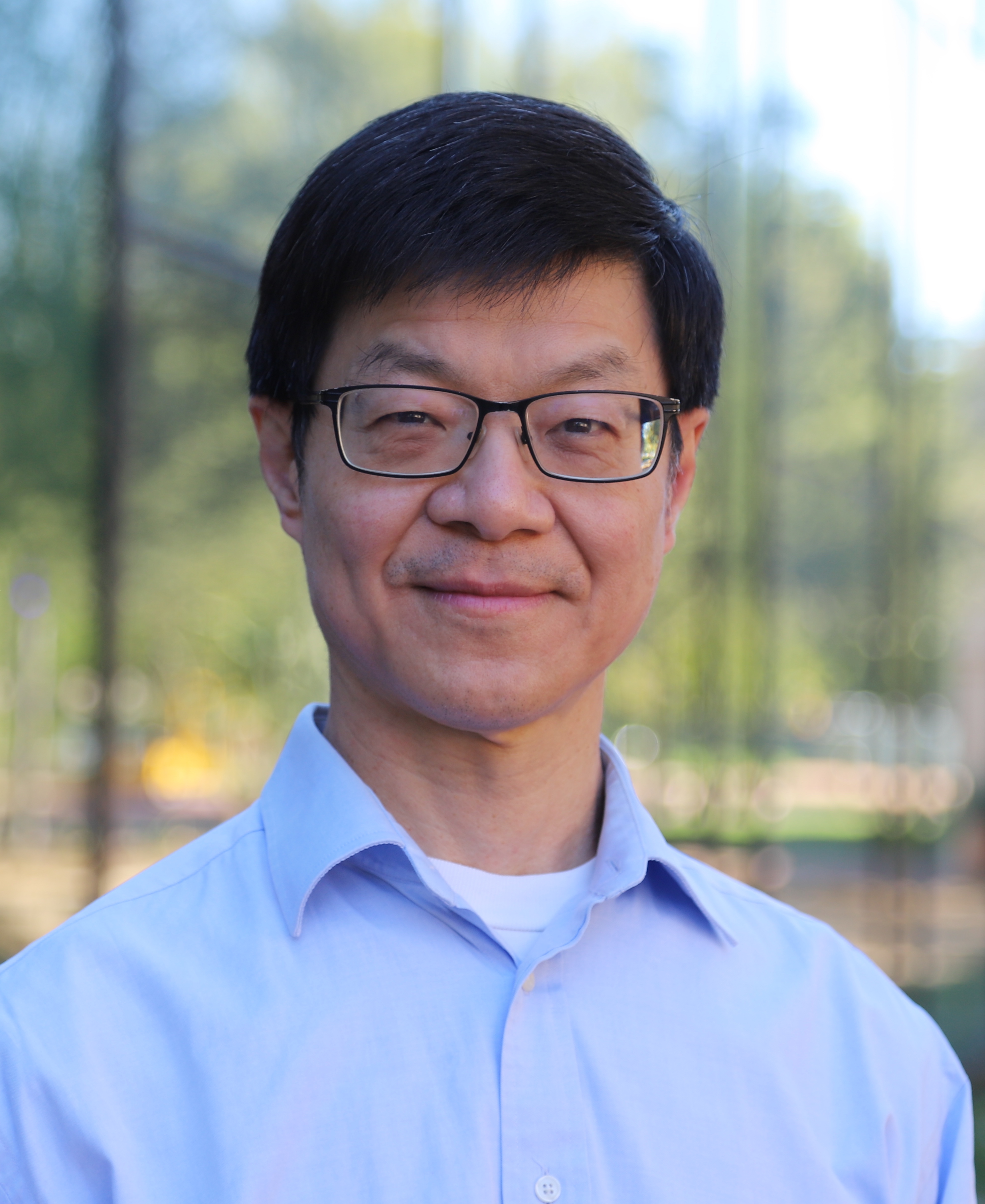}}]{Steven Low}
(F’08) is the F. J. Gilloon Professor of the Department of Computing \& Mathematical Sciences and the Department of Electrical Engineering at Caltech, and an Honorary Professor of the University of Melbourne.  Before that, he was with AT\&T Bell Laboratories, Murray Hill, NJ, and the University of Melbourne, Australia.  He was a co-recipient of IEEE best paper awards, an awardee of the IEEE INFOCOM Achievement Award and the ACM SIGMETRICS Test of Time Award, and is a Fellow of IEEE, ACM, and CSEE.  He was well-known for work on Internet congestion control and semidefinite relaxation of optimal power flow problems in smart grid.  His research on networks has been accelerating more than 1TB of Internet traffic every second since 2014.  His research on smart grid is providing large-scale cost effective electric vehicle charging to workplaces.  He received his B.S. from Cornell and Ph.D. from Berkeley, both in EE.
\end{IEEEbiography}

\vskip 0pt plus -1fil

\begin{IEEEbiography}[{\includegraphics[width=1in,height=1.25in,clip,keepaspectratio]{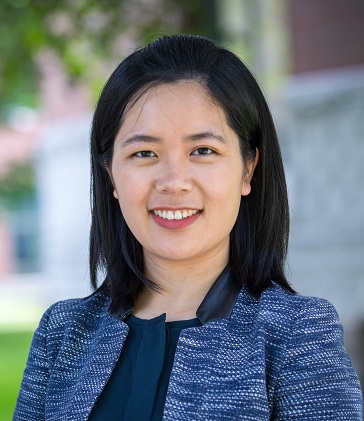}}]{Na Li}
 is a Gordon McKay professor in Electrical Engineering and Applied Mathematics at Harvard University. She received her Bachelor degree in Mathematics from Zhejiang University in 2007 and Ph.D. degree in Control and Dynamical systems from California Institute of Technology in 2013. She was a postdoctoral associate at Massachusetts Institute of Technology 2013-2014. Her research lies in control, learning, and optimization of networked systems, including theory development, algorithm design, and applications to real-world cyber-physical societal system in particular power systems. She received NSF career award (2016), AFSOR Young Investigator Award (2017), ONR Young Investigator Award(2019), Donald P. Eckman Award (2019), McDonald Mentoring Award (2020), along with some other awards.
\end{IEEEbiography}


\end{document}